%% file: main.tex
\documentclass[10pt]{article}

\usepackage[preprint]{tmlr}
\usepackage{graphicx}
\usepackage{flafter}
\usepackage{placeins}
\usepackage{booktabs}
\usepackage{algorithm}
\usepackage{algpseudocode}
\usepackage[table]{xcolor}
\input{style/colors}
\usepackage{style/paper}
\usepackage{subcaption}
\usepackage{thmtools}
\usepackage{hyperref}
\usepackage{orcidlink}
\hypersetup{
  colorlinks=true,
  linkcolor=primary,
  citecolor=primary,
  urlcolor=accent,
  pdftitle={Contraction-Gauge Preconditioning for Quantized Matrix Multiplication},
  pdfauthor={Piyush Sao, Narasinga Miniskar, Pedro Valero-Lara, Keita Teranishi, and Sudip Seal}
}
\usepackage{cleveref}

\declaretheoremstyle[headfont=\bfseries\color{primary},notefont=\normalfont,
  bodyfont=\itshape,headpunct={.},spaceabove=0pt,spacebelow=0pt]{thmplain}
\declaretheoremstyle[headfont=\bfseries\color{defcolor},notefont=\normalfont,
  bodyfont=\normalfont,headpunct={.},spaceabove=0pt,spacebelow=0pt]{thmdef}
\declaretheoremstyle[headfont=\itshape\bfseries\color{neutral},notefont=\normalfont,
  bodyfont=\normalfont,headpunct={.},spaceabove=0pt,spacebelow=0pt]{thmrem}

\declaretheorem[style=thmplain,name=Theorem,numberwithin=section]{theorem}

\declaretheorem[style=thmplain,name=Proposition,sibling=theorem]{proposition}
\declaretheorem[style=thmplain,name=Corollary,sibling=theorem]{corollary}

\declaretheorem[style=thmrem,name=Remark,sibling=theorem]{remark}

\tcbset{
  thmbox/.style={enhanced,breakable,sharp corners,boxrule=0.7pt,
    colback=primary!6,colframe=black!55,
    borderline west={2.5pt}{0pt}{primary},
    left=10pt,right=8pt,top=6pt,bottom=6pt,before skip=8pt,after skip=8pt},
  defbox/.style={enhanced,breakable,sharp corners,boxrule=0.7pt,
    colback=defcolor!7,colframe=black!55,
    borderline west={2.5pt}{0pt}{defcolor},
    left=10pt,right=8pt,top=6pt,bottom=6pt,before skip=8pt,after skip=8pt},
  rembox/.style={enhanced,breakable,sharp corners,boxrule=0.7pt,
    colback=neutral!5,colframe=black!55,
    borderline west={2.5pt}{0pt}{neutral},
    left=10pt,right=8pt,top=6pt,bottom=6pt,before skip=8pt,after skip=8pt},
}
\tcolorboxenvironment{theorem}{thmbox}
\tcolorboxenvironment{lemma}{thmbox}
\tcolorboxenvironment{proposition}{thmbox}
\tcolorboxenvironment{corollary}{thmbox}
\tcolorboxenvironment{definition}{defbox}
\tcolorboxenvironment{claim}{defbox}
\tcolorboxenvironment{remark}{rembox}

\crefname{appendix}{Appendix}{Appendices}
\Crefname{appendix}{Appendix}{Appendices}

\title{\raggedright
Contraction-Gauge Preconditioning for\\
Quantized Matrix Multiplication\thanks{This manuscript has been authored by
UT-Battelle, LLC under Contract No.\ DE-AC05-00OR22725 with the U.S. Department
of Energy. The publisher, by accepting the article for publication,
acknowledges that the United States Government retains a non-exclusive,
paid-up, irrevocable, world-wide license to publish or reproduce the published
form of this manuscript, or allow others to do so, for United States Government
purposes. The Department of Energy will provide public access to these results
of federally sponsored research in accordance with the DOE Public Access Plan
(\url{http://energy.gov/downloads/doe-public-access-plan}).}}
\author{
  \name Piyush Sao\,\orcidlink{0000-0002-9432-5855}\thanks{Corresponding author.}
    \email saopk@ornl.gov\\
  \name Narasinga Miniskar\,\orcidlink{0000-0001-8259-8891}
    \email miniskarnr@ornl.gov\\
  \name Pedro Valero-Lara\,\orcidlink{0000-0002-1479-4310}
    \email valerolarap@ornl.gov\\
  \name Keita Teranishi\,\orcidlink{0000-0001-6647-2690}
    \email teranishik@ornl.gov\\
  \name Sudip Seal\,\orcidlink{0000-0003-3233-0656}
    \email sealsk@ornl.gov\\
  \addr Oak Ridge National Laboratory, Oak Ridge, Tennessee 37831, USA
}

\begin{document}
\maketitle

\begin{abstract}
\input{src/abstract}
\end{abstract}

\input{src/notation}

\input{src/introduction}
\input{src/related}
\input{src/model}
\input{src/scaling}
\input{src/partition}
\input{src/coherence}
\input{src/householder}
\input{src/hier}
\input{src/quantizer}
\input{src/experiments}
\input{src/discussion}
\input{src/extensions}
\input{src/conclusion}
\input{src/acknowledgments}

\bibliographystyle{tmlr}
\bibliography{references}

\appendix
\crefalias{section}{appendix}
\crefalias{subsection}{appendix}
\input{src/quantizer_appendix}
\input{src/certificate_appendix}
\input{src/appendix}
\input{src/reproducibility}
\input{src/related_chronology}

\end{document}

%% file: style/colors.tex
\definecolor{primary}{HTML}{006BA2}
\definecolor{accent}{HTML}{E3120B}
\definecolor{neutral}{HTML}{4A4A4A}
\definecolor{boxfill}{HTML}{E8F4FD}
\definecolor{boxborder}{HTML}{006BA2}
\definecolor{defcolor}{HTML}{2E7D32}
\definecolor{todobg}{HTML}{FFF4E5}
\definecolor{todoframe}{HTML}{B45309}

%% file: src/abstract.tex
We study low-precision computation of $C=AB$ with both factors quantized. We
derive an exact finite-dimensional identity for the expected squared product
error under mutually independent, zero-mean entrywise errors with known
variance fields. It applies
exactly to non-overloading subtractive dither and to independent stochastic
rounding with input-dependent variances; we empirically assess deterministic
round-to-nearest (RTN). Using the product-preserving equivalence
$AB=(AT)(T^{-1}B)$, we formulate \emph{contraction-gauge preconditioning} by
jointly choosing a factor representation and its sharing pattern before
quantization.

Preconditioning can reduce product error but may require additional transformed
copies of an operand. We measure this reuse cost by the number of transformed,
quantized copies of the opposite factor: a shared transform requires one copy;
a block-specific transform can require one per block. Within the
bounded family of positive diagonal gauges, or \emph{folds}, we show that a
globally optimal shared fold can be computed using a \emph{geometric
program} and that a linear program can decide
whether the identity fold is already optimal. For other families, we derive
computable statistics to guide selection: a tail index for scaling, profile
spread for partitioning, coherence and weighted-Gram energy for full and
partial rotations, and slice-energy covariance for hierarchy depth. For these
rotations and hierarchies, we develop computable upper bounds to evaluate
heuristic candidates rather than compute exact optima.

We test these predictions in controlled synthetic experiments. Across twelve
linear products from a trained three-block image classifier, median
within-product rank correlations between dither-model predictions and
deterministic-RTN errors are $0.937$ at 8 bits and $0.918$ at 4 bits. Using the
GP fold instead of the identity fold reduces held-out product error by
$18.0\%$ at 8 bits and $20.5\%$ at 4 bits in geometric mean. At each precision,
we achieve a lower geometric-mean error than a baseline selected from a
SmoothQuant-style grid and lower error on ten of twelve products. Quantizing all twelve
products together reduces logit MSE by $15.4\%$ and $26.4\%$, respectively,
relative to the identity-fold model. We thus provide exact stochastic
product-error accounting, globally certified selection within the bounded
diagonal family, and a common objective for evaluating reusable transform
candidates under deterministic RTN.

%% file: src/notation.tex
\begin{table}[t]
\centering
\caption{\textbf{Core notation and terminology.} The complete symbol reference appears in
\Cref{app:notation}.}
\label{tab:notation}
\small
\setlength{\tabcolsep}{4pt}
\ra{1.12}
\rowcolors{2}{gray!25}{white}
\begin{tabular}{@{}ll@{}}
\toprule
Symbol & Description \\
\midrule
$A\in\R^{m\times K},\ B\in\R^{K\times n},\ C=AB$
  & Full-precision factors and their exact product \\
$m,n;\ K$ & Free output dimensions; shared summation index (contraction dimension) \\
$i,j,k;\ I,J,S$ & Entry indices; row block, column block, and contraction slice \\
$\hat A=A+E_A,\ \hat B=B+E_B$
  & Quantized factors and their error matrices \\
$v^A_{ik},v^B_{kj}$ & Entrywise quantization-error variances \\
$b_\bullet,b_{\mathrm{sum}},\Delta,R,c$
  & Operand bit widths, bit-width sum, step, range, and variance coefficient \\
$T,T_r$ & Contraction-gauge choices in $AB=(AT)(T^{-1}B)$ \\
$D,H;\ U,U_t$ & Row-local/domain-shared diagonal folds; full/partial orthogonal gauges \\
$\mathcal P;\ g,s$ & Partition; number and size of blocks or slices \\
$n_{\mathrm{gauge}},n_{\mathrm{opp}}$ & Distinct gauges; opposite-factor quantized-copy count \\
$\mathcal E,\mathcal E_A,\mathcal E_B$
  & Expected product error and one-sided leading terms \\
$P_A,P_B,P_{AB}$ & Operand-allocation and exact cross-term coefficients \\
$\Xi_{\Delta,\ell_{\max}}$ & Truncated characteristic-function diagnostic \\
\bottomrule
\end{tabular}

\medskip
\begin{tabular}{@{}>{\raggedright\arraybackslash}p{0.27\textwidth}p{0.65\textwidth}@{}}
\toprule
Term & Convention \\
\midrule
Noise; error & Noise is the stochastic model; error refers to the matrices
  $E_A,E_B$, their realizations, and propagated product error. \\
Contraction-gauge equivalence & $(A,B)\sim(AT,T^{-1}B)$ for
  $T\in\mathrm{GL}(K)$; choosing $T$ selects an equivalent factor representation. \\
Gauge domain; structural family & A domain is a fixed output block, typically a
  row block, column block, or rectangle $I\times J$, assigned a single gauge; a family
  constrains the form of $T$. A hierarchy uses one structured gauge on one domain. \\
Block-constant; transform reuse & A block-constant pattern assigns one gauge
  within each domain; transform reuse assigns the same gauge across domains. \\
Quantized-copy count & $n_{\mathrm{gauge}}$ counts distinct gauges;
  $n_{\mathrm{opp}}$ counts distinct transformed, quantized opposite-factor
  copies---including quantizer metadata---and serves as a representation-reuse
  descriptor. \\
Scale gauge & In the fold model, the scalar non-identifiability
  $(h,r^A,r^B)\mapsto(\lambda h,\lambda r^A,\lambda^{-1}r^B)$
  preserves the modeled error; we select one representative by imposing a
  condition such as $\sum_k\log h_k=0$. \\
Output-axis scaling & The scaling acts on a free index and is undone after
  multiplication; contraction gauges instead act on the shared index. \\
Row-local fold benchmark & This benchmark is the error infimum when each row may
  use its own diagonal contraction gauge, or fold; it serves as a theoretical reference
  rather than a reusable design. \\
Surrogate & A surrogate is a tractable proxy---an explicit upper bound where
  stated---for comparing transform designs. \\
Dither assumption & Subtractive dither adds a known random offset before quantization and subtracts it
  afterward; without overload, the resulting uniform error is input-independent. \\
Profile coordinates & These coordinates use log magnitudes to turn multiplicative profile ratios into
  additive differences; scalar-norm sorting serves as the one-number baseline. \\
\bottomrule
\end{tabular}
\end{table}

%% file: src/introduction.tex
\section{Introduction}
\label{sec:intro}

Given $A\in\R^{m\times K}$ and $B\in\R^{K\times n}$, we seek low-precision
factors whose product accurately approximates $C=AB$. We call the shared
summation index $K$ the \emph{contraction dimension} by tensor-network
convention. This task recurs throughout
numerical computing and underlies many large-model workloads. Quantization loss
does not enter the product uniformly: perturbing $A_{ik}$ changes an output row in
proportion to the energy of $B_{k,:}$, and perturbing $B_{kj}$ changes an output
column in proportion to the energy of $A_{:,k}$. A product-level design must
therefore weight each factor's error by the other factor's energy.
Activation outliers in transformer
models can create this imbalance: one outlier can
set its group's range, while the opposite channel's energy determines its error
propagation.

Existing work controls this sensitivity with folds, channel grouping, and
orthogonal transforms, while product-weighted analyses supply a principled
objective. We connect these lines with comparative rules for when
transform families substitute, compose, or justify additional copies
(\Cref{sec:related}).

The factor pair admits equivalent representations along the shared
index. Every
invertible $T\in\mathrm{GL}(K)$ preserves the product:
\begin{equation}
AB=(AT)(T^{-1}B).
\label{eq:preserve}
\end{equation}
We call $(A,B)$ and $(A',B')$ \emph{contraction-gauge equivalent} when
$(A',B')=(AT,T^{-1}B)$ for some $T\in\mathrm{GL}(K)$. This $T$ selects a
\emph{contraction gauge}; contraction-gauge preconditioning chooses this
representation before quantization. Like numerical preconditioning,
this changes the representation while preserving the target computation.
Output-axis scaling instead acts on a free index and is undone after
multiplication, whereas contraction gauges act on the shared index.

A \emph{gauge domain} is a fixed output block (typically a row block, column
block, or rectangle $I\times J$) assigned one gauge. A \emph{block-constant}
domain pattern uses one gauge per domain, though gauges
may differ between domains. When domains share one gauge, we call the pattern
\emph{transform reuse}. Such reuse can reduce the number of transformed,
quantized copies
of the opposite factor that must be stored or produced; we denote this
\emph{quantized-copy count} by $n_{\mathrm{opp}}$. This reuse descriptor records
representation multiplicity, which can affect storage and memory
traffic; we measure platform runtime and bandwidth separately.

The gauge-domain pattern and the structural family of $T$ are independent design
choices. A
positive diagonal gauge is a \emph{fold} rescaling
coordinates as $(AD,D^{-1}B)$; an orthogonal gauge is a rotation; and a
block-diagonal orthogonal gauge within one output domain is a hierarchy.
Within any
family, we choose (i)~$T$, (ii)~its sharing pattern and any internal
contraction slices, and (iii)~the quantizer, including bit allocation, clipping,
and rounding. We evaluate all three choices with one product-error objective.

\begin{decisionbox}
\textbf{Decision statistics.}\par\smallskip
\begin{tabular}{@{}>{\raggedright\arraybackslash}p{0.22\linewidth}>{\raggedright\arraybackslash}p{0.33\linewidth}>{\raggedright\arraybackslash}p{0.36\linewidth}@{}}
\textbf{Statistic} & \textbf{What it measures} & \textbf{Decision it informs}\\
\textbf{Tail index} & heaviness of the range tails & whether per-vector scaling is
worthwhile (\Cref{sec:exp-scaling})\\
\textbf{Profile spread} & within-block range heterogeneity & which rows or columns
to group (\Cref{sec:partition})\\
\textbf{Block coherence} & coordinate energy concentration & whether a full rotation
can help (\Cref{sec:coherence})\\
\textbf{Weighted-Gram spectrum} & low-rank weighted energy concentration & whether a
partial rotation can capture most of the available gain (\Cref{sec:householder})\\
\textbf{Slice-energy covariance} & correlation between factors' slice energies &
whether hierarchical refinement lowers the modeled leading-error surrogate
(\Cref{sec:hier})
\end{tabular}
\par\smallskip
A \emph{surrogate} is a tractable proxy---an explicit upper bound where
stated---used when directly comparing transform families is intractable.
\end{decisionbox}

\paragraph{Noise model.}
Unweighted entrywise error ignores opposite-factor propagation and coupling
between quantized operands. Under independent zero-mean errors, we derive an
exact expected product-error identity with its bilinear cross term
(\Cref{thm:master}) and extend it to
weighted output norms
(\Cref{cor:weighted}). This identity is exact under non-overloading subtractive
dither with independent entrywise dithers (known random offsets added before
quantization and removed afterward) and under independent stochastic rounding
with residue-dependent variances. We test model transfer for deterministic RTN
on controlled products and products from a trained classifier.

\paragraph{Folds and partitions.}
The row-local fold benchmark for $A$-side leading error has the
energy-matched infimum
$c\sum_k\norm{A_{:,k}}_2^2\norm{B_{k,:}}_2^2$ (\Cref{thm:fold}).
Fold selection for one shared domain is a geometric program: a logarithmic
change of variables makes it convex, and finite scale bounds guarantee an
optimizer (\Cref{thm:gpfold}). The program's optimality conditions yield an exact
identity-fold test
(\Cref{prop:foldsplit}). Partitioning then decides which rows share a fold.
Scalar-norm sorting can lose $\Theta(g)$ on a constructed family, whereas
clustering log-magnitude profiles controls regularized spread
(\Cref{prop:sortfail,thm:cluster}).

\paragraph{Rotations and hierarchies.}
Coherence measures coordinate energy concentration. Evaluated on the grouping
blocks, coherence bounds rotation gain; on maximally coherent blocks, random
transforms
come within a logarithmic factor of maximum gain
(\Cref{thm:coherence}). A conditional uniform-weight comparison explains when
rotation removes structure a later fold could exploit
(\Cref{sec:rotation-folding-substitution}). When weighted energy is
low-dimensional, a Householder construction flattens the leading eigenspace
and bounds the remainder by the trailing
spectrum (\Cref{thm:reflector}).

Anti-correlated slice energies favor hierarchical refinement. This criterion
ranks the displayed upper-bound surrogates; we evaluate realized-RTN ordering
empirically. A telescoping identity accumulates node increments and selects the
surrogate-minimizing depth
(\Cref{thm:hier,thm:telescope}).

\paragraph{Quantizer design.}
The optimal continuous high-rate split under $b_A+b_B=b_{\mathrm{sum}}$ obeys
$b_A-b_B=\tfrac12\log_2(P_A/P_B)$ once the transform and sharing pattern are fixed
(\Cref{thm:bitalloc}). A truncated
characteristic-function statistic measures whether input residues cluster on the
quantization lattice, a regime where deterministic rounding may violate the
noise model.
Product weights also determine codebook density, while an exact clipping identity
separates overload bias from granular error (\Cref{thm:clip}).

We build on the classical equivalence
$AB=(AT)(T^{-1}B)$: we couple it to an exact finite-dimensional product-error
identity, formulate transform sharing through gauge domains and
$n_{\mathrm{opp}}$, and derive optimization and selection rules. These rules
include the domain-shared fold GP, profile-aware grouping guarantees, block-local
transform tests, hierarchy telescoping, and a product-aware bit-width split.

The product-error and weighted identities, the clipped-dither identity, the
row-local fold benchmark, the bounded domain-shared fold GP, and the identity-fold
criterion are exact under their stated stochastic assumptions. The coherence,
partial-rotation, and hierarchy results provide explicit upper-bound comparisons.
Controlled and trained-classifier experiments evaluate realized deterministic-RTN
ordering and copy-error tradeoffs.

\paragraph{Scope.}
We cover scalar quantization over diagonal, orthogonal,
partial-rotation, and hierarchical gauge families. The full $\mathrm{GL}(K)$
family, vector and lattice quantizers, and platform cost models define
complementary extensions.

Controlled instances isolate bit allocation, profile clustering, heavy-tail
scaling, and the hierarchy crossover (\Cref{sec:experiments}). A trained
three-block image classifier then tests candidate ranking and shared-fold gains
under deterministic RTN across twelve internal products at two precisions.

%% file: src/related.tex
\section{Related Work}
\label{sec:related}

We analyze scaling, grouping, rotation, and quantizer choices under a
product-weighted error functional and use the opposite-factor quantized-copy
count $n_{\mathrm{opp}}$ as a proxy for transform reuse. We use \emph{gauge} to denote
the inverse-pair freedom on contracted tensor-network bonds, where gauge fixing
selects a representative that canonicalizes the network and improves numerical
conditioning
\citep{evenbly2018gauge,tindall2023gauging}. In our setting, this gauge acts on the
contracted dimension of a matrix product. This usage is distinct from a
positively homogeneous convex gauge function and the associated framework of
gauge optimization and duality~\citep{friedlander2014gauge}.
A one-page chronology of representative milestones appears in
\Cref{app:chronology}.

\paragraph{Scaling, folding, and channel grouping.}
This line of work examines how channel rescaling and grouping can equalize dynamic
ranges. Function-preserving channel rescaling predates recent transformer quantization:
\citet{meller2019factorization} rescale factors across adjacent
layers, and \citet{nagel2019equalization} equalize weight ranges
through scale-equivariant activations. Earlier transformer studies also
identified a small number of high-magnitude, functionally important dimensions
\citep{kovaleva2021bertbusters}. LLM.int8() subsequently showed that systematic
activation outlier features produce large errors in low-precision transformer matrix
multiplication~\citep{dettmers2022llmint8}. SmoothQuant
\citep{xiao2023smoothquant} shifts the activation-outlier burden to the weights
through a diagonal fold, while RPTQ~\citep{yuan2023rptq} reorders and
clusters channels with similar ranges. AWQ searches a grid of activation-derived
per-channel scales to minimize layer-output quantization
error~\citep{lin2024awq}, OmniQuant learns equivalent scales and shifts by
gradient descent~\citep{shao2024omniquant}, and Outlier Suppression+ adds
channel-wise shifting, an affine operation outside the multiplicative fold
family~\citep{wei2023outlier}. MagR, by contrast, preprocesses weights
nonlinearly to reduce their magnitudes while preserving
outputs~\citep{zhang2024magr}. These methods select scales using fixed rules, grid
search, or gradient descent. We instead show that diagonal-fold selection under
a shared contraction gauge is a geometric program: the log-domain problem
solves the bounded shared-fold range-law objective to certified global
optimality, finite lower and upper bounds on the fold scales guarantee an
optimizer, and an exact linear-program
test decides when the identity fold is
optimal (\Cref{prop:foldsplit}). We also show why clustering full magnitude
profiles in log-magnitude coordinates controls regularized spread, whereas
sorting by scalar norms can fail on tied profile classes.

\paragraph{Rotation and learned equivalent transforms.}
QuIP introduced incoherence processing for low-bit quantization
\citep{chee2023quip}; QuaRot, QuIP\#, and SpinQuant extend this with
Hadamard, lattice-codebook, and learned rotations
\citep{ashkboos2024quarot,tseng2024quipsharp,liu2025spinquant}. OSTQuant jointly
learns orthogonal and scaling transformations using a quantization-space
utilization objective~\citep{hu2025ostquant}. FlatQuant learns structured affine
transforms and implements them with Kronecker
factors~\citep{sun2025flatquant}, while AffineQuant optimizes dense
invertible affine transforms with a gradual-mask safeguard for
invertibility~\citep{ma2024affinequant}. \citet{sanjeet2026blockrot} analyze
block-Hadamard outlier suppression non-asymptotically and use
permutations to redistribute mass across blocks. \citet{feng2026rht} prove
near-random-rotation mean-square guarantees for a
dithered randomized Hadamard quantizer. This rotate-then-quantize mechanism,
with explicit mean-squared-error guarantees, predates the LLM line
in distributed mean estimation~\citep{suresh2017distributed,vargaftik2022eden}.
We instantiate this rotate-then-quantize mechanism with the randomized Hadamard construction of
\citet{ailon2006fast}. The closest prior work, CAT, decomposes layer SQNR into
concentration and dominant-direction alignment, then calibrates block linear
transforms to improve both metrics~\citep{federici2026cat}. We evaluate
rotation inside an expected block-local \emph{product-error}
functional, record the additional opposite-factor copies that differing gauges
across gauge domains can require, and derive a slice-energy anti-correlation
criterion and a telescoping identity to determine the contraction-space
refinement depth within a structured shared gauge.

\paragraph{Quantization-error and clipping models.}
\citet{schuchman1964dither} established conditions making
uniform-quantizer error independent of the input;
\citet{sripad1977quantization} characterize when quantization errors are
uniform and white, and \citet{gray1998quantization} survey the broader
high-resolution theory. For scalar products with distributional inputs,
\citet{kuzmin2022fp8} derive an expected quantization-error expression whose
full form includes simultaneous-input terms; \Cref{thm:master} instead gives an
exact finite-dimensional identity for arbitrary fixed matrices and entrywise
variance fields. Unbiased stochastic rounding, as in
QSGD~\citep{alistarh2017qsgd}, also satisfies \cref{eq:noise} when the coordinate
draws and the two operands are sampled independently, with input-dependent
variance fields. The certificate in \Cref{app:certificate} additionally assumes
variance-normalized sub-Gaussian errors, which non-overloading dither satisfies;
stochastic rounding requires a separate step-size-based bound. ACIQ selects clipping
thresholds analytically by
minimizing tensor-level reconstruction error for post-training
quantization~\citep{banner2019aciq}. Our additive model follows the dither
tradition: the product identity is exact for non-overloading subtractive dither
under the stated independence assumptions, and it approximates
deterministic round-to-nearest quantization. Unlike tensor-local clipping
criteria, our clipping identity propagates both factors' errors through the
product and separates overload bias from granular error.

\paragraph{Transform coding and bit allocation.}
Transform coding decorrelates correlated Gaussian sources before
scalar quantization and allocates a fixed bit budget among transform
coefficients~\citep{huang1963block}. More recently, CVXQ formulates scalable
weight-only compression to a prescribed model size through convex
optimization~\citep{young2024cvxq}. In deep networks, Hessian-based sensitivity
drives the assignment of mixed precision across layers~\citep{dong2019hawq}. Our
continuous allocation rule instead splits a fixed bit-width sum between the two
operands of a single product according to their product-weighted leading errors.
A raw-storage constraint weights those widths by the operand dimensions and
therefore defines a different allocation problem.

\paragraph{Product-weighted distortion.}
\citet{ordentlich2024optimal} derive information-theoretic limits and
nested-lattice quantizers for matrix multiplication; NestQuant
and the lookup-table construction make this line of work increasingly practical
\citep{savkin2025nestquant,kaplan2025highrate}. Subsequent high-rate analysis
separates calibration-free two-factor quantization
\citep{ordentlich2026highrate1} from covariance-aware weight-only quantization,
where reverse water-filling allocates the rate and the choice of basis interacts
with rotation~\citep{ordentlich2026highrate2}. Concurrently,
\citet{ang2026product} derive asymptotically optimal scalar densities for a
pair-i.i.d. product model. Layer-wise weight-only methods minimize the same
product-level quantity implicitly: GPTQ's
$\norm{(W-\widehat W)X}_F^2$ reconstruction criterion is the one-sided
counterpart of \cref{eq:master}, with the activation Gram supplying the
opposite-factor weighting~\citep{frantar2023gptq}. These works establish
product-weighted distortion as the optimization objective and study limits,
codebooks, or statistical rate allocation. Building on this common objective,
we derive reuse-aware selection criteria among restricted gauge families: the
domain-shared fold geometric program, the log-magnitude clustering guarantee,
the slice-energy anti-correlation criterion, and the contraction-refinement
telescoping identity. We record representation reuse with $n_{\mathrm{opp}}$;
platform measurements translate that descriptor into runtime and traffic.

%% file: src/model.tex
\section{The Quantized Product-Error Model}
\label{sec:model}

We derive the expected error of quantized matrix multiplication under the
independent zero-mean noise model and show how sharing a contraction gauge changes
the opposite-factor quantized-copy count, our representation-reuse descriptor.

\subsection{Notation and noise model}

Let $A \in \R^{m\times K}$ and $B \in \R^{K\times n}$ be matrices with product
$C = AB$. The shared summation index $K$ is the \emph{contraction dimension};
the free dimensions $m$ and $n$ index the output. For $b\ge2$, we quantize these
matrices using a signed $b$-bit format with maximum magnitude $R$ and scalar step size
$\Delta = R/(2^{b-1}-1)$. The quantized factors are $\hat A = A + E_A$ and
$\hat B = B + E_B$. Different quantization groups can have different ranges and therefore
different per-entry error variances. We record these variances in the
\emph{variance fields} $v^A$ and $v^B$: the entries of $E_A$ and $E_B$ are
mutually independent and satisfy
\begin{equation}
\E(E_A)_{ik} = 0, \quad \E(E_A)_{ik}^2 = v^A_{ik}, \qquad
\E(E_B)_{kj} = 0, \quad \E(E_B)_{kj}^2 = v^B_{kj}.
\label{eq:noise}
\end{equation}
Throughout, \emph{noise} denotes the stochastic model, whereas \emph{error}
denotes the matrices $E_A,E_B$ and their realized values.
Subtractive dithering adds a known random offset before quantization and removes
it after reconstruction. Under non-overloading subtractive
dithering~\citep{schuchman1964dither,sripad1977quantization}, each error entry is uniform on
$[-\Delta/2,\Delta/2]$, with variance
$v = \int_{-\Delta/2}^{\Delta/2}x^2\,dx/\Delta = \Delta^2/12 = c R^2$, where
$c = 1/(12(2^{b-1}-1)^2)$.

For a product-preserving transform, write
$\widetilde A=AT$ and $\widetilde B=T^{-1}B$. A quantization group $G$ in either
transformed factor has range $R_G=\max_{x\in G}\abs{x}$, so every entry in that
group has the common variance $v_G=cR_G^2$ under non-overloading dither. Thus the
transformed factors and their grouping determine the ranges, which in turn determine the
variance fields $v^A$ and $v^B$ used below.

\begin{remark}[Dither convention]
The exact uniform-error law uses the following \emph{non-overloading}
subtractive-dither convention. We add a uniform half-step dither, quantize on the
uniform lattice, and subtract the same dither during reconstruction. With indices
$-q,\ldots,q$ and $q=2^{b-1}-1$, the outer reconstruction level
$R=q\Delta$ lies inside the decision boundaries at
$\pm(q+\tfrac12)\Delta$. If rounding precedes index saturation, values in
$[-R,R]$ remain non-overloading after a half-step dither, apart from
measure-zero ties. Subtracting the dither may move the reconstructed value just
outside $[-R,R]$; the same quantizer-index set still suffices. An implementation
that clips the dithered value to an endpoint before rounding instead introduces
input-dependent overload, which \Cref{sec:clip} models separately.
\end{remark}

\begin{remark}[Independent stochastic rounding]
Let a non-overloading value $x$ lie between adjacent reconstruction values $\ell$ and
$u=\ell+\Delta$. Independent stochastic rounding sets
\begin{equation*}
\Pr(\hat x=u)=\frac{x-\ell}{\Delta},\qquad
\Pr(\hat x=\ell)=\frac{u-x}{\Delta},
\end{equation*}
yielding
\begin{equation*}
\E(\hat x-x)=0,\qquad
\operatorname{Var}(\hat x-x)=(x-\ell)(u-x).
\end{equation*}
With independent draws across entries and operands, \cref{eq:noise} holds with
these residue-dependent variances, so \Cref{thm:master} exactly scores any fixed
transform and grouping. The fold GP of \Cref{thm:gpfold} optimizes the dither
range law $v_G=cR_G^2$; for stochastic rounding, it supplies candidates that can
be rescored exactly with the displayed variances.
\end{remark}

The rounding rule determines whether this independence model is valid. The model in
\cref{eq:noise} is \emph{exact} under the specified non-overloading subtractively
dithered quantizer, where adding and subtracting a known dither makes the error
uniform and independent. For deterministic round-to-nearest, we use the model as
a high-resolution approximation and measure its accuracy empirically. Structured
inputs, such as values concentrated on a lattice, can induce bias and
correlations. As \Cref{sec:scaling} explains, the sharper stochastic $\ell_2$
weighting requires dither, random signs, or an explicit cancellation assumption.

\subsection{The product-error identity}

Transform design requires a product-level objective: minimizing the factors'
entrywise errors separately would ignore both opposite-factor propagation and
the interaction between simultaneous errors.
The perturbed product expands as
$\hat A \hat B - AB = E_A B + A E_B + E_A E_B$. The first two components
propagate one factor's error through the other, yielding opposite-energy-weighted
contributions; $E_AE_B$ yields a bilinear
simultaneous-error contribution. Taking the expected squared Frobenius norm gives
the following product-error identity, which serves as our design objective.

\begin{theorem}[Expected squared product-error identity]
\label{thm:master}
Under the independent zero-mean model \cref{eq:noise},
\begin{equation}
\E\Fnorm{\hat A \hat B - AB}^2
= \sum_{i,k} v^A_{ik}\,\norm{B_{k,:}}_2^2
+ \sum_{k,j} v^B_{kj}\,\norm{A_{:,k}}_2^2
+ \sum_k \Big(\sum_i v^A_{ik}\Big)\Big(\sum_j v^B_{kj}\Big).
\label{eq:master}
\end{equation}
\end{theorem}

\begin{figure}[t]
\centering
\includegraphics[width=\linewidth]{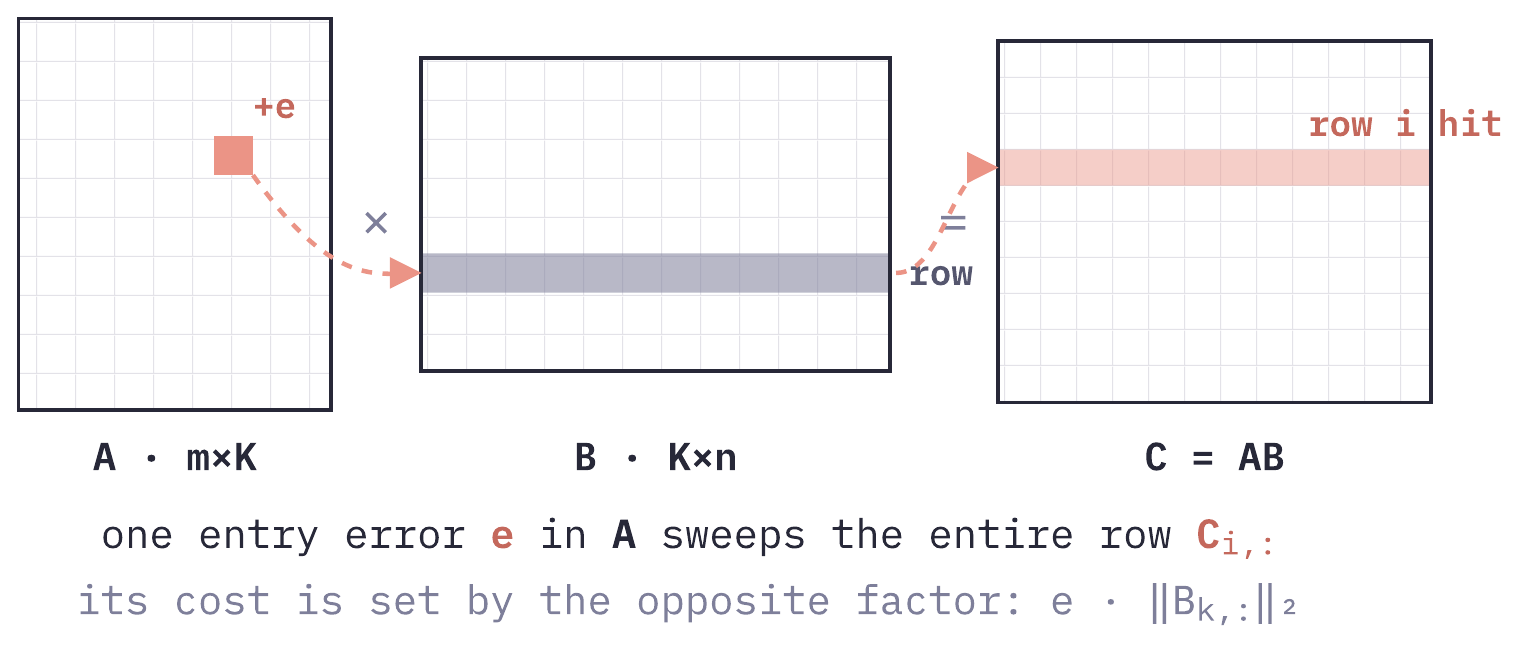}
\caption{\textbf{Opposite-factor weighting of a single entry error.}
An error $(E_A)_{ik}=e$ propagates through row $B_{k,:}$ and perturbs the
entire output row $C_{i,:}$ by $eB_{k,:}$. Its output norm is
$|e|\norm{B_{k,:}}_2$, so its squared contribution is
$e^2\norm{B_{k,:}}_2^2$. Symmetrically, an error in $B_{kj}$ perturbs an output
column and is weighted by $\norm{A_{:,k}}_2^2$.}
\label{fig:error-propagation}
\end{figure}

The first term weights the error $(E_A)_{ik}$ by the row energy
$\norm{B_{k,:}}_2^2$. The second term is its symmetric
counterpart for $B$, and the third is the bilinear contribution of simultaneous
errors in both factors. \Cref{fig:error-propagation} visualizes the first of
these propagation paths.

For fixed factors and dimensions, the third term is $O(c^2)$ as
$c\to0$, whereas the first two are $O(c)$; we call them the
\emph{leading error}. At coarse precision or large contraction dimension, the
bilinear term can remain material.
The identity exposes the design chain:
$T\longrightarrow(\widetilde A,\widetilde B)\longrightarrow
\{R_G\}\longrightarrow(v^A,v^B)\longrightarrow
\E\Fnorm{\hat A\hat B-AB}^2$. The transform $T$ changes the factors; grouping sets
each range $R_G$; and the law $v_G=cR_G^2$ converts ranges into the variance fields
used in \cref{eq:master}. \Cref{fig:design-chain} summarizes this chain.

Equation~\eqref{eq:master} concerns the mean; a single computation uses one noise
realization. \Cref{app:certificate} records a conservative single-run bound for
non-overloading subtractive dither. Calibrating this bound for transform
selection requires an explicit Hanson--Wright constant.

\subsection{Contraction-gauge sharing and quantized-copy count}
\label{sec:reuse-copy}

The equivalence in \cref{eq:preserve} acts on the contraction dimension,
changing both factor representations. A \emph{gauge domain} is a fixed
output block---typically a row block $I_r$, column block $J_r$, or rectangle
$I_r\times J_r$---assigned one contraction gauge $T_r\in\mathrm{GL}(K)$. A
gauge-domain pattern specifies these domains and assignments. A
\emph{block-constant} pattern uses one gauge throughout each domain,
although gauges may differ between domains.

We make two independent choices. The \emph{gauge-domain pattern}
determines where we share each transform, affecting copy count. The
\emph{structural family} constrains the form of $T_r$: diagonal gauges are folds,
orthogonal gauges are rotations, and block-diagonal orthogonal gauges are
hierarchies. Hierarchies specify within-domain transform structure; gauge domains
specify sharing across outputs.

\begin{figure}[t]
\centering
\includegraphics[width=\linewidth]{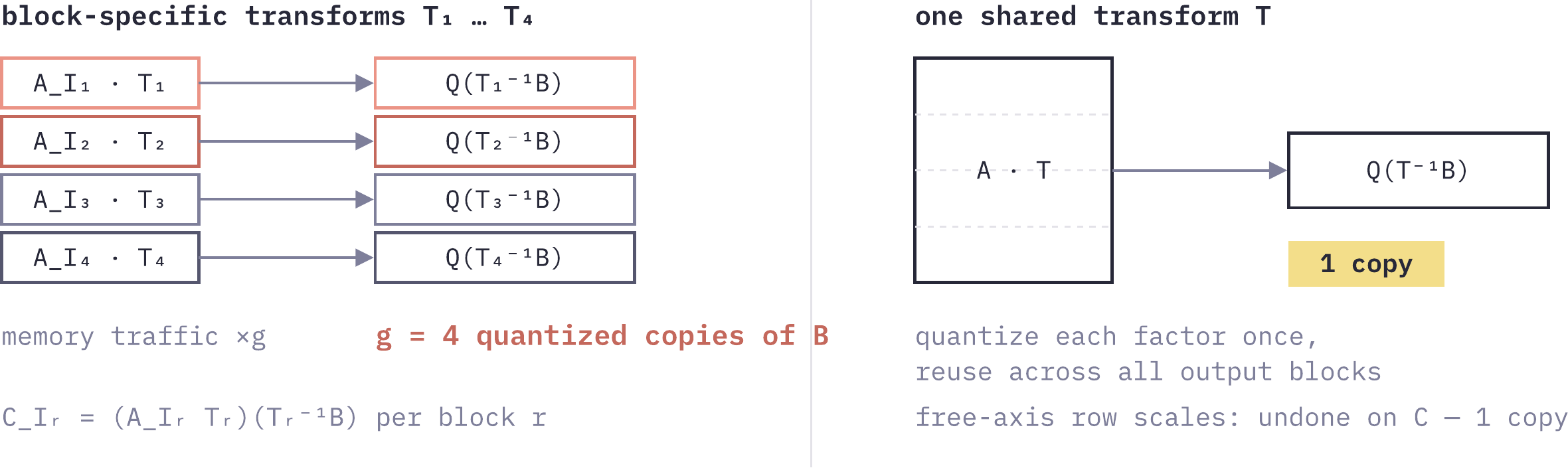}
\caption{\textbf{Gauge sharing controls opposite-factor quantized-copy count.}
With four distinct block-specific gauges, the row blocks pair with four
transformed and quantized representations of $B$ (left). Sharing one gauge
across all row blocks permits one reusable representation (right). The displayed
equalities $n_{\mathrm{opp}}=n_{\mathrm{gauge}}\in\{4,1\}$ assume common
quantizer settings, distinct quantized outputs for distinct gauges, and no
collisions; the surrounding text states the general accounting.}
\label{fig:reuse-copies}
\end{figure}

For row domains $I_1,\ldots,I_g$,
\[
C_{I_r,:}=(A_{I_r,:}T_r)(T_r^{-1}B), \qquad r=1,\ldots,g.
\]
The number of distinct gauge choices is
\[
n_{\mathrm{gauge}}
:=\left|\{T_r:r=1,\ldots,g\}\right|.
\]
Let $\operatorname{Rep}_r(T_r^{-1}B)$ denote the opposite-factor copy that has
been transformed and quantized---whether stored or produced---together with its
quantizer metadata. The \emph{opposite-factor quantized-copy count} serves as a
representation-reuse descriptor:
\[
n_{\mathrm{opp}}
:=\left|\{\operatorname{Rep}_r(T_r^{-1}B):r=1,\ldots,g\}\right|.
\]
The counts agree when all domains use the same quantizer settings and distinct
gauges produce distinct quantized outputs. Transform or quantization collisions can instead make
$n_{\mathrm{opp}}<n_{\mathrm{gauge}}$, whereas domain-dependent quantizer
settings can make $n_{\mathrm{opp}}>n_{\mathrm{gauge}}$ even when gauges coincide.
Thus \emph{transform reuse} means sharing a gauge across domains;
$n_{\mathrm{opp}}$ is the resulting representation-reuse descriptor. Whereas
$n_{\mathrm{opp}}$ describes representation reuse, preprocessing time,
bandwidth, and runtime are platform measurements recorded separately.
\Cref{fig:reuse-copies} contrasts the typical distinct- and shared-gauge cases.

\begin{figure}[t]
\centering
\includegraphics[width=\textwidth]{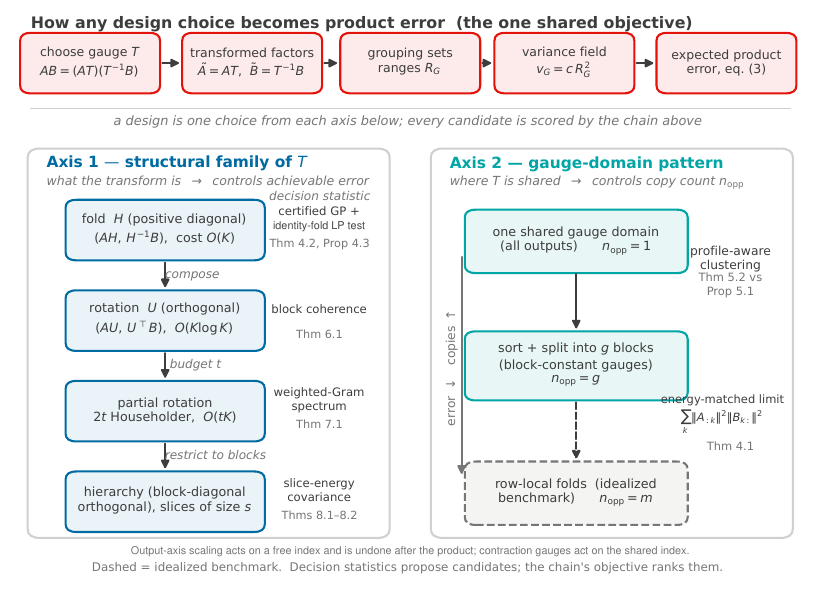}
\caption{\textbf{Gauge design and taxonomy.}
The top chain maps a product-preserving gauge through group ranges to the
variance field in \cref{eq:master}. Below it, diagonal, orthogonal, and
block-diagonal gauges give folds, rotations, and hierarchies, while sorting and
splitting define a block-constant gauge-domain pattern. The domain pattern controls copy count;
the within-domain structure controls achievable error. With a common quantizer
and distinct quantized outputs, $g$ distinct gauges give
$n_{\mathrm{gauge}}=n_{\mathrm{opp}}=g$. Dashed styling marks idealized
benchmarks; output-axis scales act on free indices outside the contraction-gauge
family. See
\Cref{sec:reuse-copy}.}
\label{fig:gauge-framework}
\label{fig:design-chain}
\label{fig:taxonomy}
\end{figure}

Symmetrically, column domains of $B$ produce copies of $AT_r$.
Output-axis row or column scalings act on free indices and are undone on the
output in $O(mn)$ work, so they leave $n_{\mathrm{opp}}=1$ and require no
additional opposite-factor copy. A hierarchy
is one structured block-diagonal orthogonal gauge on a single gauge domain,
typically the full output domain.
\Cref{fig:taxonomy} summarizes the structural gauge families and gauge-domain
patterns.

\subsection{Weighted output norms}
\label{sec:weighted}

Many applications---including Krylov methods, Hessian-weighted
quantization~\citep{frantar2023gptq}, and finite-element analysis---measure error
in a weighted output norm rather than the Frobenius norm. We model such errors by
introducing left and
right weight matrices $\mathsf L$ and $\mathsf R$ and defining the weighted error as
$\Fnorm{\mathsf L(\hat A\hat B - AB)\mathsf R}^2$. The product-error identity generalizes directly to
this weighted-norm setting because the weights simply scale the row and column energies. Here
$e_i$ and $e_j$ denote standard basis vectors of the compatible free dimensions.

\begin{corollary}[Weighted product-error identity]
\label{cor:weighted}
Under the model \cref{eq:noise}, for any matrices $\mathsf L, \mathsf R$ of compatible size,
\begin{align}
\E\Fnorm{\mathsf L(\hat A\hat B - AB)\mathsf R}^2
&= \sum_{i,k} v^A_{ik}\,\norm{\mathsf L e_i}_2^2\,\norm{B_{k,:}\mathsf R}_2^2
+ \sum_{k,j} v^B_{kj}\,\norm{\mathsf L A_{:,k}}_2^2\,\norm{e_j^\top \mathsf R}_2^2 \nonumber\\
&\quad + \sum_{i,k,j} v^A_{ik} v^B_{kj}\,\norm{\mathsf L e_i}_2^2\,\norm{e_j^\top \mathsf R}_2^2 .
\label{eq:weighted}
\end{align}
\end{corollary}

In this identity, the output weights replace raw opposite-factor energy with the
energy of the factors after weighting by $\mathsf L$ and $\mathsf R$, which we
call \emph{opposite-factor sensitivity}.
The identity thus supplies the corresponding weighted objective. Results
whose proofs use only propagated energies can replace them with these
sensitivities. Results involving unweighted range or coherence require the
corresponding weighted argument. This distinction connects the framework to
Hessian-weighted quantization and to the residual- and energy-norm targets of
\Cref{sec:extensions}.

%% file: src/scaling.tex
\section{Scaling and Diagonal Folds}
\label{sec:scaling}

Diagonal preprocessing acts on a free output axis or the shared contraction
axis. Output-axis scales act on a free index and are undone after multiplication;
contraction gauges act on the shared index. A positive diagonal \emph{fold} $D$
is the contraction gauge
$(A,B)\mapsto(AD,D^{-1}B)$, preserving $AB$ while redistributing
contraction-coordinate ranges. We first compare global, block, and per-vector
output-axis schemes, then turn to contraction-axis folds. Refining output-axis groups
weakly decreases modeled error (\Cref{sec:scaling-refinement}), while the row-local
fold benchmark gives an exact diagonal reference that block splitting approximates
(\Cref{thm:fold}). \Cref{app:review} derives the standard output-axis formulas.

Two statistics recur: error on contraction coordinate $k$
propagates through row $k$ of $B$, and the largest entry in a block sets its
shared quantizer range. Thus we write
$\beta_k=\norm{B_{k,:}}_2$, so $\beta_k^2$ is the opposite-factor energy, and
$\alpha_{I,k}=\max_{i\in I}\abs{A_{ik}}$ for the coordinate-$k$ range in row
block $I$.

\subsection{Variance fields and monotone refinement}
\label{sec:scaling-refinement}

To compare output-axis schemes, we use their induced variance fields in the
product-error identity \cref{eq:master}. Per-vector scaling assigns one scale per
row of $A$, giving $v^A_{ik} = c\,(r_i^A)^2$ with
$r_i^A = \norm{A_{i,:}}_\infty$, the per-vector field from \Cref{app:review}.
Row-block scaling instead assigns every row in $I$ the common range
$\rho_I=\max_{i\in I}\norm{A_{i,:}}_\infty$, the largest row range in that
output block. This gives $v^A_{ik} = c\,\rho_I^2$, and its
$A$-side leading error is $c\sum_I \sum_{i\in I}\rho_I^2 \beta_{\mathrm{tot}}$ with
$\beta_{\mathrm{tot}} = \sum_k \beta_k^2$. Refining an output-axis block
partition can only decrease each child range, so its variance field decreases
entrywise. Because every coefficient in \cref{eq:master} is nonnegative, the modeled
error weakly decreases; global scalar, block, and per-vector scaling therefore
form a decreasing refinement chain.

The \emph{contraction-axis} fold objective instead uses the per-coordinate range
$\alpha_{I,k} = \max_{i\in I}\abs{A_{ik}}$. Let
$\mathcal E_A^{\mathrm{fold}}(\mathcal P)$ be the infimum of the $A$-side
leading error when each block $I\in\mathcal P$ may use one shared diagonal fold.
Then
\begin{equation}
\mathcal E_A^{\mathrm{fold}}(\mathcal P) \le c \sum_{I\in\mathcal P} \abs{I} \sum_k \alpha_{I,k}^2 \beta_k^2 .
\label{eq:blockbound}
\end{equation}
For each block, choosing $d_k=1/\alpha_{I,k}$ on the nonzero support normalizes
the folded block range to one and yields the bound in \cref{eq:blockbound}. Coordinates
that are identically zero can be omitted or handled by a limiting argument.
This bound defines a variance field and refinement chain distinct from those of the
output-axis schemes. The row-local fold benchmark below gives the corresponding
row-by-row limit.

\subsection{The row-local fold benchmark}

A fold can outperform output-axis scaling when the opposite factor has little
energy on coordinates with large range. For one row $a$ of $A$, write
$D=\diag(d_k)$. Quantizing $aD$ by its folded range and pairing it with
$D^{-1}B$ gives the $A$-side objective
\begin{equation}
f(d) = \Big(\max_k a_k^2 d_k^2\Big)\Big(\sum_k \beta_k^2/d_k^2\Big),
\end{equation}
the product of the squared folded range and the folded opposite-factor energy.
The benchmark is the infimum of this product. Its value is
\emph{energy-matched} because $\sum_k a_k^2\beta_k^2$ pairs each squared
coordinate magnitude with the energy through which its error propagates, rather
than charging all coordinates through one maximum.

A row-local fold can adapt to each row's range profile, whereas a fold shared by
several rows approximates those profiles with one common transform. To measure the
resulting spread in the block bound, define
\begin{equation}
\gamma_{I,k}
= \frac{\abs{I}\,\alpha_{I,k}^2}{\sum_{i\in I} A_{ik}^2}
\label{eq:fold-spread}
\end{equation}
whenever the denominator is nonzero. Its numerator charges all $\abs{I}$ rows with the
largest squared magnitude on coordinate $k$, whereas its denominator is their
actual squared energy. Hence $1\le\gamma_{I,k}\le\abs{I}$ on this active support;
an identically zero block coordinate contributes nothing and is omitted.

The benchmark value exists as an infimum for every row, but zeros determine
whether a finite fold attains it.

\begin{theorem}[Row-local fold benchmark]
\label{thm:fold}
For any row $a$,
\begin{equation}
\inf_{d_k>0} f(d) = \sum_k a_k^2 \beta_k^2 .
\label{eq:foldoracle}
\end{equation}
If $a=0$, every positive fold attains the value zero. Otherwise, the infimum is
attained if and only if $\beta_k=0$ whenever $a_k=0$; under this condition,
$d_k\propto1/\abs{a_k}$ on the support of $a$ is a minimizing choice. If
$a_k=0$ and $\beta_k>0$ for some $k$, no finite fold attains the infimum; it is
approached by sending the corresponding $d_k\to\infty$. Summing the row-local fold
infima gives the energy-matched benchmark
$\mathcal E_A^\star = c \sum_k \norm{A_{:,k}}_2^2 \beta_k^2$.
\end{theorem}

For a single fold shared across block $I$, the block bound
\cref{eq:blockbound} is
$c\,\abs{I}\sum_k\alpha_{I,k}^2\beta_k^2$. Relative to the coordinatewise
benchmark, its spread on coordinate $k$ is precisely $\gamma_{I,k}$ from
\cref{eq:fold-spread}. Thus $\gamma_{I,k}$ isolates the within-block spread
from sharing one fold across a gauge domain.

The row-local fold benchmark \cref{eq:foldoracle} isolates the infimum of the
$A$-side leading error under row-local diagonal folds. For a nonzero row with
uniform nonzero opposite-factor weights, its improvement over the range-based
value lies between $1$ and $K$. For nonuniform weights, this improvement can
exceed $K$ and become arbitrarily large: a coordinate
with large range can receive negligible opposite-factor weight, making the
range-based baseline arbitrarily worse.
When the infimum is attained, the fold direction
$d_k \propto 1/\abs{a_k}$ follows coordinate range rather than $\beta_k$. Folding by
energy instead may shrink outlier coordinates too little and increase the error
(\Cref{sec:foldsplit}).

The range-following fold direction generally differs by row. We use the
gauge-domain and copy-count conventions of \Cref{sec:reuse-copy}. A gauge shared
over all rows is the one-domain special case $n_{\mathrm{gauge}}=1$. The
row-local gauges considered here are diagonal folds on singleton row domains, so
they can use up to $\abs{I}$ distinct gauges and, under one-copy-per-gauge
accounting, as many transformed and quantized copies $D_i^{-1}B$.
Folds shared on larger domains instead define a block-constant gauge-domain
pattern; \cref{eq:fold-spread} quantifies their within-domain spread. They
interpolate between one global domain and the row-local fold benchmark.
Output-axis per-vector scaling remains a separate baseline. A complete design first scales each
row, then applies a domain-shared fold, and finally substitutes the combined variance field into
the product-error identity. The partitioning theory in \Cref{sec:partition} addresses the
tradeoff between the accuracy of the row-local fold benchmark and the number of
distinct gauges.

\subsection{Domain-shared folds form a geometric program}

One fold shared over a rectangular output domain couples every transformed row
range of $A$ to every transformed column range of $B$, so independent
coordinate-balancing rules can be suboptimal for the joint objective. Auxiliary
range variables can, however, represent these maxima, exposing a geometric
program (GP) with a convex log-coordinate formulation. For a fixed domain
$(I,J)$, let
$H = \diag(h_1,\dots,h_K)$ with $h_k > 0$ be its diagonal contraction gauge, so
$A_{I,:}B_{:,J}=(A_{I,:}H)(H^{-1}B_{:,J})$. Write
$h=(h_1,\dots,h_K)$, and introduce positive variables $r_i^A,r_j^B$ that
upper-bound the transformed row and column ranges:
$r_i^A \ge \abs{A_{ik}}h_k$ and
$r_j^B \ge \abs{B_{kj}}h_k^{-1}$ for every $i \in I$, $j \in J$, and
$k \in [K]$. Define the leading-error epigraph objective
\begin{equation}
\mathcal E_{\rm fold}(h,r^A,r^B) = c\Big[\Big(\sum_{i\in I} (r_i^A)^2\Big)
\Big(\sum_k \norm{B_{k,J}}_2^2 h_k^{-2}\Big)
+ \Big(\sum_{j\in J} (r_j^B)^2\Big)
\Big(\sum_k \norm{A_{I,k}}_2^2 h_k^2\Big)\Big].
\label{eq:foldgp}
\end{equation}
Because identically zero rows of $A_{I,:}$ and columns of $B_{:,J}$ contribute
nothing, we remove them before introducing the positive epigraph variables. If this
preprocessing empties either block, we handle its zero contribution separately.
We henceforth assume both blocks remain nonempty.

A positive \emph{monomial} has the form $a\prod_\ell x_\ell^{p_\ell}$, where
$a>0$ and the exponents are real; a \emph{posynomial} is a sum of such monomials.
After division by their positive right-hand variables, the nonzero range constraints
become monomial inequalities, for example
$\abs{A_{ik}}h_k/r_i^A\le 1$ when $A_{ik}\ne0$; zero-entry constraints are vacuous.
A standard GP minimizes a posynomial subject to inequalities $f_\ell\le1$ with
posynomial $f_\ell$ and equalities $g_q=1$ with monomial $g_q$, all in positive
variables.

\begin{theorem}[Domain-shared fold as a geometric program]
\label{thm:gpfold}
The leading objective \cref{eq:foldgp} is posynomial in $(h,r^A,r^B)$, and the
range constraints are monomial inequalities. The exact simultaneous-error term is
\begin{equation}
\mathcal E_{\rm cross}(r^A,r^B)
=Kc^2\Big(\sum_{i\in I}(r_i^A)^2\Big)
      \Big(\sum_{j\in J}(r_j^B)^2\Big),
\label{eq:foldgp-cross}
\end{equation}
which is also posynomial.
Therefore, the leading objective and the full objective
$\mathcal E_{\rm full}:=\mathcal E_{\rm fold}+\mathcal E_{\rm cross}$ each
define a geometric program. The change of variables
$x_k = \log h_k$, $u_i^A = \log r_i^A$, $u_j^B = \log r_j^B$ makes this program convex.
Every solution of the convex log-domain problem is globally optimal. Here $H$ is
the product-preserving contraction gauge assigned to the domain. Distinct from
that transform choice, the modeled error has a one-dimensional \emph{scale gauge}:
before absolute bounds are imposed, the objective and range constraints are
invariant under
\[
(h,r^A,r^B)\longmapsto
(\lambda h,\lambda r^A,\lambda^{-1}r^B),\qquad \lambda>0.
\]
A compact reduced feasible set guarantees attainment. Finite positive box bounds
$\underline h\le h_k\le\overline h$ provide one compactification after the
zero-row/column preprocessing. A scale-gauge condition such as
$\sum_k\log h_k=0$ or $h_1=1$, together with finite monomial pairwise-ratio
bounds, provides another. The unreduced epigraph feasible set may remain
noncompact because its range variables are unbounded above. Degenerate
disjoint-support inputs can fail to attain a finite infimum in the absence of
either compactification, with some $h_k$ tending to zero or infinity.
\end{theorem}

Taking $I=[m]$ and $J=[n]$ gives the one-domain shared-gauge corollary, with
$n_{\mathrm{gauge}}=1$.

\begin{decisionbox}
\setlength{\abovedisplayskip}{4pt}
\setlength{\belowdisplayskip}{4pt}
\textbf{Two-channel counterexample.}
Let
$A=\left[\begin{smallmatrix}2&3\\3&2\end{smallmatrix}\right]$,
$B=(3,2)^\top$, and $H=\diag(1,t)$ for $t>0$. The tight ranges are
$r_1^A=\max(2,3t)$, $r_2^A=\max(3,2t)$, and
$r_1^B=\max(3,2/t)$. After substituting these tight ranges, abbreviate
$\mathcal E_{\rm fold}(t):=\mathcal E_{\rm fold}((1,t),r^A(t),r^B(t))$.
The squared row/column norm profiles $(9,4)$ and $(13,13)$ then give
\[
\Phi(t)=\frac{\mathcal E_{\rm fold}(t)}{c}
=\begin{cases}
169+104t^{-2},&0<t\le 2/3,\\
234+198t^2+36t^{-2},&2/3\le t\le 3/2,\\
169+234t^2,&t\ge 3/2.
\end{cases}
\]
The first branch decreases. On the middle branch, with $y=t^2$,
$d\Phi/dy=198-36/y^2$ is positive at $y=4/9$ and increases thereafter;
the third branch increases. Hence $t_\star=2/3$ and $\Phi(t_\star)=403$.
Both the range rule
$h_k\propto\sqrt{\max_j\abs{B_{kj}}/\max_i\abs{A_{ik}}}$ and the norm rule
$h_k\propto\sqrt{\norm{B_{k,:}}_2/\norm{A_{:,k}}_2}$ instead select
$t=\sqrt{2/3}$.
\par\smallskip
{\centering
\setlength{\tabcolsep}{5pt}
\renewcommand{\arraystretch}{1.05}
\begin{tabular}{@{}lrrr@{}}
\toprule
Fold & $t$ & $\mathcal E_{\rm fold}/c$ & $\mathcal E_{\rm full}/c$\\
\midrule
Identity fold & $1$ & $468$ & $468+324c$\\
Range/norm & $\sqrt{2/3}$ & $420$ & $420+270c$\\
GP optimum & $2/3$ & $403$ & $403+234c$\\
\bottomrule
\end{tabular}
\par}
\smallskip
Adding \cref{eq:foldgp-cross} changes the outer $y$-dependent terms to
$104(1+c)/y$ and $234(1+c)y$; the middle derivative at its left endpoint is
$63/4+162c>0$ and increases with $y$. Thus $t_\star$ remains globally optimal
for every $c>0$; the heuristic's normalized full-objective gap is
$17+36c>0$, and its leading relative gap is $420/403-1\approx4.22\%$ for this
counterexample.
\end{decisionbox}

The domain-shared fold balances the active per-row and per-column ranges, whereas the
two coordinatewise heuristics balance their statistics independently.
\Cref{fig:gpfold} measures how large this distinction can become on a constructed
family. The GP representation depends on epigraph variables that upper-bound the
entries; their maximum defines each scale. Eliminating these variables by
substituting the maxima yields a convex objective only after the log parameterization
$x_k=\log h_k$, but obscures the standard GP form.

\paragraph{Deterministic worst-case weighting.}
Under deterministic round-to-nearest, each entrywise error satisfies
$\abs{e_k}\le\Delta/2$. If the rows $B_{k,:}$ are aligned with equal norm,
choosing $e_k=\Delta/2$ gives
$\norm{eB}_2=\tfrac{\Delta}{2}\sum_k\beta_k$, a factor $\sqrt K$ larger than the
corresponding $\ell_2$-weighted expression. Thus the sharper stochastic weighting
requires dither, random signs, or an explicit cancellation assumption.

\subsection{Block-local folds: an exact identity-fold optimality criterion}
\label{sec:foldsplit}

For a chosen rectangular domain, the first design question is whether a
nontrivial fold improves on the identity fold. Deciding strict improvement
requires the full block profiles. Convexity lets directional derivatives at the
identity fold answer this question exactly.

For a rectangular domain $(I,J)$, parameterize the fold by
$H(x)=\diag(e^{x_1},\ldots,e^{x_K})$, so $x=0$ is the identity fold, and define
\begin{align*}
R_A^{\mathrm{fold}}(x)&=\sum_{i\in I}\max_k A_{ik}^2e^{2x_k},
&W_A^{\mathrm{fold}}(x)&=\sum_k\norm{A_{I,k}}_2^2e^{2x_k},\\
R_B^{\mathrm{fold}}(x)&=\sum_{j\in J}\max_k B_{kj}^2e^{-2x_k},
&W_B^{\mathrm{fold}}(x)&=\sum_k\norm{B_{k,J}}_2^2e^{-2x_k}.
\end{align*}
The exact block contribution of \cref{eq:master} is
\begin{equation}
F_{I,J}(x)
=c\big[R_A^{\mathrm{fold}}(x)W_B^{\mathrm{fold}}(x)
+R_B^{\mathrm{fold}}(x)W_A^{\mathrm{fold}}(x)\big]
+Kc^2R_A^{\mathrm{fold}}(x)R_B^{\mathrm{fold}}(x).
\label{eq:block-fold-full}
\end{equation}
We may omit the last term when optimizing only the leading error. Multiplying
all diagonal entries of $H$ by the same scalar leaves \cref{eq:block-fold-full}
unchanged. This invariance is the modeled-error scale gauge, distinct from the
product-preserving contraction gauge $H$. We fix the scale gauge by imposing
$\mathbf1^\top x=0$.

At the identity fold, define
\begin{align*}
a_k&=\norm{A_{I,k}}_2^2,
&b_k&=\norm{B_{k,J}}_2^2,\\
\alpha_i^2&=\max_k A_{ik}^2,
&S_i^A&=\arg\max_k A_{ik}^2,\\
\chi_j^2&=\max_k B_{kj}^2,
&S_j^B&=\arg\max_k B_{kj}^2,
\end{align*}
and abbreviate
$R_A=\sum_i\alpha_i^2$, $R_B=\sum_j\chi_j^2$,
$W_A=\sum_k a_k$, and $W_B=\sum_k b_k$. For a direction $d$, set
\begin{align*}
\dot R_A(d)&=2\sum_i\alpha_i^2\max_{k\in S_i^A}d_k,
&\dot W_A(d)&=2\sum_k a_kd_k,\\
\dot R_B(d)&=-2\sum_j\chi_j^2\min_{k\in S_j^B}d_k,
&\dot W_B(d)&=-2\sum_k b_kd_k.
\end{align*}

The active sets $S_i^A,S_j^B$ identify which coordinates currently determine
each quantizer range. The proposition combines their directional effects into an
exact linear-program test.

\begin{proposition}[Computable identity-fold criterion]
\label{prop:foldsplit}
The function $F_{I,J}$ is convex, and its directional derivative at the identity-fold
point is
\begin{align}
F'_{I,J}(0;d)
&=c\big[\dot R_A(d)W_B+R_A\dot W_B(d)
       +\dot R_B(d)W_A+R_B\dot W_A(d)\big]\nonumber\\
&\quad+Kc^2\big[\dot R_A(d)R_B+R_A\dot R_B(d)\big].
\label{eq:fold-directional}
\end{align}
For the leading-only objective, omit the second line. This derivative is convex
and piecewise linear in $d$, so
\[
\eta_{I,J}
=\min_{\mathbf1^\top d=0,\ \norm{d}_\infty\le1}F'_{I,J}(0;d)
\]
is a linear program after epigraph reformulation. On this scale-gauge-fixed
constraint set, the identity fold is optimal if and only if
$\eta_{I,J}=0$; a strict improvement exists if and only if $\eta_{I,J}<0$.
\end{proposition}

\begin{figure}[t]
\centering
\begin{subfigure}[t]{0.49\textwidth}
\centering
\includegraphics[width=\textwidth]{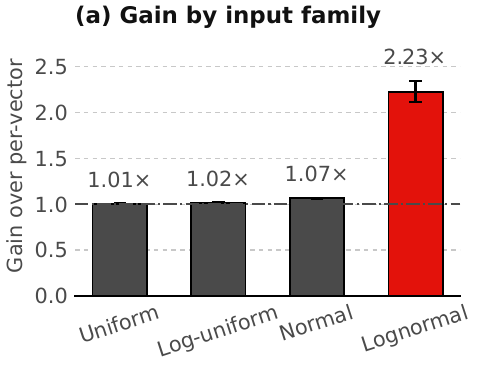}
\phantomsubcaption
\label{fig:foldgain-family}
\end{subfigure}\hfill
\begin{subfigure}[t]{0.49\textwidth}
\centering
\includegraphics[width=\textwidth]{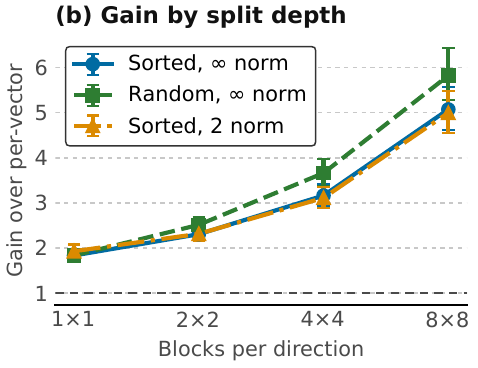}
\phantomsubcaption
\label{fig:foldgain-depth}
\end{subfigure}
\caption{Heuristic folding gains over per-vector scaling.
\textup{(a)} Gain by input family for $2\times2$ blocks on
$256\times256\times256$ products. The comparison uses 20 independent instances;
error bars are $95\%$ log-Student-$t$ intervals for geometric mean
within-instance gain ratios. The displayed range heuristic is
$d_k=\sqrt{\max_j\abs{B_{kj}}/\max_i\abs{A_{ik}}}$;
\Cref{prop:foldsplit} gives the exact identity-fold test through
\cref{eq:fold-directional}.
\textup{(b)} Gain versus split depth on lognormal
$256\times256\times256$ products, using 20 independent instances per depth.
The curves compare range- and energy-based folds under sorted and random
partitions, with $95\%$ log-Student-$t$ intervals for geometric mean
within-instance gain ratios. The intervals quantify the displayed
configurations; the preferred heuristic can vary with the input family. See
\Cref{sec:foldsplit}.}
\label{fig:foldgain}
\end{figure}

\Cref{fig:foldgain-family,fig:foldgain-depth} show how two deployable fold
heuristics vary by input family and split depth.

%% file: src/partition.tex
\section{Partitioning Theory}
\label{sec:partition}

Partitioning selects the gauge domains over which a contraction transform is
shared. Coarser partitions lower the possible gauge and copy counts but increase
within-domain spread; finer partitions reverse the tradeoff. A common baseline,
\emph{scalar-norm sorting}, sorts one norm per row and cuts contiguous domains.
Because a scalar norm discards which coordinates are large, tied keys can mix
incompatible profiles and incur an expected $\Theta(g)$ loss on a constructed family
(\Cref{prop:sortfail}). In contrast, clustering log-magnitude profiles retains coordinate-magnitude information
(\Cref{thm:cluster}); for rank-one profiles, an interval dynamic program is exact.

\subsection{Scalar-norm sorting can lose a linear factor}

Equal-norm rows can have disjoint supports, yet scalar sorting treats them as
interchangeable. The proposition quantifies this random-tie penalty
without adversarial ordering.

\begin{proposition}[Random-tie scalar-norm sorting failure]
\label{prop:sortfail}
Construct $g$ pairwise disjoint-support profile classes with $g$ identical
unit-norm rows per class and uniform opposite-factor weights. If scalar-norm sorting breaks
ties uniformly at random and forms $g$ contiguous blocks of size $g$, measure a
partition by the unweighted block-bound cost
$\mathcal C(\mathcal P)=\sum_{I\in\mathcal P}\abs I\sum_k\alpha_{I,k}^2$.
Write $\mathcal C_{\rm sort}$ for the resulting cost and
$\mathcal C_{\rm pure}$ for the profile-pure optimum. Then
\begin{equation}
\E\!\left[\frac{\mathcal C_{\rm sort}}{\mathcal C_{\rm pure}}\right]
=g\left(1-\frac{\binom{g^2-g}{g}}{\binom{g^2}{g}}\right)
=(1-e^{-1}+o(1))g .
\label{eq:random-tie-gap}
\end{equation}
\end{proposition}

The expectation counts distinct profile classes in a random size-$g$ block;
\Cref{app:proofs} gives the calculation. Thus a scalar key can lose a linear
factor without adversarial ordering; this loss motivates profile-aware clustering.

\subsection{Clustering in log-magnitude coordinates}

The magnitude profile of row $i$ is
$(\abs{A_{i1}},\ldots,\abs{A_{iK}})$. Block spread depends on multiplicative
coordinatewise ratios. Logarithms turn these ratios into additive differences.

For a nonempty block $I$, we write $a_{ik}=\abs{A_{ik}}$,
$\alpha_{I,k}=\max_{i\in I}a_{ik}$, and
$\mathcal S_I=\{k:\sum_{i\in I}a_{ik}^2>0\}$. The raw per-coordinate and
worst-coordinate spreads are
\begin{equation}
\gamma_{I,k}
=\frac{\abs{I}\alpha_{I,k}^2}{\sum_{i\in I}a_{ik}^2},\quad k\in\mathcal S_I,
\qquad
\gamma_I=\max_{k\in\mathcal S_I}\gamma_{I,k}.
\label{eq:rawspread}
\end{equation}
On $\mathcal S_I$, $1\le\gamma_{I,k}\le\abs{I}$; zero coordinates are excluded
because their raw ratio is $0/0$.

To handle $\log 0$, we introduce $\tau_{\log}>0$ and define the log-profile
coordinates and maximum-coordinate distance
\begin{equation}
x_i(k)=\log(a_{ik}+\tau_{\log}),\qquad
d(i,i')=\norm{x_i-x_{i'}}_\infty.
\label{eq:logmetric}
\end{equation}
The corresponding regularized spread is
\begin{equation}
\tilde\gamma_{I,k}
=\frac{\abs{I}(\alpha_{I,k}+\tau_{\log})^2}
{\sum_{i\in I}(a_{ik}+\tau_{\log})^2},\quad k\in[K],
\qquad
\tilde\gamma_I=\max_{k\in[K]}\tilde\gamma_{I,k}.
\label{eq:regspread}
\end{equation}
If only one entry is nonzero, $\gamma_{I,k}=\abs{I}$ even though a large
$\tau_{\log}$ makes the log-profile diameter small. Thus the metric $d$ directly
controls regularized spread; common support and a positive lower bound transfer
this control to raw spread.

\begin{theorem}[Regularized spread control]
\label{thm:cluster}
If block $I$ has diameter at most $r$ in the metric $d$, then for every
$k\in[K]$,
\begin{equation}
\max_{i,i'\in I}\frac{a_{ik}+\tau_{\log}}
{a_{i'k}+\tau_{\log}}\le e^r,
\qquad
\tilde\gamma_{I,k}\le e^{2r}.
\label{eq:clustercontrol}
\end{equation}
Consequently, $\tilde\gamma_I\le e^{2r}$. If all rows in $I$ have the same
support $\mathcal S_I$ and
$a_{\min,k}=\min_{i\in I}a_{ik}>0$ for $k\in\mathcal S_I$, then
\begin{equation}
\tau_{\log}\le\epsilon a_{\min,k}
\quad\Longrightarrow\quad
\gamma_{I,k}\le(1+\epsilon)^2\tilde\gamma_{I,k}.
\label{eq:rawtransfer}
\end{equation}
If the premise holds for every $k\in\mathcal S_I$, then
$\gamma_I\le(1+\epsilon)^2\tilde\gamma_I$.
\end{theorem}

For example, let $R^\star$ be the optimal $g$-center radius. Gonzalez's
farthest-point traversal returns a radius $R\le2R^\star$; assigning each profile to
its nearest center gives a block diameter of at most $2R$. Consequently, every
returned block satisfies $\tilde\gamma_I\le e^{4R}\le e^{8R^\star}$.

For rank-one profiles $a_{ik}=s_i c_k$, sorting by $s_i$ makes the optimal partition
contiguous: an exchange removes interleaving without increasing a block maximum.
An interval dynamic program then finds the exact $g$-block optimum in
$O(m^2g)$ time.

For common support, a lower bound
$a_{\min}=\min_{k\in\mathcal S_I}a_{\min,k}$ permits
$\tau_{\log}\le\epsilon a_{\min}$ and the transfer in
\cref{eq:rawtransfer}. We can generate candidates
over a logarithmic grid of $\tau_{\log}$ and select
by the unregularized product-weighted objective
$\sum_{I\in\mathcal P}\abs{I}\sum_k\alpha_{I,k}^2\beta_k^2$ from
\cref{eq:blockbound}. Our experiments fix $\tau_{\log}$ in advance to isolate
the clustering comparison.

Low-dimensional profile structure sharpens the profile-clustering guarantee;
see \Cref{app:latent-profiles} for latent-dimension and tropical refinements.

%% file: src/coherence.tex
\section{Orthogonal Incoherence Preconditioners}
\label{sec:coherence}

Uniform quantization sets a vector's step from its largest coordinate, whereas
product error propagates the vector's total energy. The worst mismatch occurs when a
vector is supported on a single coordinate: its squared $\ell_\infty$ and $\ell_2$ norms are
equal, rather than differing by a factor of $K$. Diagonal scaling reweights the
existing coordinates, whereas an orthogonal transform can spread their energy while
preserving both the $\ell_2$ norm and the product. We quantify this range
reduction by \emph{block-local coherence}, showing that a random transform
controls all vectors in a block within a logarithmic factor with high probability.

One orthogonal matrix $U$ applied over all outputs is a single shared contraction
gauge with $n_{\mathrm{gauge}}=1$. If we instead use distinct
matrices on output blocks, we obtain block-local rotations:
$n_{\mathrm{gauge}}$ equals the number of distinct
rotations. Under the one-copy-per-gauge accounting of
\Cref{sec:reuse-copy}, $n_{\mathrm{opp}}=n_{\mathrm{gauge}}$.

\subsection{Block-local coherence factors}

We measure how much a rotation reduces squared quantizer ranges on the same
output blocks used for folding and partitioning. Fix a block $(I,J)$ and an
orthogonal matrix $U$, and apply it to the factor pair as
$A_{I,:}U$ and $U^\top B_{:,J}$. Define the block coherence sums
\begin{equation}
R_A^{\mathrm{coh}}(I,J;U)
= \sum_{i\in I}\norm{A_{i,:}U}_\infty^2, \qquad
R_B^{\mathrm{coh}}(I,J;U)
= \sum_{j\in J}\norm{U^\top B_{:,j}}_\infty^2,
\end{equation}
so the leading block error is
\begin{equation}
\mathcal E_{I,J}(U) = \frac{1}{12(2^{b-1}-1)^2}\Big[
R_A^{\mathrm{coh}}(I,J;U)\,\norm{B_{:,J}}_F^2
+ R_B^{\mathrm{coh}}(I,J;U)\,\norm{A_{I,:}}_F^2 \Big].
\label{eq:blockcoh}
\end{equation}
The full expected error also contains the bilinear term from
\cref{eq:master}. The $R^{\mathrm{coh}}$ terms in \cref{eq:blockcoh} sum squared
coordinate maxima, while the Frobenius energies are invariant under $U$.
When $\norm{A_{I,:}}_F>0$, we define
$\eta_A(U)=K R_A^{\mathrm{coh}}/\norm{A_{I,:}}_F^2$; when
$\norm{B_{:,J}}_F>0$, we define
$\eta_B(U)=K R_B^{\mathrm{coh}}/\norm{B_{:,J}}_F^2$.
If a block factor has zero Frobenius norm, its leading contribution vanishes, so
we omit its coherence factor.
The factor $K$ calibrates a perfectly flat block to $\eta=1$, whereas a block whose
nonzero vectors are each supported on one coordinate has $\eta=K$. The theorem
first bounds every rotation between these endpoints, then shows that one shared
random $U$ controls all $N_{IJ}$ block vectors with only a logarithmic loss.

\begin{theorem}[Coherence bounds]
\label{thm:coherence}
Each defined coherence factor satisfies $1 \le \eta \le K$. If $U$
is a Haar-random orthogonal matrix, or if a normalized Hadamard matrix of order
$K$ exists and $U=DH$, where $H$ is normalized Hadamard and $D$ has independent
Rademacher diagonal entries, then with probability
$1-\delta$, every vector $z$---either a row of $A_{I,:}$ or a transposed column
of $B_{:,J}$---satisfies
$\norm{zU}_\infty^2 \le \norm{z}_2^2 \cdot
C\log(2KN_{IJ}/\delta)/K$ for a universal constant $C$, where
$N_{IJ} = \abs{I}+\abs{J}$. Consequently,
every defined coherence factor is $O(\log(KN_{IJ}/\delta))$.
\end{theorem}

For the leading equal-bit block surrogate in \cref{eq:blockcoh}, let
$c=1/[12(2^{b-1}-1)^2]$. When both block energies are nonzero,
\begin{equation}
\mathcal E_{I,J}^{\mathrm{lead}}(U)
=\frac{c}{K}\norm{A_{I,:}}_F^2\norm{B_{:,J}}_F^2
\big[\eta_A(U)+\eta_B(U)\big].
\label{eq:rotation-ceiling}
\end{equation}
This gives a computable ceiling on the gain from any baseline
$U_0\in\mathrm O(K)$:
\begin{equation}
\frac{\mathcal E_{I,J}^{\mathrm{lead}}(U_0)}
{\inf_{U\in\mathrm O(K)}\mathcal E_{I,J}^{\mathrm{lead}}(U)}
\le \frac{\eta_A(U_0)+\eta_B(U_0)}{2}\le K.
\end{equation}
This ceiling characterizes the leading, equal-bit, block-local surrogate; it is
not a bound on realized deterministic-RTN error. The full objective includes the
bilinear term, and \Cref{sec:experiments} measures realized deterministic-RTN
error directly.

The standard fast Walsh--Hadamard transform supplies the Hadamard case when $K$
is a power of two. For other dimensions, zero-padding changes the transformed
dimension, and a block-Hadamard alternative changes the mixing structure; each
choice therefore requires a guarantee for its altered setting.

These bounds on $\eta$ are sharp. If both baseline terms have coherence
of order $K$, the theorem guarantees a gain of
$\Omega(K/\log(KN_{IJ}/\delta))$; if both equal one, no orthogonal transform
can lower the leading surrogate. The logarithmic factor is a uniform bound rather
than a measure of realized error; some instances can improve more.
Prior work establishes random-transform incoherence~\citep{ailon2006fast}; the
block-local objective \cref{eq:blockcoh} lets those gains combine across a
partition and supports the flat-versus-hierarchy comparison in \Cref{sec:hier}.

\subsection{Conditional substitution of rotation and folding}
\label{sec:rotation-folding-substitution}

Rotations and folds address large coordinates differently:
a rotation redistributes energy, whereas a diagonal fold transfers coordinate scale
between factors. A rotation that lowers the maximum normalized coordinate
energy can therefore remove variation that a later fold would exploit. For the
one-sided functional with fixed uniform opposite-factor weights, let
$G(a)=K\max_k p_k$, where $p_k=a_k^2/\norm{a}_2^2$. Because the maximum coordinate
is Schur-convex~\citep{hardy1952inequalities}, $p(U)\prec p$ implies
$G(aU)\le G(a)$. This substitution principle therefore applies to fixed profiles
with uniform opposite-factor weights.

Random rotations change the expected profile. For a flat row
$a=\mathbf1/\sqrt K$, $G(a)=1$, whereas a Haar rotation produces
$\E G(aU)=\E\,K\max_k(aU)_k^2=\Theta(\log K)$, so its expected coherence
increases from $1$ to $\Theta(\log K)$. Nonuniform product weights also remove
the ordinary-majorization ordering. For
$G(a;\beta)=\norm{a}_\infty^2\sum_k\beta_k^2/\sum_k a_k^2\beta_k^2$ and
$\beta^2=(M,1)$, the permutation profiles $a^2=(1,0)$ and $a^2=(0,1)$ yield
$1+1/M$ and $M+1$, respectively. Composition under general product weights can
therefore vary with the weighted profile, so we compare candidate transform
chains using the full product-error objective.

%% file: src/householder.tex
\section{Local Householder Reflectors}
\label{sec:householder}

The coherence bound uses a full orthogonal preconditioner that mixes all $K$
coordinates. When most joint weighted energy is concentrated in $t$ directions,
a partial transform can target them at an application cost of $O(tK)$.
We build this transform from Householder reflectors
$I-2vv^\top$, rank-one updates that align one direction at a time. Two
Householder-QR alignments use at most $2t$ of these reflectors to map the selected
source frame and an incoherent target frame through the same reference frame. The
construction maps the selected subspace onto the target subspace, spreading vector
components. The orthogonal transform preserves complementary-subspace energy,
which controls the remainder. A randomized target yields
the stated high-probability guarantee.

A distinct reflector product on each output block creates block-local rotations,
with $n_{\mathrm{gauge}}$ equal to the number of distinct products. Sharing one
product across all outputs keeps $n_{\mathrm{gauge}}=1$. Under the standard
one-copy-per-gauge accounting,
$n_{\mathrm{opp}}=n_{\mathrm{gauge}}$ in both cases.

\subsection{The weighted Gram matrix and a constructive bound}

We can factor the block error in \cref{eq:blockcoh} as
\begin{equation}
\mathcal E_{I,J}(U)
= c\norm{B_{:,J}}_F^2\big[
R_A^{\mathrm{coh}}(I,J;U)+\mu R_B^{\mathrm{coh}}(I,J;U)\big],
\qquad
\mu=\frac{\norm{A_{I,:}}_F^2}{\norm{B_{:,J}}_F^2}.
\label{eq:householder-mu}
\end{equation}
Here $\mu$ is the \emph{balance ratio} of the two rotation-invariant block
energies. Factoring $\norm{B_{:,J}}_F^2$ from the leading block-error expression
forces this ratio, which requires no additional tuning parameter. If either
block factor has zero Frobenius norm, the leading block error vanishes and no
reflector selection is needed. Other positive values of $\mu$ define alternative
design objectives: increasing $\mu$ emphasizes the columns of $B_{:,J}$, whereas
decreasing it emphasizes the rows of $A_{I,:}$.

The weighted Gram matrix guides how we spend the $t$-direction reflector budget:
\begin{equation}
M_\mu = A_{I,:}^\top A_{I,:} + \mu\, B_{:,J} B_{:,J}^\top.
\end{equation}
For a unit direction $x$, its Rayleigh quotient
$x^\top M_\mu x=\norm{A_{I,:}x}_2^2+\mu\norm{B_{:,J}^\top x}_2^2$
is the joint weighted energy along $x$. The eigenvalues
$\lambda_1 \ge \lambda_2 \ge \cdots$ therefore order directions by their
contribution to the objective in \cref{eq:householder-mu}.
For a budget $t$, call the leading eigenspace the \emph{spectral head} and its
orthogonal complement the \emph{tail}. This head contains the most weighted energy
available to any $t$-dimensional subspace. The construction maps its basis to an
incoherent target; the theorem discounts the head by the incoherence factor and
bounds the tail by its unchanged total energy.

\begin{theorem}[Spectral head/tail bound]
\label{thm:reflector}
Let $w_1,\dots,w_t$ be the top-$t$ eigenvectors of $M_\mu$, and let $P_t$ be the
projection onto their span. Draw a target frame $Q\in\R^{K\times t}$ from the Haar
measure on the Stiefel manifold, independent of the
$N_{IJ}=\abs{I}+\abs{J}$ rows and columns being transformed. There is an
orthogonal $U_t$, expressible as a product
of at most $2t$ Householder reflectors, such that
$U_t^\top[w_1,\ldots,w_t]=Q$. With probability at least $1-\delta$ over $Q$, all
$N_{IJ}$ weighted rows and columns $z$ simultaneously satisfy
\begin{equation}
\norm{(zP_t)U_t}_\infty^2
\le C\frac{\log(2KN_{IJ}/\delta)}{K}\norm{zP_t}_2^2
\label{eq:head-incoherence}
\end{equation}
for a universal constant $C$, and hence
\begin{equation}
\mathrm{Obj}(U_t) \;\lesssim\;
\frac{\log(KN_{IJ}/\delta)}{K}\sum_{\ell\le t}\lambda_\ell(M_\mu)
\;+\; \sum_{\ell>t}\lambda_\ell(M_\mu),
\label{eq:headtail}
\end{equation}
where $\mathrm{Obj}(U) = \sum_{i\in I}\norm{A_{i,:}U}_\infty^2
+ \mu\sum_{j\in J}\norm{U^\top B_{:,j}}_\infty^2$.
\end{theorem}

\begin{corollary}[Deterministic Hadamard target]
\label{cor:reflector-hadamard}
Suppose a normalized Hadamard matrix of order $K$ exists. If $Q$ consists of any
$t$ of its columns, possibly after random sign flips and permutations, the same
$2t$-reflector
construction gives the deterministic alternative
\begin{equation}
\mathrm{Obj}(U_t)
\le 2\frac{t}{K}\sum_{\ell\le t}\lambda_\ell(M_\mu)
+2\sum_{\ell>t}\lambda_\ell(M_\mu).
\label{eq:headtail-hadamard}
\end{equation}
\end{corollary}

In \cref{eq:headtail}, the head receives the
$\log(KN_{IJ}/\delta)/K$ discount and the tail retains its full weight. This
logarithm comes from the union bound over all coordinates of the block vectors. A
randomized target gives the logarithmic head discount, whereas a deterministic
Hadamard target retains the $t/K$ worst-case factor under Cauchy--Schwarz; random
signs and permutations preserve that worst-case factor.

The tail bound uses preserved energy rather than bounding individual coordinates:
$\norm{z(I-P_t)U_t}_\infty\le\norm{z(I-P_t)}_2$. Consequently, the top-$t$
eigenspace minimizes the displayed bound precisely when the coefficient on the
head is below one; above that threshold, the estimate supplies no useful
preference for targeting high-energy directions. Computing or sketching
this eigenspace and applying its $O(t)$ reflectors costs $O(tK)$ per vector. A
low-rank weighted-Gram head can therefore capture most of the bound's reduction.
At $t=K$, the construction recovers the bound in \Cref{thm:coherence}.

\begin{remark}[What the reflector budget implies]
\label{rem:hhlower}
If $U$ is a product of $r$ reflectors, then
$\operatorname{rank}(U-I)\le r$. This rank condition supplies a necessary check
on any claimed reflector budget; a matching spectral lower bound requires
additional geometric information and remains open.
\end{remark}

%% file: src/hier.tex
\section{Hierarchy versus Flat Rotation}
\label{sec:hier}

Full rotations (\Cref{sec:coherence}) are flat: they mix all $K$ contraction
coordinates. A hierarchy instead uses one block-diagonal transform shared over
the output domain, making it a structured shared contraction gauge with
$n_{\mathrm{gauge}}=1$; its slices are transform blocks rather than separate gauge
domains. Restricting the transform to blocks reduces mixing strength but increases
selectivity. A slice of
size $s$ has a weaker $1/s$ incoherence factor than the flat transform's $1/K$,
but it couples only the two factors' energies within that slice. Complementary
slice energies can therefore favor the hierarchy.
Application cost depends on the transform: a dense orthogonal or Haar matrix
costs $O(K^2)$ per vector, a fast Walsh--Hadamard transform costs $O(K\log K)$,
$t$ Householder reflectors cost $O(tK)$, and a diagonal fold costs $O(K)$.
A block-diagonal transform costs the sum of its blockwise costs. Negative
correlation between the factors' per-slice energies favors the hierarchical
leading-error surrogate.

\subsection{The anti-correlation criterion}

Partition the contraction dimension into $g$ slices $S_1,\dots,S_g$ of size
$s=K/g$. We compare the flat and hierarchical candidates under one matched
quantization-group convention. For either transformed pair
$(\widetilde A,\widetilde B)$, every row-slice
$\widetilde A_{i,S_r}$ is one quantization group, and every column-slice
$\widetilde B_{S_r,j}$ is one quantization group. Both candidates use
$g(m+n)$ scale groups and one transformed copy of each factor. The flat
candidate uses one full orthogonal gauge
$U\in\mathrm O(K)$; the hierarchical candidate uses the single shared gauge
$U=\operatorname{diag}(U_1,\ldots,U_g)$ with
$U_r\in\mathrm O(s)$. Under the fixed-quantizer accounting of
\Cref{sec:reuse-copy}, both candidates have
$n_{\mathrm{gauge}}=n_{\mathrm{opp}}=1$.

Let $A_r=\norm{A_{:,S_r}}_F^2$ and
$B_r=\norm{B_{S_r,:}}_F^2$ be the slice energies. Both surrogates vanish if
either factor has zero Frobenius norm; below we assume both are nonzero.
The normalized profiles
$p_r=A_r/\norm{A}_F^2$ and $q_r=B_r/\norm{B}_F^2$ each sum to one. Write
$N_{\mathrm{out}}=m+n$. A full transform controls the
$N_{\mathrm{out}}$ length-$K$ rows and columns; its full-vector maximum also
controls every matched slice group. The hierarchy instead controls
$gN_{\mathrm{out}}$ length-$s$ row and column slices. Because $sg=K$,
allocating total failure probability $\delta$ gives the same union-bound penalty
in both cases. Denote this penalty by
\begin{equation}
\Lambda = 2\log(2KN_{\mathrm{out}}/\delta)
  = 2\log(2s(gN_{\mathrm{out}})/\delta).
\label{eq:incoherence-factor}
\end{equation}
Applying \Cref{thm:coherence} at dimensions $K$ and $s$ gives bounds proportional
to the common-normalized leading-error surrogates, with the same universal factor:
\begin{equation}
\mathcal B_{\mathrm{flat}}
= \frac{c\Lambda}{K}\norm{A}_F^2\norm{B}_F^2,
\qquad
\mathcal B_{\mathrm{hier}}
= \frac{c\Lambda}{s}\sum_r A_rB_r.
\end{equation}
The flat surrogate $\mathcal B_{\mathrm{flat}}$ contains the product of total
energies, while the hierarchical surrogate $\mathcal B_{\mathrm{hier}}$ sums
only within-slice energy products. The common
normalization suppresses the same symmetric two-sided and incoherence constants,
which cancel in comparison.

\begin{theorem}[Anti-correlation criterion]
\label{thm:hier}
Under the matched quantization groups above, the ratio of the displayed
hierarchical and flat leading-error surrogates is
\begin{equation}
\frac{\mathcal B_{\mathrm{hier}}}{\mathcal B_{\mathrm{flat}}}
= g\sum_{r=1}^g p_r q_r .
\end{equation}
\end{theorem}

\begin{figure}[t]
\centering
\includegraphics[width=0.96\linewidth]{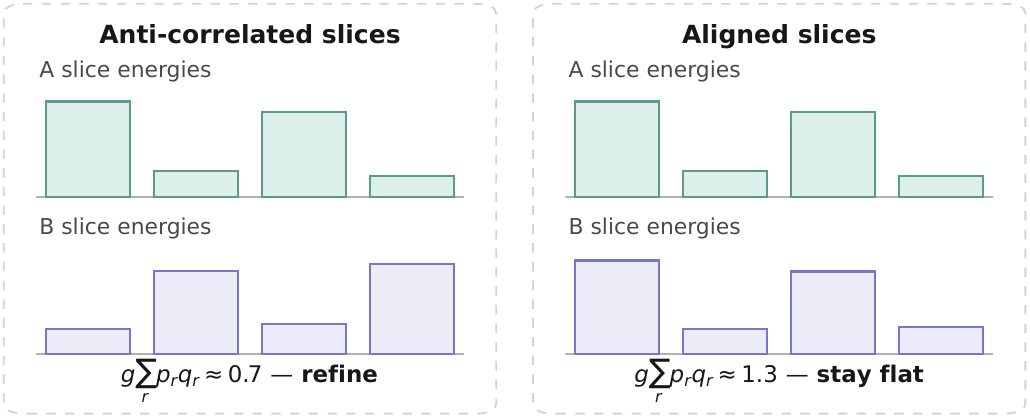}
\caption{\textbf{Slice-energy alignment controls the single-level hierarchy
decision.} The normalized overlap $g\sum_r p_rq_r$ equals the ratio of the
hierarchical and flat leading-error surrogates in \Cref{thm:hier}.
Complementary (anti-correlated) slice-energy profiles produce a ratio below one
and favor refinement (left); aligned profiles produce a ratio above one and
favor the flat candidate (right). Bars and displayed ratios are schematic.}
\label{fig:hierarchy-overlap-schematic}
\end{figure}

Because $p$ and $q$ both have mean $1/g$, the hierarchical surrogate is lower
precisely when their centered inner product is negative:
\begin{equation}
\sum_{r=1}^g \Big(p_r - \tfrac1g\Big)\Big(q_r - \tfrac1g\Big)
=\sum_{r=1}^g p_rq_r-\frac1g<0 .
\label{eq:anticorr}
\end{equation}
This sign orders the two upper-bound surrogates. A negative value means the factors
favor complementary slices, reducing $\sum_rA_rB_r$, while aligned energies favor
the flat surrogate. \Cref{fig:hierarchy-overlap-schematic} visualizes this
ordering; a zero centered inner product gives equality. \Cref{sec:experiments}
evaluates the corresponding realized-RTN crossover on controlled data.

\FloatBarrier

\subsection{Multilevel telescoping and surrogate-optimal depth}

A multilevel hierarchy repeats the single-level decision at every internal node.
Its levels refine the contraction space rather than the gauge domains, so
$n_{\mathrm{gauge}}$ remains one. To measure one such decision, let node $S$
cover a slice of size $K_S$, with energies
$A_S=\norm{A_{:,S}}_F^2$ and $B_S=\norm{B_{S,:}}_F^2$, and define its unsplit
contribution as $\mathcal J(S)=A_SB_S/K_S$.
If $S$ has $g_S$ equal children $R$, define $p_R=A_R/A_S$ when $A_S>0$
and set $p_R=1/g_S$ when $A_S=0$; define $q_R$ analogously. These
zero-energy conventions make both child profiles sum to one and leave the
surrogate unchanged: if $A_SB_S=0$, all descendant contributions and the node increment
vanish. Direct substitution then gives
$\sum_R\mathcal J(R)=g_S(\sum_Rp_Rq_R)\mathcal J(S)$.
Splitting replaces one parent contribution by its child sum. These replacements
telescope without double counting, yielding a bottom-up depth rule.

\begin{theorem}[Telescoping and depth]
\label{thm:telescope}
For any hierarchy tree $\mathcal T$ with root covering all $K$ coordinates, the
total leaf surrogate satisfies the exact identity
\begin{equation}
\mathcal J_{\mathrm{leaves}} = \mathcal J_{\mathrm{root}}
+ \sum_{S \in \mathrm{internal}(\mathcal T)}
\frac{A_S B_S}{K_S}\Big[g_S \sum_{R\in\mathrm{ch}(S)} p_R q_R - 1\Big].
\label{eq:telescope}
\end{equation}
Each bracket is negative exactly when the normalized child energy profiles have
negative centered inner product,
so splitting a node reduces its surrogate if and only if
$g_S\sum_R p_R q_R < 1$. For a fixed candidate split, the bracket gives its
exact surrogate increment after multiplication by $A_SB_S/K_S$.
\end{theorem}

For a binary split, the normalized energy pairs are $(p,1-p)$ for $A$ and
$(q,1-q)$ for $B$. Their orthonormal Haar averages are both fixed at
$1/\sqrt2$; the informative coefficients are the details
$d_A=[p-(1-p)]/\sqrt2$ and $d_B=[q-(1-q)]/\sqrt2$, which record each factor's
left--right energy imbalance. With this normalization, the bracket in
\cref{eq:telescope} is
\begin{equation}
2\big[pq+(1-p)(1-q)\big]-1
=(2p-1)(2q-1)=2d_A d_B.
\label{eq:haar-increment}
\end{equation}
An immediate split helps exactly when the two details have opposite
signs.
This factor of $2$ arises from the orthonormal $1/\sqrt2$ normalization.
Because these covariance increments can change sign across scales, stopping at the
first nonnegative increment can miss a favorable deeper split. Optimal stopping is
thus bottom-up: compare the node surrogate $\mathcal J(S)$ with the sum of the
optimal child surrogates, and expand only when the latter is smaller. For a
levelwise hierarchy, one can equivalently evaluate the cumulative telescoping sum
at every permitted depth and choose its minimum.

For a binary hierarchy, each node increment is
$2\mathcal J(S)d_A d_B$, so the identity is a weighted sum of cross-factor Haar
detail products across scales. A coarse split can be favorable
(anti-correlated pooled energies) even if all finer nested splits are unfavorable
(aligned fine energies), producing an interior surrogate optimum.
In \cref{eq:telescope}, the leaf surrogate is exactly additive. A
deployment-oriented stopping rule can augment this surrogate with level-dependent
incoherence and transform-cost corrections, which can shift the practical
stopping depth.
In \Cref{sec:experiments}, the telescoped increments predict
the surrogate-minimizing depth.

\subsection{Designing the slices}

The preceding criteria evaluate a prescribed slicing and its depth. To construct
the slices, we reduce the design problem to per-coordinate energies by setting
$a_k = \norm{A_{:,k}}_2^2$ and
$b_k = \norm{B_{k,:}}_2^2$. Under this scalar-energy model, single-level slice
design reduces to the optimization problem
\begin{equation}
\min_{S_1,\dots,S_g}\; \sum_{r=1}^g \Big(\sum_{k\in S_r} a_k\Big)\Big(\sum_{k\in S_r} b_k\Big),
\qquad \abs{S_r} = s .
\label{eq:slicedesign}
\end{equation}
Expanding the objective gives
$\sum_r A_rB_r=\sum_k a_kb_k+
\sum_r\sum_{k<\ell\in S_r}(a_kb_\ell+a_\ell b_k)$.
Because the first term is partition-independent, \cref{eq:slicedesign}
is a balanced graph-partitioning problem with within-slice edge weights
$w_{k\ell}=a_kb_\ell+a_\ell b_k$. A simple constructive heuristic sorts coordinates by
$\log(a_k/b_k)$ and groups similar ratios, thereby tending to form
$A$-heavy and $B$-heavy slices. The following example quantifies a factor-two
gap between this heuristic and a mixed partition:
for $(a,b)=(1000,100),(1000,100),(1,1),(1,1)$ and $s=2$, grouping
like ratios costs approximately $4\times10^5$, whereas mixing them
costs approximately $2\times10^5$. The associated decision problem is
weakly NP-complete even for two slices with $a_k=b_k$, although that
restricted case admits a pseudo-polynomial dynamic program; the formal
statement and proof appear in \Cref{prop:slice-hard}.

%% file: src/quantizer.tex
\section{Quantizer Design After the Transform}
\label{sec:quantizer}

Choosing the transform $T$ and the grouping fixes the shape of the variance field.
The remaining scalar-quantizer choices set its resolution, rounding law, and
clipping threshold.

\subsection{Asymmetric allocation under a bit-width-sum constraint}

Opposite-factor amplification can favor assigning more precision to one operand.
To expose this asymmetry without sweeping bit splits, we use the high-rate law
$v\propto2^{-2b}$ and factor the bit dependence as
$v^A_{ik} = \bar v^A_{ik}2^{-2b_A}$ and
$v^B_{kj} = \bar v^B_{kj}2^{-2b_B}$, absorbing bit-independent format constants
into $\bar v^A$ and $\bar v^B$. Substituting these expressions into
\cref{eq:master} gives
\begin{equation}
\mathcal E(b_A, b_B) = P_A\, 2^{-2b_A} + P_B\, 2^{-2b_B} + P_{AB}\, 2^{-2(b_A+b_B)},
\end{equation}
where the three coefficients are
\begin{equation}
\begin{aligned}
P_A &= \sum_{i,k}\bar v^A_{ik}\norm{B_{k,:}}_2^2, &
P_B &= \sum_{k,j}\bar v^B_{kj}\norm{A_{:,k}}_2^2,\\
P_{AB} &= \sum_k\Big(\sum_i\bar v^A_{ik}\Big)
                    \Big(\sum_j\bar v^B_{kj}\Big). &&
\end{aligned}
\label{eq:bitcoeff}
\end{equation}
$P_A$ is the $A$-side product-error cost, $P_B$ its $B$-side counterpart, and
$P_{AB}$ the simultaneous-error coefficient. We can compute all three from the
transformed factors and their groups. For normalized
per-vector fields
$\bar v^A_{ik}=(r_i^A)^2$ and
$\bar v^B_{kj}=(r_j^B)^2$, these formulas reduce to
$P_A=(\sum_i (r_i^A)^2)\norm{B}_F^2$,
$P_B=(\sum_j (r_j^B)^2)\norm{A}_F^2$, and
$P_{AB}=K(\sum_i (r_i^A)^2)(\sum_j (r_j^B)^2)$.

\begin{theorem}[Asymmetric bit-width-sum allocation]
\label{thm:bitalloc}
Assume $P_A,P_B>0$ and allow unconstrained real bit widths under the fixed
bit-width sum $b_A + b_B = b_{\mathrm{sum}}$. The cross term
$P_{AB}2^{-2b_{\mathrm{sum}}}$ is constant under this constraint. The interior
continuous optimum therefore minimizes the two leading terms:
\begin{equation}
b_A^\star = \frac{b_{\mathrm{sum}}}{2} + \frac14\log_2\frac{P_A}{P_B}, \qquad
b_B^\star = \frac{b_{\mathrm{sum}}}{2} - \frac14\log_2\frac{P_A}{P_B},
\qquad\text{i.e.}\qquad
b_A^\star - b_B^\star = \tfrac12\log_2\frac{P_A}{P_B}.
\end{equation}
\end{theorem}

If the available formats impose
$b_A\ge b_A^{\min}$ and $b_B\ge b_B^{\min}$, we clip $b_A^\star$ to
$[b_A^{\min},b_{\mathrm{sum}}-b_B^{\min}]$ and set
$b_B=b_{\mathrm{sum}}-b_A$. For integer widths, we compare the nearest feasible
splits.

The bit-width sum is a precision constraint, distinct from raw dense-factor
storage. A raw storage constraint has the form
\begin{equation}
mK\,b_A+Kn\,b_B=K(mb_A+nb_B)=S.
\label{eq:raw-bit-storage}
\end{equation}
When $m\ne n$, moving along \cref{eq:raw-bit-storage} changes $b_A+b_B$, so the
bilinear term $P_{AB}2^{-2(b_A+b_B)}$ participates in the storage-constrained
optimum.

\begin{remark}[Groupwise allocation]
For a groupwise weighted-storage constraint, the cross term couples the groups,
and the stationarity
conditions become coupled water-filling equations
$2\ln 2\,[P^A_g 2^{-2b^A_g} + \sum_h P^{AB}_{gh}2^{-2(b^A_g+b^B_h)}] = \lambda n^A_g$
with a symmetric equation for $B$.
\end{remark}

The operand with the larger product-weighted coefficient therefore receives more bits.
The exact cross term is constant for one fixed bit-width sum, but it couples
groupwise allocations with unequal group costs. The closed-form bit-width gap
$b_A^\star-b_B^\star$ in \Cref{thm:bitalloc} requires the stated $2^{-2b}$
high-rate law; other formats require their own distortion model.

\subsection{When does deterministic rounding match the noise model?}

After we allocate bits, we choose the rounding rule. The identity of
\Cref{thm:master} is exact under subtractive dither and independent stochastic
rounding with its residue-dependent variances. Deterministic round-to-nearest,
however, can introduce periodic bias and correlations. At a lattice frequency, the phase
$\exp(2\pi\mathrm{i}\ell a_k/\Delta)$ depends only on the residue of $a_k$ modulo the
quantizer step $\Delta$; its group average is therefore an empirical Fourier
coefficient of those residues. Prior quantization-noise analyses use these
coefficients at frequencies
$2\pi\ell/\Delta$, $\ell\in\mathbb Z\setminus\{0\}$,
\citep{widrow1961statistical,sripad1977quantization,gray1998quantization}.
Independent uniform residues have zero population coefficients there. Small
empirical coefficients are consistent with this null, whereas large ones flag
lattice alignment for held-out calibration.

For a scale group of $K$ values $a_k$, aggregate the first $\ell_{\max}$
harmonics into the dimensionless diagnostic
\begin{equation}
\Xi_{\Delta,\ell_{\max}}(a) = \sum_{\ell=1}^{\ell_{\max}}\frac{1}{\ell^2}
\Big| \frac{1}{K}\sum_{k=1}^K
\exp\!\Big(\frac{2\pi\mathrm{i}\,\ell\, a_k}{\Delta}\Big)\Big|^2 .
\label{eq:xi}
\end{equation}
Here $\mathrm{i}^2=-1$, and $\ell_{\max}$ is the largest retained harmonic. Since every
coefficient magnitude is at most one, the omitted tail of the infinite series obeys
\begin{equation}
0 \le \Xi_{\Delta,\infty}(a)-\Xi_{\Delta,\ell_{\max}}(a)
\le \sum_{\ell>\ell_{\max}}\ell^{-2} \le \frac{1}{\ell_{\max}}.
\label{eq:xitail}
\end{equation}
Thus, choosing
$\ell_{\max}\ge\lceil1/\varepsilon_{\rm tail}\rceil$ ensures a truncation
tolerance of $\varepsilon_{\rm tail}$. Under $K$ independent uniform
residues modulo $\Delta$,
$\E\Xi_{\Delta,\ell_{\max}}=H_{\ell_{\max},2}/K$, where
$H_{\ell_{\max},2}=\sum_{\ell=1}^{\ell_{\max}}\ell^{-2}$. Values near this null
value are consistent with uniform residues; substantially larger values indicate
lattice alignment.

A held-out calibration defines an application-specific threshold by limiting
the mismatch between measured and modeled error. \Cref{fig:xi} relates the
diagnostic $\Xi$ to this mismatch on controlled regimes.

\begin{remark}[Quantizer-dependent transform ordering]
Consider
\[
A=\begin{bmatrix}1&1&1&0\end{bmatrix},\qquad B=I_4,
\qquad
H_4=\frac12
\begin{bmatrix}
1&1&1&1\\
1&-1&1&-1\\
1&1&-1&-1\\
1&-1&-1&1
\end{bmatrix}.
\]
Use signed INT8 RTN ($q=127$), one scale for the row of transformed $A$, and one
scale for each column of transformed $B$, without clipping. At $T=I_4$, every
nonzero entry is a range endpoint and the product error is zero. At $T=H_4$,
\[
AH_4=\begin{bmatrix}3/2&1/2&1/2&-1/2\end{bmatrix},
\qquad H_4^\top B=H_4.
\]
The transformed $B$ is represented exactly, while the $A$-row step is
$(3/2)/127$ and $1/2$ is not a lattice point; the RTN product error is therefore
positive. Under non-overloading subtractive dither with the same groups,
\Cref{thm:master} instead gives
\[
\mathcal E_{\mathrm{dither}}(I_4)=16c+16c^2,
\qquad
\mathcal E_{\mathrm{dither}}(H_4)=12c+9c^2.
\]
Thus transform ordering depends on the quantizer. The dither objective ranks
$H_4$ first on this pair, whereas deterministic INT8 RTN ranks $I_4$ first;
independent stochastic rounding can be rescored exactly with the
residue-dependent variances in \Cref{sec:model}.
\end{remark}

The transform-dependent product weights also govern nonuniform codebooks and
weighted Lloyd--Max updates (\Cref{sec:compander}).

\subsection{Clipping under the product-error identity}
\label{sec:clip}

At fixed bit width, the clipping threshold trades two errors: lowering it reduces
the quantizer step and hence \emph{granular} rounding variance, but values outside
the retained range incur deterministic \emph{overload}. Under the assumptions
below, the product-error identity separates overload as a deterministic bias and
granular error as a stochastic variance field. Let
$\tilde A=AT$ and $\tilde B=T^{-1}B$ denote the transformed full-precision
factors. Define
$A' = \mathrm{clip}(\tilde A; \tau^A)$,
$B' = \mathrm{clip}(\tilde B; \tau^B)$, and the clipping bias fields
$C^A = A' - \tilde A$ and $C^B=B'-\tilde B$. We quantize the clipped values
with non-overloading subtractive dither so the granular errors obey the model with
variance fields $v^A(\tau), v^B(\tau)$. For a finite signed codebook, this exact
result requires half-step headroom or guard reconstruction levels as specified in
\Cref{sec:model}. A saturating implementation incorporates endpoint overload in
its bias model.

\begin{theorem}[Clipped product-error identity]
\label{thm:clip}
Under these assumptions,
\begin{equation}
\begin{aligned}
\E\Fnorm{\hat A\hat B - \tilde A\tilde B}^2
&= \Fnorm{C^A \tilde B + \tilde A C^B + C^A C^B}^2 \\
&\quad + \sum_{i,k} v^A_{ik}(\tau)\norm{B'_{k,:}}_2^2
       + \sum_{k,j} v^B_{kj}(\tau)\norm{A'_{:,k}}_2^2 \\
&\quad + \sum_k\Big(\sum_i v^A_{ik}(\tau)\Big)
                    \Big(\sum_j v^B_{kj}(\tau)\Big).
\end{aligned}
\label{eq:clip}
\end{equation}
\end{theorem}

The identity is exact: overload involves the \emph{unclipped} opposite factor,
whereas granular error is weighted by the \emph{clipped} opposite factor's
energies. To obtain a separable threshold rule, we use a diagonal surrogate. For one
scale group, index the unclipped values by $z_s$ and their nonnegative
opposite-factor energy weights by $w_s$. With
$c_b=1/[12(2^{b-1}-1)^2]$, define, for $\tau\ge0$,
\[
M(\tau) = c_b\tau^2\sum_s w_s + \sum_s w_s(\abs{z_s}-\tau)_+^2.
\]
Here $(x)_+=\max(x,0)$. The first term models granular variance and the second
models per-entry overload.
The objective $M(\tau)$ is convex in $\tau$. When the total weight is positive and at least
one weighted sample is nonzero, the optimal per-group threshold is the unique root
of $c_b\tau\sum_s w_s = \sum_s w_s(\abs{z_s}-\tau)_+$. This first-order
condition balances the marginal product-weighted granular and overload terms. The
surrogate replaces the squared norm of the summed systematic bias with the sum of
squared per-entry contributions, dropping cross-coordinate overload terms. The
magnitude of those dropped correlations determines its target-data accuracy.

For a Gaussian population, $\tau^\star/\sigma$ depends only on the number of
quantization levels, whereas max-scaling grows as
$\tau_{\max}\approx\sigma\sqrt{2\ln G}$. \Cref{fig:clip} evaluates the resulting
finite-sample gap. Because $(\tau_{\max}/\tau^\star)^2$ compares granular variance
only, the figure reports the full $M(\tau_{\max})/M(\tau^\star)$ ratio, including
overload.

For fixed thresholds, the granular fold sub-problem remains the geometric program
of \Cref{thm:gpfold}. The overload term requires a general nonlinear treatment
of the joint $(T,\tau)$ problem. A practical alternating scheme uses the convex
threshold update at fixed transform and the granular fold GP at fixed thresholds,
then accepts each fold proposal through the full objective or a line search that
includes overload. A jointly convex formulation remains open.

%% file: src/experiments.tex
\section{Numerical Experiments}
\label{sec:experiments}

Controlled synthetic data let us vary one decision statistic at a time and test
whether error follows the predicted direction. A final trained-classifier experiment
tests whether the dither-derived fold objective transfers to deterministic
round-to-nearest (RTN) on held-out data.
Unless a subsection states otherwise, we quantize the factors to signed 8-bit
using the relevant transform, compute the quantized product, and report the
squared relative Frobenius error
\begin{equation}
\operatorname{relF}^2(\hat C,C)
=\frac{\norm{\hat C-C}_F^2}{\norm{C}_F^2}.
\label{eq:experimental-metric}
\end{equation}
Fixed seeds make the figures reproducible. For arithmetic means, we use two-sided
$95\%$ Student-$t$ confidence intervals (CIs). For paired comparisons on shared
instances, we instead use geometric-mean ratios with two-sided $95\%$
Student-$t$ CIs on the paired log ratios. Captions identify each ratio's
numerator and denominator. The trained-classifier study instead shows all twelve
correlated products and reports descriptive geometric means and quartiles. We
provide complete code, seeds, raw results, checksums, and regeneration instructions in
\Cref{app:reproducibility}.

\subsection{Scaling: logarithmic gain on Gaussian, polynomial on heavy tails}
\label{sec:exp-scaling}

The refinement fact in \Cref{sec:scaling-refinement} orders the scaling schemes,
and \Cref{app:review} predicts that
the range heterogeneity $\rho_A = m\max_i r_i^2/\sum_i r_i^2$ governs the
per-vector gain. The tail of the per-vector scale distribution distinguishes two
regimes. For i.i.d.\
Gaussian factors, the maximum row range concentrates, so
$\rho_A \approx 1 + \log m/\log(2K)$ grows logarithmically. For factors with
heavy-tailed per-vector scales of tail index $\xi$, the maximum follows a
Fr\'echet law, so $\rho_A \approx \Gamma(1-2/\xi)\frac{\xi-2}{\xi}
m^{2/\xi}$ grows polynomially.

\begin{figure}[t]
\centering
\begin{subfigure}[t]{0.49\textwidth}
\centering
\includegraphics[width=\textwidth]{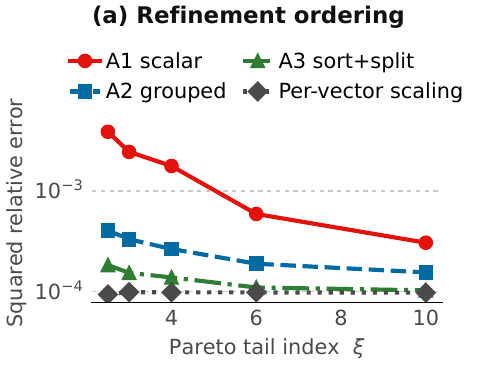}
\phantomsubcaption
\label{fig:ordering}
\end{subfigure}\hfill
\begin{subfigure}[t]{0.49\textwidth}
\centering
\includegraphics[width=\textwidth]{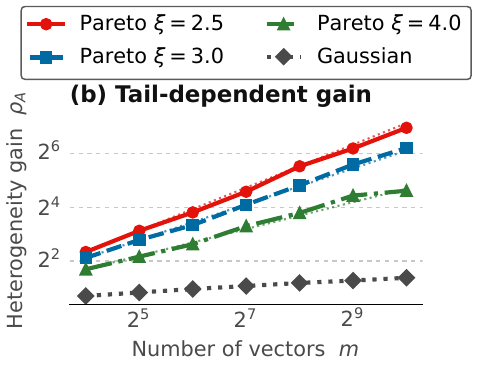}
\phantomsubcaption
\label{fig:heavytail}
\end{subfigure}
\caption{Scaling behavior across tail regimes.
\textup{(a)} Refinement ordering from \Cref{sec:scaling-refinement} on simulated signed
INT8 $256\times256\times256$ products across Pareto tail indices. Means are
computed from 20 independent trials; lower error is better.
\textup{(b)} Per-vector heterogeneity gain $\rho_A$ versus the number of vectors.
It grows polynomially as $m^{2/\xi}$ for heavy tails and logarithmically for
Gaussian data. Dotted lines show the predicted $m^{2/\xi}$ slopes; each point
uses 400 independent trials with $K=128$. See \Cref{sec:exp-scaling}.}
\label{fig:scaling-regimes}
\end{figure}

\Cref{fig:ordering} confirms the ordering
across tail indices, and \Cref{fig:heavytail} confirms the predicted gain:
the measured log-log slopes are $0.79, 0.68, 0.52$ for $\xi = 2.5, 3.0, 4.0$
against the predicted $2/\xi = 0.80, 0.67, 0.50$, while the Gaussian curve is
nearly flat at slope $0.11$. The close slope agreement supports the tail index
as a predictor of the error reduction available from per-vector scaling; the
metadata break-even remains platform-dependent.

\subsection{The full domain-shared fold GP and an idealized one-sided lower bound}
\label{sec:exp-gpfold}

For a fixed rectangular gauge domain, the domain-shared fold theorem establishes
that the bounded shared-fold range-law objective is globally solvable. We generate
dense factors with heterogeneous, approximately reciprocal contraction profiles
so that shared ranges couple the two factors and the joint optimization is nontrivial. We
then solve the theorem's log-domain GP, including both leading terms, the per-row
and per-column range
epigraphs, and the bilinear cross term. We compare it with the identity-fold baseline
and with the sum of two separately optimized one-sided leading bounds. The
latter separately optimizes the two sides and omits the cross term, making it an
idealized lower bound rather than a jointly feasible fold. The reported ratios
compare these modeled objectives
rather than realized $\operatorname{relF}^2$.

Across ten paired trials, \Cref{fig:gpfold} shows a modeled identity-to-GP
objective ratio of $161.2$ ($95\%$ CI $[132.5,196.1]$). The modeled
GP-to-separate-bound ratio is
$4.86$ ($95\%$ CI $[4.73,4.98]$).

\subsection{Asymmetric bit allocation beats the symmetric split}
\label{sec:exp-bitalloc}

\begin{figure}[t]
\centering
\begin{subfigure}[t]{0.49\textwidth}
\centering
\includegraphics[width=\textwidth]{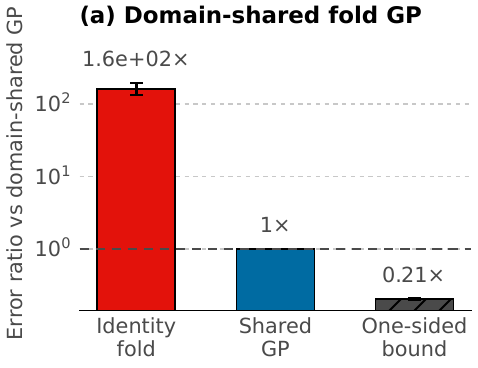}
\phantomsubcaption
\label{fig:gpfold}
\end{subfigure}\hfill
\begin{subfigure}[t]{0.49\textwidth}
\centering
\includegraphics[width=\textwidth]{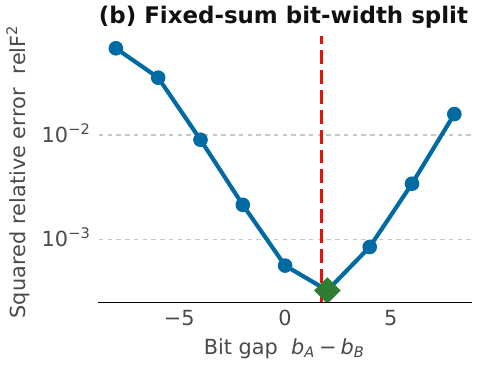}
\phantomsubcaption
\label{fig:bitalloc}
\end{subfigure}
\caption{Optimization predictions for domain-shared folds and bit allocation.
\textup{(a)} Modeled error relative to the numerically optimized domain-shared fold
GP (\Cref{thm:gpfold}) across ten shared-instance trials with $m=n=20$ and
$K=32$. Bars show paired geometric mean ratios with $95\%$
log-Student-$t$ intervals; the hatched one-sided bound is not a feasible joint
fold and omits the cross term.
\textup{(b)} Squared relative Frobenius error versus $b_A-b_B$ at the fixed
bit-width sum $b_A+b_B=16$. The dashed line marks the predicted $1.7$-bit gap from
\Cref{thm:bitalloc}; the diamond marks the measured $9/7$ optimum (gap two),
which beats the symmetric split by $1.7\times$, for a
$256\times256\times256$ product with $10\%$ spiky rows in $A$ and Gaussian
$B$. See \Cref{sec:exp-gpfold,sec:exp-bitalloc}.}
\label{fig:optimization-predictions}
\end{figure}

\Cref{thm:bitalloc} predicts the optimal operand bit split from the
coefficient ratio $P_A/P_B$. We construct operands with different range and
energy profiles: $A$ has a fraction of spiky rows, whereas $B$ is homogeneous.
This construction yields $P_A/P_B \approx 11$ and a predicted gap of $1.7$ bits.
\Cref{fig:bitalloc} sweeps all integer splits at fixed bit-width sum
$b_{\mathrm{sum}} = 16$. The $9/7$ split achieves the measured optimum; its
two-bit gap is nearest the $1.7$-bit prediction, and it beats the symmetric
$8/8$ split by $1.7\times$ in squared relative Frobenius error. This rule
determines the split directly from the product-weighted coefficients without a
search.

\subsection{Log-magnitude clustering beats scalar-norm sorting}
\label{sec:exp-clustering}

\begin{figure}[t]
\centering
\begin{subfigure}[t]{0.49\textwidth}
\centering
\includegraphics[width=\textwidth]{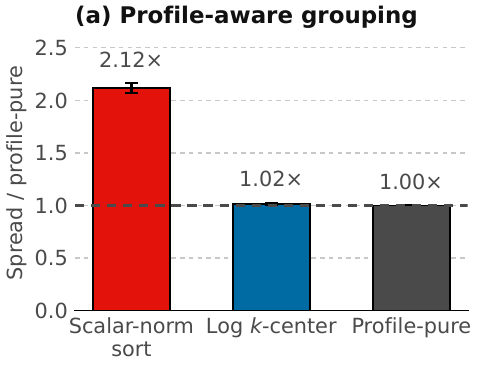}
\phantomsubcaption
\label{fig:clustering}
\end{subfigure}\hfill
\begin{subfigure}[t]{0.49\textwidth}
\centering
\includegraphics[width=\textwidth]{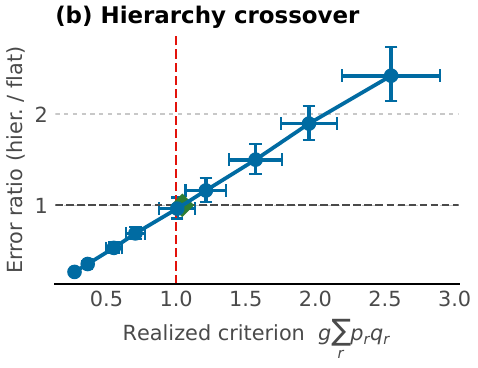}
\phantomsubcaption
\label{fig:hier}
\end{subfigure}
\caption{Controlled tests of data-driven transform-selection criteria.
\textup{(a)} Block-spread ratios relative to the profile-pure partition on
disjoint-support data. Log-magnitude clustering nearly matches one, whereas
scalar-norm sorting is worse in this related profile-structured regime. Bars show
geometric mean paired ratios and $95\%$ log-Student-$t$ intervals over 30
shared-instance trials with five profiles, $m=200$, and $K=60$.
\textup{(b)} Hierarchy-to-flat squared-error ratio against the realized
slice-energy-overlap criterion of \Cref{thm:hier}. The dashed vertical line is
the surrogate boundary and the diamond is the interpolated measured crossover.
Each point uses 28 independent instances with $K=128$ and eight slices. Both
candidates use $g(m+n)$ matched range groups and one shared gauge; horizontal
intervals are Student-$t$ intervals for the mean criterion and
vertical intervals are paired log-Student-$t$ intervals for the error ratio.
See \Cref{sec:exp-clustering,sec:exp-hierarchy}.}
\label{fig:selection-criteria}
\end{figure}

\Cref{thm:cluster} controls the spread of clusters formed in log-magnitude
coordinates, while \Cref{prop:sortfail} quantifies the expected loss of
scalar-norm sorting under random ties. We compare both methods in a related
profile-structured regime: rows drawn from five disjoint-support profiles with
random per-row scales. We measure the block-spread objective
$\sum_I \abs{I}\sum_k \alpha_{I,k}^2\beta_k^2$ for each partition.
For this synthetic instance, we set entries outside each profile's support to
$10^{-3}$ and use the log-magnitude regularizer $\tau_{\log}=10^{-3}$.
Fixing $\tau_{\log}$ in advance isolates the partitioning criterion from the
regularizer choice.
\Cref{fig:clustering} shows that log $k$-center clustering (Gonzalez's farthest-point
traversal in the $\ell_\infty$ log-metric) achieves $1.015$ times the profile-pure
reference, whose blocks each contain a single profile
($95\%$ CI $[1.005,1.025]$), while scalar-norm sorting is $2.116$ times the reference
($95\%$ CI $[2.073,2.161]$). On this instance, choosing the partition in
log-magnitude coordinates recovers almost all of the profile-pure reduction.

\subsection{Rotation on an injected coordinate-outlier family}
\label{sec:exp-rotation}

\Cref{thm:coherence} predicts larger gains for more concentrated coordinate
profiles, but an exactly flat profile offers no coherence gain to an orthogonal
transform.
\Cref{fig:rotation} sweeps the magnitude of one outlier coordinate per row over a dense Gaussian
background, comparing per-vector scaling with and without a
randomized Hadamard rotation.

On the zero-injection Gaussian control, the paired unrotated-to-rotated error
ratio is $0.999$ ($95\%$ CI $[0.989,1.009]$); the interval contains one, consistent
with equal error for an already nearly flat profile.
As the injected outlier grows, the paired gain reaches $59.82$
($95\%$ CI $[59.50,60.14]$) at the largest outlier. The
$K/\log(KN) \approx 12$ factor is a lower-bound scale up to constants; the
realized ratio can be larger. Both range flattening and incoherence contribute
to the measured gain. Together, the flat control and the $59.82\times$ outlier
gain support coordinate concentration as a rotation-selection statistic on
this family.

\subsection{Householders: a few reflectors match full rotation}
\label{sec:exp-householder}

\begin{figure}[t]
\centering
\begin{subfigure}[t]{0.49\textwidth}
\centering
\includegraphics[width=\textwidth]{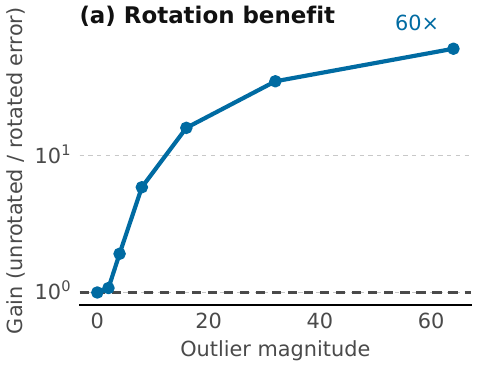}
\phantomsubcaption
\label{fig:rotation}
\end{subfigure}\hfill
\begin{subfigure}[t]{0.49\textwidth}
\centering
\includegraphics[width=\textwidth]{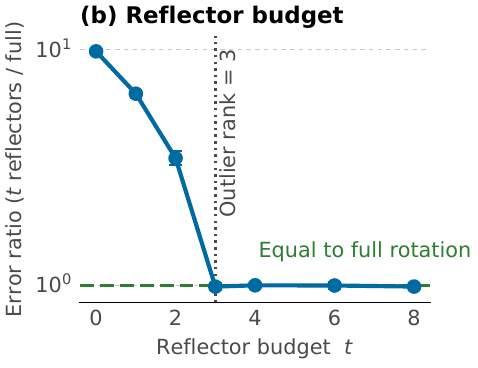}
\phantomsubcaption
\label{fig:householder}
\end{subfigure}
\caption{Orthogonal preconditioning on constructed coordinate-outlier families.
\textup{(a)} Paired unrotated-to-rotated error gain against injected outlier
magnitude for $K=128$ and 30 shared-instance trials.
\textup{(b)} Paired error for $t$ explicit Householder reflectors relative to a
full rotation on rank-$3$ outlier data with $K=128$ and 24 shared-instance
trials. Both panels show geometric mean paired ratios with $95\%$
log-Student-$t$ intervals; the horizontal value one denotes equal error. See
\Cref{sec:exp-rotation,sec:exp-householder}.}
\label{fig:orthogonal-preconditioners}
\end{figure}

\Cref{thm:reflector} predicts that when the weighted-Gram spectrum has a
low-rank head, $O(t)$
reflectors mapping the top-$t$ eigenspace of $M_\mu$ to a flat target frame reduce the spectral
head's contribution while leaving the tail undiscounted. \Cref{fig:householder} sweeps the number of
explicit reflectors over a rank-$3$ coordinate-aligned outlier on a Gaussian
background. The implementation applies the stored reflector vectors directly.

The error falls steeply through $t = 1, 2, 3$ and then flattens. At $t=3$, the
partial-to-full error ratio is $0.988$ ($95\%$ CI $[0.955,1.023]$), so it is
statistically consistent with the full rotation on these 24 pairs. The observed
knee is consistent with the
head/tail mechanism on this constructed low-rank instance.

\subsection{Hierarchy on an anti-correlated slice-energy sweep}
\label{sec:exp-hierarchy}

\Cref{thm:hier} treats the hierarchy as a structured shared gauge and provides
the instance-specific surrogate ratio $g\sum_r p_rq_r$ for comparing it with a flat
rotation. \Cref{fig:hier} sweeps the correlation between the two factors' slice
energies, computes this criterion for every realized pair, and compares both
schemes using the same $g(m+n)=2048$ row-slice and column-slice quantization
groups and one shared gauge per design.

The measured ratio rises monotonically with the criterion: strongly
anti-correlated instances have a hierarchical-to-flat ratio $0.270$
($95\%$ CI $[0.263,0.277]$), while strongly aligned instances give $2.422$
($95\%$ CI $[2.147,2.732]$). Linear interpolation places the measured crossover
at criterion value $1.045$, closely tracking the point where the displayed
upper-bound surrogates cross in this controlled family.

A separate constructed $K=64$ six-level hierarchy illustrates the
contraction-space refinement in \Cref{thm:telescope}: the surrogate reaches its
minimum at depth one because the first covariance increment is negative and the
next five are positive.

\subsection{Clipping and deterministic-rounding diagnostics}
\label{sec:exp-clipping}

\Cref{fig:clip} evaluates the diagonal clipping surrogate $M(\tau)$, including
both granular variance and overload. Varying group size isolates the growth of the
sample maximum from the population threshold. The sample-optimal threshold
approaches the finite-level Gaussian optimum $\tau^*=3.92$ as the group grows;
the common $\sqrt{2\log 255}=3.33$ approximation underestimates this optimum. The measured
max-to-optimal MSE
ratio and the full $M(\tau_{\max})/M(\tau^*)$ ratio agree up to the displayed
precision: both equal $1.00$ at $G=64$ and rise to $1.18\times$ at
$G=65{,}536$.

\FloatBarrier

Varying residue structure while holding the quantizer step and group size
fixed, we evaluate $\Xi_{\Delta,\ell_{\max}}$ in \Cref{fig:xi} without assigning
it a universal threshold.
Near-lattice factors have $\Xi$ about $62$ times the
value under uniform residues and an average absolute log-mismatch of $0.34$ dex
(about a factor of $2.2$) between measured round-to-nearest (RTN) error and the product-error
prediction. Adding residue jitter reduces both; a randomized Hadamard rotation
brings $\Xi$ to $1.04$ times the uniform-residue value and the mismatch to
$0.02$ dex. Thus the roughly $62\times$ diagnostic separation tracks a reduction
in RTN-model mismatch from $0.34$ to $0.02$ dex. Held-out products can map this
relation to an application-specific threshold.

\subsection{Trained-classifier RTN validation on held-out digit images}
\label{sec:exp-trained-digits}

We evaluate candidate selection on all twelve block-linear products of a trained
three-block, width-$64$, four-head ViT-like classifier. The inputs are $8\times8$
handwritten images from scikit-learn, a copy of the UCI
Optical Recognition of Handwritten Digits data
\citep{pedregosa2011scikit,alpaydin1998digits}. The fixed stratified split has
$1{,}078$ training, $323$ calibration, and $396$ held-out test images; the
full-precision test accuracy is $93.94\%$. A fixed subset of 128 calibration
images supplies all $2{,}176$ token rows for fold fitting and
SmoothQuant-style $\alpha$ selection.

For each QKV projection, attention output projection, and pair of MLP products,
we compare thirteen candidates: identity, eleven $\alpha$ values from zero to
one, and a numerically fitted leading-objective GP fold. The SmoothQuant-style
candidate is the scale-gauge--normalized fold
\[
 h_k(\alpha)\propto
 \frac{\max_j\abs{B_{kj}}^{1-\alpha}}
      {\max_i\abs{A_{ik}}^{\alpha}},
 \qquad \alpha\in\{0,0.1,\ldots,1\},
\]
applied as $(A,B)\mapsto(AH,H^{-1}B)$. ``Calibration-selected'' minimizes
measured calibration RTN error over these eleven candidates; the dither-model
selection reported below minimizes the calibration dither prediction over all
thirteen candidates. The same calibration rows define every candidate. Both
factors use symmetric ties-to-even RTN at 8 and 4 bits, with one activation
scale per token row, one weight scale per output column, and no clipping. The
fold vectors are written once by the product sweep and loaded unchanged for the
composed-model evaluation.

\begin{figure}[t]
\centering
\includegraphics[width=\textwidth]{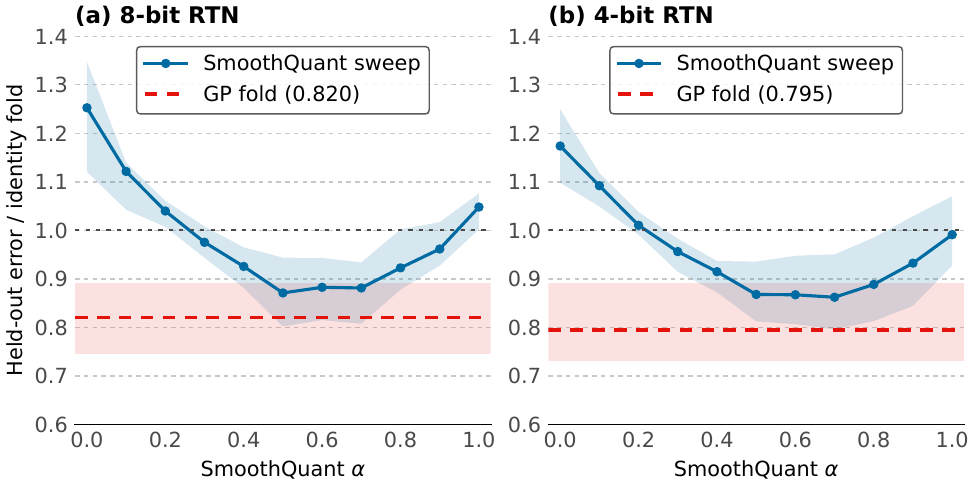}
\caption{\textbf{Shared-fold selection across the SmoothQuant-style $\alpha$
sweep.} Each blue point is the geometric-mean held-out RTN error ratio for one
common $\alpha$ across twelve products; blue shading shows the interquartile
range. The red line and band show the geometric mean and interquartile range of
the product-specific leading-objective GP folds. The horizontal value one is the
identity-fold reference. Both factors are quantized at the displayed precision. See
\Cref{sec:exp-trained-digits}.}
\label{fig:trained-alpha}
\end{figure}

\Cref{fig:trained-alpha} reports held-out product error. Relative to identity,
the GP fold has geometric-mean ratios $0.820$ at 8 bits and $0.795$ at 4 bits,
and improves all twelve products at each precision. The calibration-selected
$\alpha$ grid gives $0.860$ and $0.842$; a test-selected $\alpha$ oracle over
the grid gives $0.858$ and $0.841$. The GP is lower than this oracle in geometric
mean and on ten of twelve products, with worst GP-to-oracle ratios $1.018$ and
$1.007$.

\begin{figure}[t]
\centering
\includegraphics[width=\textwidth]{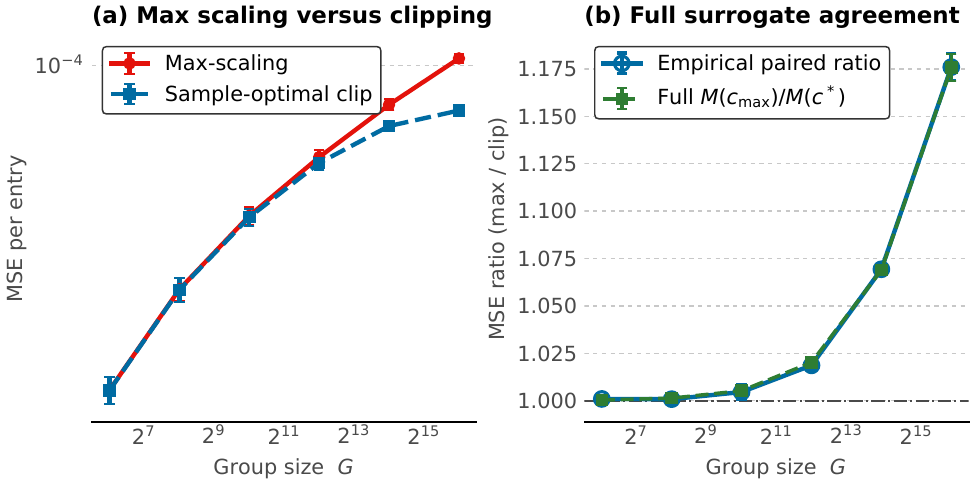}
\vspace{0.5em}
\begin{subfigure}[t]{0.49\textwidth}
\centering
\setcounter{subfigure}{2}
\includegraphics[width=\textwidth]{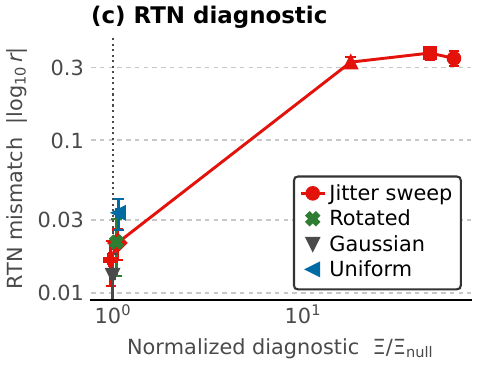}
\phantomsubcaption
\label{fig:xi}
\end{subfigure}\hfill
\begin{subfigure}[t]{0.49\textwidth}
\centering
\includegraphics[width=\textwidth]{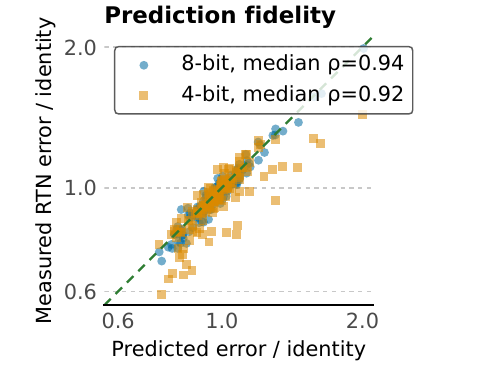}
\phantomsubcaption
\label{fig:trained-prediction}
\end{subfigure}
\caption{Quantizer design and deterministic-rounding model assessment.
\textup{(a)} On Gaussian data, max scaling degrades with group size while the
sample-optimal clipping threshold approaches the finite-level optimum
$\tau^*=3.92$. \textup{(b)} The measured max-to-optimal ratio tracks the full
granular-plus-overload surrogate. Both clipping panels use 255 levels and 240
common-sample trials per group size; their intervals are respectively
Student-$t$ intervals for arithmetic means and paired log-Student-$t$
intervals for geometric mean ratios.
\textup{(c)} The normalized characteristic-function diagnostic against RTN
model mismatch for jittered near-lattice, rotated, Gaussian, and uniform
factors. Intervals are Student-$t$ intervals over 24 independently sampled
draws per regime at a fixed 6-bit step, $K=64$, and $\ell_{\max}=16$.
\textup{(d)} Predicted versus measured held-out RTN error for all thirteen
candidates on twelve products from the trained classifier at each precision, normalized by
the identity within each product. Median within-product Spearman correlations
are $0.937$ and $0.918$. See
\Cref{sec:exp-clipping,sec:exp-trained-digits}.}
\label{fig:clip}
\label{fig:rtn-model-assessment}
\end{figure}

Across all thirteen candidates, the median within-product Spearman correlation
between dither-model prediction and measured error is $0.937$ at 8 bits and
$0.918$ at 4 bits (\Cref{fig:trained-prediction}). Calibration-set dither
prediction selects the held-out winner on ten of twelve products at each
precision; its geometric-mean selection
regret is $1.00194$ and $1.00100$. The simultaneous-error term contributes
$0.00198\%$ and $0.646\%$ of the predicted GP error on average, supporting the
leading-objective fit in this experiment. When all twelve products are quantized
simultaneously, the GP fold gives logit-MSE ratios $0.846$ and $0.736$ relative
to the corresponding identity-fold model; \Cref{app:trained-digits-results} shows the
per-product and composed comparisons. In the composed evaluation, the twelve
block-linear products are quantized while biases, attention QK and AV products,
normalization, nonlinearities, patch embedding, and the classifier head remain
in FP32. These results show dither-model transfer on this trained classifier
using disjoint calibration activations and held-out RTN measurements.

%% file: src/discussion.tex
\section{Discussion}
\label{sec:discussion}

\subsection{Deterministic rounding}
The product-error identity is exact under non-overloading subtractive dither and
independent stochastic rounding with residue-dependent variances.
In contrast, deterministic RTN produces input-dependent bias and inter-coordinate
correlations. To assess these departures, the truncated characteristic-function
diagnostic $\Xi_{\Delta,\ell_{\max}}$ in \Cref{sec:quantizer} summarizes lattice
alignment
at frequencies $2\pi\ell/\Delta$. In the controlled sweep of \Cref{fig:xi},
this diagnostic takes large values alongside the tested near-lattice mismatches.
We use held-out representative products to set an application-specific threshold
and calibrate the modeled-to-realized error ratio.

\subsection{Rotation--scaling composition}
The realized-profile comparison in
\Cref{sec:rotation-folding-substitution} characterizes fixed profiles under
uniform opposite-factor weights. Random rotations introduce a distribution over
profiles, and general product weights couple each coordinate to the opposite
factor. In these regimes, the full product-error objective ranks candidate
transform chains. Prior work explores gradient-based optimization over affine
and structured families \citep{ma2024affinequant,sun2025flatquant}; exact
optimization of the finite-set max-range objective over the full invertible
family $\mathrm{GL}(K)$, including practical condition-number constraints,
remains open.

\subsection{Relation to vector and lattice quantization}
The row-local fold benchmark $\sum_k \norm{A_{:,k}}_2^2\beta_k^2$ of
\Cref{thm:fold} and the incoherence bound of \Cref{thm:coherence} characterize
scalar quantization. \citet{ordentlich2024optimal} establish
information-theoretic limits and nested-lattice constructions for matrix-product
quantization under specified distributional models, and related work develops
practical lattice quantizers
\citep{savkin2025nestquant,kaplan2025highrate}.

Several factors contribute to the gap between scalar performance and these
information-theoretic limits: scalar-to-lattice shaping, overload and clipping,
and rate allocation. The $\pi e/6$ lattice-shaping constant captures the shaping
component, while rotation and clipping control data-dependent range growth
(\Cref{fig:clip}). Identifying the resulting shaping-and-allocation constant
requires a cross-family comparison in a specified data and rate regime.

\subsection{Reuse descriptor: quantized-copy count}
\label{sec:reuse-cost}
Error-based analysis ranks gauges, while distinct gauges assigned across gauge
domains can require additional transformed-and-quantized opposite-factor copies.
Let $n_{\mathrm{opp}}$ denote the number of copies actually required; we call
it the quantized-copy count, a representation-reuse descriptor. By contrast,
$n_{\mathrm{gauge}}$ counts distinct contraction-gauge choices. The counts agree
when each distinct gauge produces one copy. Per-gauge quantizer variants can
make $n_{\mathrm{opp}}>n_{\mathrm{gauge}}$, and coincidental quantized collisions
can make $n_{\mathrm{opp}}<n_{\mathrm{gauge}}$. The storage, preprocessing, and
memory-traffic consequences of these copies depend on the implementation and
can be measured alongside the descriptor.

With a fixed quantizer rule, a single shared gauge along the contraction dimension
requires one quantized copy of each factor. The conditional observation in
\Cref{sec:rotation-folding-substitution} suggests that a rotation can remove some
heterogeneity that a later fold would otherwise exploit. Whether the remaining
error reduction warrants additional copies is a deployment-specific trade-off.

\begin{figure}[ht]
\centering
\includegraphics[width=0.70\textwidth]{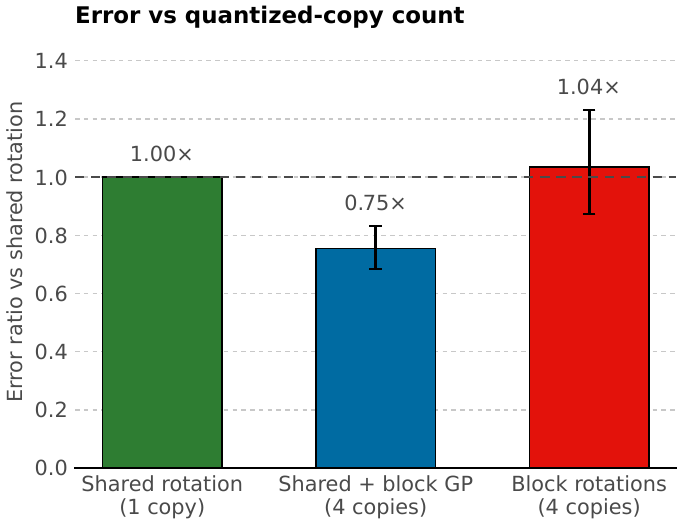}
\caption{Paired deterministic-RTN error ratios relative to a shared rotation on
heavy-tailed operands. The reference bar has $n_{\mathrm{opp}}=1$
transformed-and-quantized copy; the block GP and independent block rotations have
$n_{\mathrm{opp}}=4$. The operands have Pareto
tail index $\xi=3$, $K=16$, and ten shared-instance trials. Bars are geometric
mean ratios with $95\%$ log-Student-$t$ intervals; the block-rotation interval
crosses one. The shared design has 64 scale groups, and each four-copy design
has 160. Here $n_{\mathrm{opp}}$ is the design-level quantized-copy count and
reuse descriptor; kernel measurements supply runtime and traffic. See
\Cref{sec:reuse-cost}.}
\label{fig:reuse}
\end{figure}

We test this trade-off on heavy-tailed operands (\Cref{fig:reuse}) by comparing a
one-copy shared randomized rotation with two four-copy designs: blockwise GP
folds optimized after the shared rotation, and independently drawn block
rotations.
The shared design uses 64 scale groups; each row-block design uses 160 because
every block produces a separate transformed-and-quantized copy of the 32-column
opposite factor. Relative to the shared rotation ($n_{\mathrm{opp}}=1$), the block-GP
design ($n_{\mathrm{opp}}=4$) achieves a paired error ratio of $0.754$ ($95\%$
CI $[0.685,0.831]$), whereas four independent block rotations yield a ratio of
$1.035$ ($95\%$ CI $[0.872,1.229]$). The block-GP interval lies below one, while the
independent-rotation interval spans one. The GP's modeled objective is $0.761$
times its identity-fold value. On these paired instances, the block-GP design
exchanges three additional copies for $24.6\%$ lower error. Platform-specific
kernel measurements can translate this copy/error result into runtime and
traffic; the independent block rotations show no resolved error gain for the
same copy count.

%% file: src/extensions.tex
\section{Extensions}
\label{sec:extensions}

The regularized spread bound of \Cref{thm:cluster} controls the raw spread when
rows share a common support and their nonzero magnitudes are bounded below by a
positive constant. For clipping,
the fixed-transform threshold subproblem is convex, while joint fold--threshold
selection requires evaluating candidate folds under the full overload-aware
surrogate. The trained-classifier study in \Cref{sec:exp-trained-digits} tests twelve
products from one compact vision model. Applying the same selection protocol to
pretrained language models and real token activations would measure transfer
across architectures and data regimes.

\paragraph{Two extensions.}
In attention mechanisms using rotary position embeddings
(RoPE)~\citep{su2024roformer},
the product $QK^\top=(QU)(KU)^\top$ is invariant under any shared orthogonal
$U$ on the head dimension. Before RoPE, $U$ must lie in the commutant of the
position-dependent block rotations. When all plane frequencies are distinct,
this constraint reduces to independent per-frequency phase rotations; repeated
frequencies permit additional mixing within their shared-frequency subspaces.
After RoPE, $U$ is unrestricted but requires inference-time computation. Second,
iterative solvers use residual- or energy-weighted error, so the weighted
product-error identity of \Cref{cor:weighted} applies with $\mathsf L,\mathsf R$
encoding the spectral measure. This identity implies that the selected transform
should flatten the
weighted norms defined by that measure rather than the raw norms.

\paragraph{Open problems.}
Calibration-aware rounding, coupled finite-rate interfaces, approximate-inverse
sketching, and structured $\Sigma\Delta$ noise shaping toward low-energy
contraction coordinates remain open.

%% file: src/conclusion.tex
\section{Conclusion}
\label{sec:conclusion}

The product-error identity and the reuse descriptor $n_{\mathrm{opp}}$ organize
quantized matrix-product design around one objective and an explicit sharing
choice. Five statistics---tail index, profile spread, block coherence,
weighted-Gram spectrum, and slice-energy covariance---guide these choices by
screening the scaling, grouping, rotation, partial-rotation, and hierarchy
families. The bounded domain-shared fold GP gives the exact optimum
of its range-law objective; the broader structural criteria compare explicit
upper bounds and generate candidates that we then evaluate under the
product-error objective.

Controlled experiments isolate each mechanism. Relative to identity-fold
baselines on twelve products from a trained classifier, the GP folds lower
held-out product error by $18.0\%$ at 8 bits and $20.5\%$ at 4 bits in geometric
mean, while composed logit MSE falls by $15.4\%$ and $26.4\%$. The
quantizer-dependent ordering example clarifies the
workflow---score dither and independent stochastic rounding under the exact
identity, then evaluate RTN candidates on representative calibration data.

Natural extensions include exact max-range optimization over broader transform
families, coupled finite-rate transform chains, calibration-aware RTN theory,
validation across pretrained language models, and kernel measurements linking
copy count to storage, traffic, and runtime. The present framework already turns
product-error accounting, transform sharing, and family-specific statistics into
an executable selection procedure for quantized matrix products.

%% file: src/acknowledgments.tex
\section*{Acknowledgments}

The authors acknowledge support from the 2026 Laboratory Directed Research and
Development (LDRD) FORSEE initiative, ``CCSD Core: Foundational Research for
Smart Extreme-scale Ecosystems,'' at Oak Ridge National Laboratory.

%% file: src/quantizer_appendix.tex
\section{Supplementary Notation and Quantizer Design}
\label{app:supplementary-quantizer}

\subsection{Complete notation reference}
\label{app:notation}

\Cref{tab:notation} introduces symbols needed for the main argument.
\Cref{tab:notation-full} collects the complete notation used throughout the
paper and appendices.

\begin{table}[p]
\centering
\caption{\textbf{Complete notation reference.}}
\label{tab:notation-full}
\small
\setlength{\tabcolsep}{2pt}
\setlength{\aboverulesep}{0.25ex}
\setlength{\belowrulesep}{0.35ex}
\ra{0.90}
\rowcolors{2}{gray!25}{white}
\begin{tabular}{@{}
  >{\raggedright\arraybackslash}p{0.13\textwidth}
  >{\raggedright\arraybackslash}p{0.36\textwidth}
  p{0.45\textwidth}
  @{}}
\toprule
Symbol type & Symbol & Description \\
\midrule
& $A \in \R^{m\times K}$, $B \in \R^{K\times n}$ & Full-precision factors to be quantized \\
& $C = AB$ & Exact matrix product \\
& $m,n;N_{IJ},N_{\mathrm{out}}$ & Free dimensions; block-local and global vector counts \\
& $K$ & Contraction dimension, the shared summation index \\
& $i,j,k$ & Row, column, and contraction indices \\
\multirow{-6}{*}{\specialcell{Matrices\\and indices}}
& $I,J,S$ & Row block, column block, and contraction slice \\
\midrule
& $\hat A,\hat B$ & Quantized factors \\
& $E_A,E_B$ & Quantization-error matrices: $\hat A=A+E_A$, $\hat B=B+E_B$ \\
& $v^A_{ik},v^B_{kj}$ & Entrywise error variances \\
& $b_\bullet,b_{\mathrm{sum}},\Delta,R,c$ & Operand bit width, bit-width sum, step, range, and variance coefficient \\
\multirow{-5}{*}{Quantization}
& $\Q(\cdot),\tau$ & Scalar quantizer and clipping threshold \\
\midrule
& $T,T_r$ & Contraction-gauge choices satisfying $AB=(AT)(T^{-1}B)$ \\
& $U,U_t$ & Orthogonal transform; partial Householder transform \\
& $D,H$ & Row-local diagonal fold; domain-shared diagonal fold \\
& $\mathcal P$ & Gauge-domain or contraction-slice partition \\
& $n_{\mathrm{gauge}}$ & Number of distinct contraction-gauge choices $\abs{\{T_r\}}$ \\
& $n_{\mathrm{opp}}$ & Number of required transformed-and-quantized opposite-factor copies; representation-reuse descriptor \\
\multirow{-7}{*}{\specialcell{Transforms\\and grouping}}
& $g,s$ & Number of blocks or slices; slice size \\
\midrule
& $\mathcal E,\mathcal E_A,\mathcal E_B$ & Expected squared product error and its one-sided leading terms \\
& $\Sigma_i^A,\Sigma_j^B$ & Propagated row and column covariances \\
& $\mu_\bullet,s_\bullet^2,\ell_\bullet$ & Mean, variance proxy, and scale for $\bullet\in\{A,B\}$ \\
& $Q_A,Q_B$ & Deterministic Frobenius bounds for dither errors \\
& $P_A,P_B,P_{AB}$ & Bit-allocation coefficients and exact cross-term coefficient \\
& $\Xi_{\Delta,\ell_{\max}}$ & Truncated characteristic-function diagnostic through harmonic $\ell_{\max}$ \\
\multirow{-7}{*}{\specialcell{Error\\statistics}}
& $\mathsf L,\mathsf R$ & Left and right weights for weighted output norms \\
\midrule
& $r_i^A,r_j^B$ & Per-vector row and column ranges; domain-shared fold GP epigraphs \\
& $\beta_k$ & Opposite-factor row norm $\norm{B_{k,:}}_2$; $\beta_k^2$ is its energy \\
& $\alpha_{I,k}$ & Blockwise coordinate range $\max_{i\in I}\abs{A_{ik}}$ \\
& $\rho_I$ & Shared row-block range $\max_{i\in I}\norm{A_{i,:}}_\infty$ \\
& $\rho_A$ & Per-vector range-heterogeneity ratio \\
& $\gamma_{I,k}$ & Per-coordinate spread from sharing a fold across block $I$ \\
& $R_A^{\mathrm{fold}},R_B^{\mathrm{fold}},W_A^{\mathrm{fold}},W_B^{\mathrm{fold}}$
& Exact block-fold range and energy sums \\
\multirow{-8}{*}{\specialcell{Scaling\\and folds}}
& $x_i(k),d(i,i'),\tau_{\log}$ & Log-magnitude profile, distance, and regularizer \\
\midrule
& $R_A^{\mathrm{coh}},R_B^{\mathrm{coh}}$ & Block coherence sums after an orthogonal transform \\
& $\eta_A,\eta_B$ & Normalized coherence factors \\
& $M_\mu$ & Weighted Gram matrix coupling a block's two factors \\
& $\lambda_\ell,P_t,t$ & Eigenvalue, top-eigenspace projector, and reflector budget \\
& $A_r,B_r,p_r,q_r$ & Per-slice energies and normalized slice-energy distributions \\
\multirow{-6}{*}{\specialcell{Rotation,\\structured\\gauges}}
& $\Lambda,\mathcal J(S)$ & Global coherence log factor and contraction-refinement node surrogate \\
\bottomrule
\end{tabular}
\end{table}
\FloatBarrier

\subsection{Optimal codebook density after the transform}
\label{sec:compander}

The bit-allocation rule chooses the number of levels per operand but assumes
uniform spacing. A compander redistributes the reconstruction levels: it applies a
monotone compressor, quantizes uniformly in the compressed coordinate, and maps
back through the expander~\citep{bennett1948spectra,gray1998quantization}.
Its point density records how densely reconstruction levels cover each input
region, giving common or product-sensitive values finer resolution.

Treating this density as continuous, \citet{ang2026product} derive the optimal
codebook under a pair-i.i.d.\ model, where input--output value pairs are
independent and identically distributed. Their matrix-product objective weights
the density by the other factor's conditional second moment. In our
post-transform, per-instance setting, this conditional second moment becomes
the transformed opposite-factor energy. To formalize this, consider a scale
group of $\tilde A$, index its samples by $s=(i,k)$, write
$z_s=\tilde A_{ik}$, and let $k(s)=k$ identify the contraction coordinate. The
product-error identity assigns each sample the
product weight $w_s=\norm{\tilde B_{k(s),:}}_2^2$. If $f_A$ is the density of
$z_s$ and $\omega_A(x)=\E[w\mid z=x]$, the leading $A$-side term of the
product-error identity equals the weighted average of the local quantization variance.

Let a $K_A$-level compander have normalized point density $\lambda$, with
$\int\lambda(x)\,dx=1$. Its high-rate cell width near $x$ is
$[K_A\lambda(x)]^{-1}$, so its local variance is
$[12K_A^2\lambda(x)^2]^{-1}$. Averaging this variance field in the
product-error identity gives
\begin{equation}
D(\lambda)=\frac{1}{12K_A^2}\int f_A(x)\omega_A(x)\lambda(x)^{-2}\,dx.
\label{eq:compander}
\end{equation}
If the normalizer $I_A$ is finite and nonzero, the Lagrange
stationarity condition under $\int\lambda=1$ yields
\begin{equation}
\lambda^\star(x)=\frac{[f_A(x)\omega_A(x)]^{1/3}}{I_A},
\qquad
D(\lambda^\star)=\frac{I_A^3}{12K_A^2},
\qquad
I_A=\int[f_A(x)\omega_A(x)]^{1/3}\,dx.
\end{equation}

For a uniform density on a support of width $W$, the high-rate cell width is
$W/K_A$. A finite symmetric endpoint codebook with $K_A=2q+1$ levels instead has
the exact interior step $W/(K_A-1)=R/q$ and granular variance $cR^2$ under the
convention of \Cref{sec:model}. Because the relative difference between these
widths is $O(K_A^{-1})$, the resulting density is a high-rate surrogate for
the leading product-weighted error. It can be interpreted as a nonuniform
variance field only when the reconstruction errors are independently zero mean
in the \emph{original} coordinate. Nonlinear expansion generally destroys this
zero-mean property, so the exact product-error identity does not follow merely by adding dither in the
compressed coordinate.

A sample-based construction removes the pair-i.i.d.\ assumption. A weighted
Lloyd--Max iteration~\citep{max1960quantizing,lloyd1982least} alternates between
assigning cells and updating reconstruction levels using samples $(z_s,w_s)$.
With fixed levels, assigning samples to the nearest level minimizes weighted
squared error. For a fixed cell $C_r$ with positive total weight, the
error-minimizing level is the weighted centroid
$c_r = \sum_{s\in C_r} w_s z_s / \sum_{s\in C_r} w_s$. Each update cannot increase
the empirical weighted scalar distortion. This coordinatewise
fact guarantees neither a unique nor a globally optimal codebook. The centroid
condition makes each cell's \emph{empirical product-weighted} first error moment
zero, but it does not make errors independent or conditionally zero mean. In particular,
compressed-domain dither followed by a nonlinear inverse expander generally
introduces bias in the original coordinate.

The cross term suggests reweighting within the independent-noise
high-rate surrogate.
For entry $(i,k)$ of $A$, the marginal variance cost is
$\Omega^A_{ik} = \norm{B_{k,:}}_2^2 + \sum_j v^B_{kj}$. Alternating between the two
operands' Lloyd--Max fits weighted by $\Omega^A$ and $\Omega^B$ yields a
monotone fixed-weight scalar heuristic. We then rank finite-rate candidate codebooks
by realized product error. The cross-term correction is often small
at moderate precision but warrants explicit evaluation at low bit widths.

The transform-level high-rate companding objective
$\Phi(T) = \sum_g n_g I_g(T)^3/12 \cdot 2^{-2b_g}$ is the codebook-optimized
analogue of the max-range and coherence objectives, where $n_g$ is the number of
samples and $b_g$ is the bit width in group $g$. Within this surrogate, the best
compander has distortion no greater than the uniform density because the latter
remains feasible. At finite rate, especially with overload or deterministic
rounding, we must instead evaluate the candidate codebook on the actual product
error.

%% file: src/certificate_appendix.tex
\section{From Mean to Certificate}
\label{app:certificate}
\label{sec:concentration}

Expected error averages over noise realizations; it does not bound a single realization. To
obtain a high-probability statement, we assume entrywise \emph{sub-Gaussian}
errors, whose tails decay at least at a Gaussian rate. The Orlicz norm
\[
\norm{X}_{\psi_2}=\inf\{t>0:\E\exp(X^2/t^2)\le2\}
\]
quantifies that tail scale. We assume every error entry, after division by
its standard deviation, has $\psi_2$-norm at most a fixed $\kappa$.
Non-overloading subtractive dither satisfies this variance-normalized assumption.
By contrast, standard stochastic rounding stays within one step but does not satisfy the same
assumption uniformly: near a representable value, its standard deviation can be
arbitrarily smaller than its $\psi_2$-norm. It therefore requires a separate
step-size-based bound.

For each row $i$, the propagated error $(E_A)_{i,:}B$ has covariance
$\Sigma_i^A = B^\top \diag(v^A_{i,:})B$. The squared norm of this propagated error is a quadratic form,
and these forms are independent across rows. The Hanson--Wright concentration
inequality~\citep{rudelson2013hanson} converts the entrywise sub-Gaussian
assumption into a deviation bound for the sum of these forms. The resulting bound uses three statistics:
\begin{align}
\mu_A &= \sum_i \tr\Sigma_i^A,
& s_A^2 &= \sum_i \Fnorm{\Sigma_i^A}^2,
& \ell_A &= \max_i \norm{\Sigma_i^A}_{\mathrm{op}}.
\label{eq:concentration-stats}
\end{align}
Here $\mu_A$ is the $A$-side leading mean in \cref{eq:master}, $s_A$ controls
the square-root deviation, and $\ell_A$ controls the large-deviation linear term.
To analyze the other side, let $D_j^B=\diag(v^B_{:,j})$ and define
\begin{align}
\Sigma_j^B &= A D_j^B A^\top,
& \mu_B &= \sum_j\tr\Sigma_j^B,
& s_B^2 &= \sum_j \Fnorm{\Sigma_j^B}^2,
& \ell_B &= \max_j\norm{\Sigma_j^B}_{\mathrm{op}}.
\label{eq:concentration-stats-b}
\end{align}

\begin{theorem}[One-sided concentration]
\label{thm:concentration}
Under the variance-normalized sub-Gaussian assumption, there is a constant
$C_\kappa$ depending only on $\kappa$ such that, for all $t \ge 0$,
\begin{equation}
\Pr\!\big[\,\big|\Fnorm{E_A B}^2 - \mu_A\big|
> C_\kappa(\sqrt{s_A^2\,t} + \ell_A t)\,\big] \le 2e^{-t},
\label{eq:concentration}
\end{equation}
and symmetrically for $(A,E_B,\mu_B,s_B,\ell_B)$.
\end{theorem}

These one-sided bounds, combined with uniformly bounded subtractive-dither errors,
control all three error terms, though conservatively.

\begin{corollary}[Bounded-dither full-product certificate]
\label{cor:fullcert}
Let $0<\delta<1$ and suppose the errors come from non-overloading subtractive
dither. Write their half-step bounds as
$d^A_{ik}=\sqrt{3v^A_{ik}}$ and $d^B_{kj}=\sqrt{3v^B_{kj}}$, and set
\[
Q_A^2=\sum_{i,k}(d^A_{ik})^2,
\qquad
Q_B^2=\sum_{k,j}(d^B_{kj})^2.
\]
For $t_\delta=\log(4/\delta)$, define
\begin{align*}
U_A(\delta)&=\mu_A+C_\kappa\big(\sqrt{s_A^2t_\delta}+\ell_At_\delta\big),\\
U_B(\delta)&=\mu_B+C_\kappa\big(\sqrt{s_B^2t_\delta}+\ell_Bt_\delta\big).
\end{align*}
Then, with probability at least $1-\delta$,
\begin{equation}
\Fnorm{\hat A\hat B-AB}^2
\le
\Big(\sqrt{U_A(\delta)}+\sqrt{U_B(\delta)}+Q_AQ_B\Big)^2.
\label{eq:full-certificate}
\end{equation}
\end{corollary}

The theorem and corollary apply Hanson--Wright to the independent row quadratic
forms and symmetrically to the independent column quadratic forms. The full
certificate controls all three propagated components via the triangle
inequality and the deterministic bound
$\Fnorm{E_AE_B}\le\Fnorm{E_A}\Fnorm{E_B}\le Q_AQ_B$.
This deterministic bound is conservative because the squared bilinear term need not have a
sub-exponential tail under general sub-Gaussian noise; in the scalar Gaussian
case, for example, this squared bilinear term is a product of two chi-square variables.

Under the per-vector variance field $v^A_{ik} = c (r_i^A)^2$, the covariance
statistics simplify to directly measurable quantities. Because $\Sigma_i^A = c
(r_i^A)^2 B^\top B$, the relative deviation depends on the effective rank of
$B^\top B$ and the range-heterogeneity ratio
$\rho_A = m\max_i (r_i^A)^2/\sum_i (r_i^A)^2$ (measured in
\Cref{sec:experiments}).

These one-sided statistics describe the tail behavior of the propagated leading
terms. Because orthogonal transforms leave $B^\top B$ invariant,
these transforms can improve the Hanson--Wright bounds under per-vector
fields through the same range flattening that reduces the mean. Diagonal folds and coordinate-dependent
fields also reshape the propagated covariance spectrum, so designs with the same
mean can have different leading-term tails. The unspecified $C_\kappa$ and the
coupling of both propagated terms with $Q_AQ_B$ make the certificate a conservative
tail guarantee; calibrated transform selection additionally requires an explicit
$C_\kappa$ or empirical calibration.

%% file: src/appendix.tex
\section{Review of Standard Scaling Schemes}
\label{app:review}

We derive the closed-form errors for the global, per-vector, and block scaling
schemes quoted in \Cref{sec:scaling}, using
only the product-error identity (\Cref{thm:master}) and the noise model
\cref{eq:noise}. Throughout, $c = 1/(12(2^{b-1}-1)^2)$,
$r_i^A = \norm{A_{i,:}}_\infty$,
$r_j^B = \norm{B_{:,j}}_\infty$, and $\beta_k = \norm{B_{k,:}}_2$.

\paragraph{Global scalar.}
A single scale per factor sets $v^A_{ik} = c\norm{A}_{\max}^2$ for all entries.
An outlier in $A$ sets $\norm{A}_{\max}$, inflating the quantization step for
every entry.
The leading terms of
\cref{eq:master} yield
\begin{equation}
\mathcal E_1 = c\big(m\norm{A}_{\max}^2\Fnorm{B}^2 + n\norm{B}_{\max}^2\Fnorm{A}^2\big),
\end{equation}
where $m$ reflects the shared opposite-factor weights across rows of
$A$.

\paragraph{Per-vector.}
Assigning an output-axis scale to each row of $A$ and column of $B$ sets
$v^A_{ik} = c (r_i^A)^2$, yielding
\begin{equation}
\mathcal E_2 = c\Big(\Fnorm{B}^2\sum_i (r_i^A)^2
+ \Fnorm{A}^2\sum_j (r_j^B)^2\Big).
\label{eq:mse2}
\end{equation}
Since $\sum_i (r_i^A)^2 \le m\norm{A}_{\max}^2$ termwise,
$\mathcal E_2 \le \mathcal E_1$. The range-heterogeneity improvement factor is
\[
\rho_A
= \frac{m\max_i (r_i^A)^2}{\sum_i (r_i^A)^2}
\in [1,m].
\]
Two stylized models yield the orders measured in \Cref{sec:experiments}. First,
let $s_1,\ldots,s_m$ be i.i.d.\ Pareto row ranges with
$\Pr(s_i>t)=t^{-\xi}$ for $t\ge1$ and $\xi>2$, and let $s$ denote a generic
copy. Setting $r_i^A=s_i$ gives
\begin{align*}
\frac{\E[\max_i s_i^2]}{\E[s^2]}
&\sim
\frac{\Gamma(1-2/\xi)m^{2/\xi}}{\xi/(\xi-2)},\\
\intertext{whereas for i.i.d.\ entries $A_{ik}\sim\mathcal N(0,\sigma^2)$ and
$r_i^A=\max_k\abs{A_{ik}}$, the leading extreme-value approximation is}
\frac{2\sigma^2\log(2mK)}{2\sigma^2\log(2K)}
&=\frac{\log(2mK)}{\log(2K)}.
\end{align*}

\paragraph{Block.}
For output-axis row-block scaling, let
$\rho_I = \max_{i\in I}\norm{A_{i,:}}_\infty$. The $A$-side error is
\[
c\,\beta_{\mathrm{tot}}\sum_I \abs{I}\rho_I^2,
\]
which lies between the global and per-vector errors, as shown in
\Cref{sec:scaling-refinement}.
The contraction-axis block fold belongs to a different refinement chain: it obeys
\cref{eq:blockbound} and approaches the row-local fold benchmark as blocks shrink
(\Cref{thm:fold}).

\section{Proofs}
\label{app:proofs}

\paragraph{\Cref{thm:master} (product-error identity).}
We expand $\hat A\hat B - AB = E_A B + A E_B + E_A E_B$. The three cross inner
products vanish in expectation because each contains a single factor of
$\E(E_A)$ or $\E(E_B)$, which are zero. For the first squared term,
$\E\Fnorm{E_A B}^2 = \sum_{i,j}\sum_{k,k'}\E[(E_A)_{ik}(E_A)_{ik'}]B_{kj}B_{k'j}
= \sum_{i,k} v^A_{ik}\norm{B_{k,:}}_2^2$ by independence. The second term is symmetric.
For the bilinear term, $\E\Fnorm{E_A E_B}^2 = \sum_{i,j}\sum_{k,k'}
\E[(E_A)_{ik}(E_A)_{ik'}]\,\E[(E_B)_{kj}(E_B)_{k'j}]$, and independence forces
$k = k'$, yielding $\sum_k(\sum_i v^A_{ik})(\sum_j v^B_{kj})$.
\hfill$\square$

\paragraph{\Cref{thm:fold} (row-local fold benchmark).}
Let $S=\{k:a_k\ne0\}$. If $S$ is empty, the maximum in $f$ is zero, so every
positive $d$ attains the value zero. Now suppose $S$ is nonempty, set
$u_k=a_k^2d_k^2$ for $k\in S$, and write $M=\max_{k\in S}u_k>0$. Then
\[
f(d)
=M\left(\sum_{k\in S}\frac{a_k^2\beta_k^2}{u_k}
       +\sum_{k\notin S}\frac{\beta_k^2}{d_k^2}\right)
\ge\sum_{k\in S}a_k^2\beta_k^2
=\sum_k a_k^2\beta_k^2.
\]
To prove the reverse inequality, fix $t>0$, set
$d_k=\sqrt{t}/\abs{a_k}$ on $S$, and $d_k=L$ off $S$. The maximum is $t$
and
\[
f(d)=\sum_k a_k^2\beta_k^2
     +\frac{t}{L^2}\sum_{k\notin S}\beta_k^2,
\]
which approaches the lower bound as $L\to\infty$. If
$\beta_k=0$ off $S$, the same construction with any finite $L$ attains the
infimum. Conversely, if some $\beta_k>0$ off $S$, every finite fold has a strictly
positive second term multiplied by $M>0$, so no finite fold achieves the
infimum.
Summing the row-local infima yields $\mathcal E_A^\star$. Comparing this benchmark
with the block error yields \cref{eq:blockbound}, and the
spread $\gamma_{I,k}$ equals its ratio to the benchmark term. \hfill$\square$

\paragraph{\Cref{prop:sortfail} (random-tie sorting failure).}
Partition the $K$ coordinates into $g$ equal blocks $T_1,\dots,T_g$ and take
$m=g^2$ rows. Form $g$ profile classes of $g$ identical unit-norm rows each, with
class $\ell$ supported uniformly on $T_\ell$. Suppose the scalar sorting key gives
all classes the same value, as permutation-invariant norms do here, and break
ties uniformly at random.

For a random block of $g$ rows, let $D$ be the number of represented profile
classes. With uniform opposite-factor weights,
$\sum_k\alpha_{I,k}^2=D$, so the block cost is $gD$. The profile-pure
partition has total cost $g^2$, whereas the random partition has cost
$g\sum_{r=1}^gD_r$; its expected ratio to the optimum is therefore $\E D$.
Each profile class is absent from a random size-$g$ block with probability
\[
\frac{\binom{g^2-g}{g}}{\binom{g^2}{g}}.
\]
Linearity of expectation gives \cref{eq:random-tie-gap}. The displayed
probability converges to $e^{-1}$, proving the asymptotic form.
\hfill$\square$

\paragraph{\Cref{thm:cluster} (regularized spread control).}
For any $i,i' \in I$ and coordinate $k$,
$\abs{\log(a_{ik}+\tau_{\log})-\log(a_{i'k}+\tau_{\log})}
\le\norm{x_i-x_{i'}}_\infty\le r$. Exponentiating gives the ratio bound in
\cref{eq:clustercontrol}. If $b_i=a_{ik}+\tau_{\log}$, then
\[
\tilde\gamma_{I,k}
=\frac{\abs{I}(\max_i b_i)^2}{\sum_{i\in I}b_i^2}
\le\frac{\abs{I}e^{2r}(\min_i b_i)^2}
{\abs{I}(\min_i b_i)^2}
=e^{2r}.
\]

Now suppose the rows have common support and
$\tau_{\log}\le\epsilon a_{\min,k}$. For every $i\in I$ on an active
coordinate, $a_{ik}+\tau_{\log}\le(1+\epsilon)a_{ik}$. Therefore
\[
\frac{\gamma_{I,k}}{\tilde\gamma_{I,k}}
=\frac{\alpha_{I,k}^2}{(\alpha_{I,k}+\tau_{\log})^2}
 \frac{\sum_{i\in I}(a_{ik}+\tau_{\log})^2}
 {\sum_{i\in I}a_{ik}^2}
\le(1+\epsilon)^2,
\]
which proves \cref{eq:rawtransfer}. Finally, Gonzalez farthest-point traversal
returns a cover radius $R\le2R^\star$ in any metric~\cite{gonzalez1985clustering}.
Nearest-center assignment produces blocks of diameter at most $2R$, so applying
the first part with $r=2R$ gives the stated spread guarantee. \hfill$\square$

\paragraph{Rank-one interval dynamic program.}
In the rank-one case, if $\abs{A_{ik}} = s_i c_k$ with $s_i,c_k\ge0$, then
$\alpha_{I,k} = c_k\max_{i\in I}s_i$, so
the block cost is
$\abs{I}(\max_{i\in I}s_i)^2\sum_k c_k^2\beta_k^2$ up to the common factor $c$.
The block cost thus depends on block cardinality and the maximum scale. An exchange
preserves block cardinalities and does not increase either block maximum, so an
optimal partition has contiguous intervals in $s$-sorted order. Interval dynamic
programming over the $O(m^2)$ interval costs finds the optimal $g$-way split in
$O(m^2 g)$.

\paragraph{\Cref{thm:coherence} (coherence).}
For any unit-norm row $z$ and orthogonal $U$, $\norm{zU}_\infty \le \norm{zU}_2 =
\norm{z}_2$ gives $\eta \le K$ (attained when $zU$ is one-hot, i.e.\ when $z$ is spiky in
$U$'s basis), and $\norm{zU}_\infty \ge \norm{zU}_2/\sqrt K = \norm{z}_2/\sqrt K$
gives $\eta \ge 1$ (attained when $zU$ is flat). For Haar $U$, a fixed normalized
$zU$ is uniform on the sphere and its coordinates have sub-Gaussian tails at scale
$K^{-1/2}$. If a normalized Hadamard matrix $H$ of order $K$ exists, then for
$U=DH$ with independent Rademacher diagonal entries in $D$, each output coordinate
is a Rademacher average, so Hoeffding gives the same scale. A union bound over the
$KN_{IJ}$ coordinates of the stacked block yields the stated uniform bound, hence
$\eta = O(\log(KN_{IJ}/\delta))$. Comparing this upper bound after rotation with
baseline coherence of order $K$ proves the stated
$\Omega(K/\log(KN_{IJ}/\delta))$
gain; it allows larger gains on some instances. \hfill$\square$

\paragraph{\Cref{thm:reflector,cor:reflector-hadamard} (spectral head/tail).}
Let $P_t$ project onto the top-$t$ eigenspace of $M_\mu$, with orthonormal basis
$w_1,\dots,w_t$, and write $W=[w_1,\ldots,w_t]$. Householder QR produces
orthogonal matrices $V_W$ and $V_Q$, each a product of at most $t$ reflectors, such
that $V_W^\top W=E_t$ and $V_Q^\top Q=E_t$, where $E_t$ contains the first $t$
coordinate vectors. Hence $U_t^\top=V_QV_W^\top$ is a product of at most $2t$
reflectors and satisfies $U_t^\top W=Q$.

For Haar $Q$ and any fixed nonzero head vector $zP_t$, the normalized vector
$(zP_t)U_t/\norm{zP_t}_2$ is uniform on the ambient sphere $\mathbb S^{K-1}$.
The standard spherical-coordinate tail is sub-Gaussian with scale
$O(K^{-1/2})$. A union bound over all $K$ coordinates and all
$N_{IJ}$ fixed vectors
therefore gives, with probability at least $1-\delta$,
\[
\norm{(zP_t)U_t}_\infty^2
\le C\frac{\log(2KN_{IJ}/\delta)}{K}\norm{zP_t}_2^2
\]
simultaneously. The logarithm depends on the ambient dimension $K$ because the
head dimension $t$ does not reduce the number of output coordinates.

Given a Hadamard target $Q$, every row of $Q$ has squared norm $t/K$. Writing
$(zP_t)^\top=W a$ and using $W^\top U_t=Q^\top$ gives
\[
\norm{(zP_t)U_t}_\infty=\norm{a^\top Q^\top}_\infty
\le\sqrt{t/K}\,\norm{a}_2
=\sqrt{t/K}\,\norm{zP_t}_2.
\]
This deterministic argument also covers sign-and-permutation randomizations,
which lack the Haar distribution.

For any row $z$, decompose
$z=zP_t+z(I-P_t)$. The triangle inequality separates the two components:
\[
\norm{zU_t}_\infty
\le\norm{zP_tU_t}_\infty+\norm{z(I-P_t)U_t}_\infty.
\]
Because the construction does not fix the tail pointwise, orthogonal invariance
gives
\begin{equation}
\norm{z(I-P_t)U_t}_\infty
\le\norm{z(I-P_t)U_t}_2
=\norm{z(I-P_t)}_2.
\label{eq:reflectortail}
\end{equation}

Apply the Haar bound (or the Hadamard bound) to the head and
\cref{eq:reflectortail} to the tail. After $(a+b)^2\le2a^2+2b^2$, sum over the
weighted collection of $A$-rows and $B$-columns. The head and tail energies are,
respectively,
\[
H_t=\sum_{\ell\le t}\lambda_\ell(M_\mu),
\qquad
T_t=\sum_{\ell>t}\lambda_\ell(M_\mu).
\]
Thus the Haar incoherence factor discounts $H_t$, while $T_t$ remains
undiscounted, giving \cref{eq:headtail}. Substituting $t/K$ for the Haar factor
gives \cref{eq:headtail-hadamard}. \hfill$\square$

\paragraph{\Cref{thm:hier} (anti-correlation).}
For either candidate, define the matched group ranges
\[
R^A_{ir}=\norm{\widetilde A_{i,S_r}}_\infty,
\qquad
R^B_{rj}=\norm{\widetilde B_{S_r,j}}_\infty.
\]
The grouping assumption in \Cref{sec:hier} gives the leading error
\[
c\sum_{i,r}(R^A_{ir})^2\norm{\widetilde B_{S_r,:}}_F^2
+c\sum_{r,j}(R^B_{rj})^2\norm{\widetilde A_{:,S_r}}_F^2.
\]
For a full orthogonal $U$, \Cref{thm:coherence} at dimension $K$ bounds every
slice range by its full-vector maximum. On the common high-probability event,
\[
(R^A_{ir})^2\lesssim\frac{\Lambda}{K}\norm{A_{i,:}}_2^2,
\qquad
(R^B_{rj})^2\lesssim\frac{\Lambda}{K}\norm{B_{:,j}}_2^2.
\]
Orthogonality and summation over the matched slices then bound each of the two
leading terms by
$c\Lambda\norm{A}_F^2\norm{B}_F^2/K$, up to the same universal constant.

For the hierarchical gauge
$U=\operatorname{diag}(U_1,\ldots,U_g)$, applying
\Cref{thm:coherence} at dimension $s$ gives
\[
(R^A_{ir})^2\lesssim\frac{\Lambda}{s}\norm{A_{i,S_r}}_2^2,
\qquad
(R^B_{rj})^2\lesssim\frac{\Lambda}{s}\norm{B_{S_r,j}}_2^2.
\]
Each leading term is therefore bounded by
$c\Lambda\sum_rA_rB_r/s$, with the same universal constant. Suppressing the
common symmetric factor from the two leading terms yields the displayed
surrogates. The hierarchy is one structured shared gauge, both candidates use
the same $g(m+n)$ quantization groups, and their surrogate ratio is
\[
\frac{\mathcal B_{\mathrm{hier}}}{\mathcal B_{\mathrm{flat}}}
=g\frac{\sum_r A_r B_r}{\norm{A}_F^2\norm{B}_F^2}
=g\sum_r p_r q_r.
\]
Here the hierarchy's union bound covers $gN_{\mathrm{out}}$ vectors of length
$s$, so its logarithmic factor is
$2\log(2s(gN_{\mathrm{out}})/\delta)
=2\log(2KN_{\mathrm{out}}/\delta)$, matching
the $N_{\mathrm{out}}$ length-$K$ vectors in the flat transform. Thus the
hierarchical surrogate is lower exactly when $\sum_r p_rq_r<1/g$.
Since $p$ and $q$ sum to one, both have mean $1/g$ over the $g$ slices, and
\[
\sum_r p_r q_r-\frac1g
=\sum_r\left(p_r-\frac1g\right)\left(q_r-\frac1g\right).
\]
The right-hand side is the unnormalized covariance, so its sign orders the
displayed surrogates.
\hfill$\square$

\paragraph{\Cref{cor:weighted} (weighted product-error identity).}
We repeat the proof of \Cref{thm:master} with $\mathsf L(\cdot)\mathsf R$ inserted. The $(i,j)$
entry of $\mathsf L(E_A B)\mathsf R$ is
$\sum_{p,k,q}\mathsf L_{ip}(E_A)_{pk}B_{kq}\mathsf R_{qj}$. Independence therefore gives
$\E\Fnorm{\mathsf L E_A B \mathsf R}^2 = \sum_{p,k}v^A_{pk}\norm{\mathsf L e_p}_2^2
\norm{B_{k,:}\mathsf R}_2^2$. The remaining two terms follow by symmetry.
\hfill$\square$

\paragraph{\Cref{thm:gpfold} (domain-shared fold GP).}
In \cref{eq:foldgp}, each factor $\sum_i (r_i^A)^2$,
$\sum_k \norm{B_{k,J}}^2 h_k^{-2}$, $\sum_j (r_j^B)^2$, and
$\sum_k \norm{A_{I,k}}^2 h_k^2$ is a posynomial in the positive variables
$(h,r^A,r^B)$. Sums and products of posynomials are posynomials, and the
constraints $\abs{A_{ik}}h_k (r_i^A)^{-1}\le 1$ and
$\abs{B_{kj}}h_k^{-1}(r_j^B)^{-1}\le 1$ are monomial $\le 1$. Hence the
program is a GP.

Under $x=\log h$, $u^A=\log r^A$, and $u^B=\log r^B$, every posynomial becomes a
convex log-sum-exp, while every monomial constraint becomes affine. The bilinear
error cross term
$\sum_k(\sum_i v^A_{ik})(\sum_jv^B_{kj})$ involves
$v^A_{ik}=c(r_i^A)^2$ and $v^B_{kj}=c(r_j^B)^2$.
Because these variances are constant in $k$, summing over $k$ gives exactly
\[
Kc^2\Big(\sum_i(r_i^A)^2\Big)\Big(\sum_j(r_j^B)^2\Big),
\]
which is posynomial. The full objective remains a GP.

For any $\lambda>0$, the transformation
$(h,r^A,r^B)\mapsto(\lambda h,\lambda r^A,\lambda^{-1}r^B)$ preserves every
range constraint and all three objective terms. A normalization
$\prod_kh_k=1$ (equivalently $\sum_k\log h_k=0$) or $h_1=1$ is a monomial
equality and therefore preserves the GP form while selecting one representative
of this scale gauge. This one-dimensional normalization redundancy is distinct
from the broader contraction-gauge equivalence
$AB=(AT)(T^{-1}B)$.

To show that a minimizer exists, first eliminate the epigraph variables: because
the objective is
nondecreasing in each $r_i^A$ and $r_j^B$, their minimizing values are the
corresponding entrywise maxima. After we remove zero rows and columns as specified in
\Cref{sec:scaling}, the resulting physical objective is continuous in $h$. Finite
positive bounds on every $h_k$ give a compact reduced feasible set, so the minimum
is attained. (The unreduced epigraph feasible set need not itself be compact because
the range variables grow without bound.) Without those bounds, take
$A=(1,0)$, $B=(0,1)^\top$, and $x=(t,-t)$. The objective is proportional to
$e^{4t}$ and approaches zero as $t\to-\infty$ but never attains it, proving
the stated unconstrained caveat.
\hfill$\square$

\paragraph{\Cref{thm:bitalloc} (asymmetric bit-width-sum allocation).}
Fix $b_A+b_B=b_{\mathrm{sum}}$. The cross term is independent of the split because
\[
P_{AB}2^{-2(b_A+b_B)}=P_{AB}2^{-2b_{\mathrm{sum}}}.
\]
It therefore suffices to minimize
\[
P_A2^{-2b_A}+P_B2^{-2b_B}
\quad\text{subject to}\quad b_A+b_B=b_{\mathrm{sum}}.
\]
Setting the Lagrangian derivative to zero yields
\begin{align*}
P_A2^{-2b_A}&=P_B2^{-2b_B},\\
2^{-2(b_A-b_B)}&=\frac{P_B}{P_A},\\
b_A-b_B&=\frac12\log_2\!\left(\frac{P_A}{P_B}\right).
\end{align*}
Combining the last equality with $b_A+b_B=b_{\mathrm{sum}}$ gives the stated
$b_A^\star,b_B^\star$.

For groupwise weighted-storage constraints the cross terms
$P^{AB}_{gh}2^{-2(b^A_g+b^B_h)}$ depend on individual $b^A_g$, so differentiating
the Lagrangian $\mathcal E - \lambda(\sum_g n^A_g b^A_g + \sum_h n^B_h b^B_h)$
with respect to $b^A_g$ yields the stated coupled water-filling condition.
\hfill$\square$

\paragraph{\Cref{thm:telescope} (telescoping and depth).}
For a single split of node $S$ into children $\{R\}$,
\begin{align*}
\sum_R \mathcal J(R)
&=\sum_R\frac{A_RB_R}{K_S/g_S}
=\frac{g_S}{K_S}\sum_R A_RB_R\\
&=g_S\left(\sum_R p_R q_R\right)\frac{A_SB_S}{K_S}
=g_S\left(\sum_R p_R q_R\right)\mathcal J(S).
\end{align*}
Hence the split contributes the increment
\[
\sum_R\mathcal J(R)-\mathcal J(S)
=\mathcal J(S)\left[g_S\sum_R p_R q_R-1\right].
\]
Summing over all internal nodes telescopes: each non-root node appears once as a
child and, if internal, once as a parent. The sum therefore reduces to
$\mathcal J_{\mathrm{leaves}}-\mathcal J_{\mathrm{root}}$, giving
\cref{eq:telescope}.

The increment is negative iff $g_S\sum_R p_R q_R < 1$;
for a binary split, $2(pq + (1-p)(1-q)) - 1 = (2p-1)(2q-1)$. Because the increment
uses the pooled energies $p_R, q_R$ at node $S$'s scale, its sign is that of a
covariance at that scale and may differ across scales, so the running sum can have
an interior minimum in depth. \hfill$\square$

\begin{proposition}[Hardness of balanced slice design]
\label{prop:slice-hard}
For $g=2$ and positive integer profiles $a_k=b_k$, deciding whether a balanced
partition attains an objective value of at most
$\tfrac12(\sum_k a_k)^2$ is weakly NP-complete. Consequently, minimizing
\cref{eq:slicedesign} is NP-hard.
\end{proposition}

\paragraph{Proof.}
We reduce from \textsc{Partition}. Given positive integers
$x_1,\ldots,x_n$, form $2n$ positive weights consisting of
$M+x_1,\ldots,M+x_n$ and $n$ additional copies of $M$, where $M=1$ suffices.
Any subset of exactly $n$ constructed weights has sum
$nM+\sum_{i\in S}x_i$, where $S$ indexes the selected augmented weights; its
complement has sum $nM+\sum_{i\notin S}x_i$. Thus, the constructed weights admit
an equal-cardinality, equal-sum bipartition if and only if the original instance
admits a partition.

Now set $a_k=b_k$ equal to these constructed weights and let
$W=\sum_k a_k$. For a balanced two-slice partition whose first slice has weight
$y$, the objective in \cref{eq:slicedesign} is
\[
y^2+(W-y)^2=\frac{W^2}{2}+2\left(y-\frac W2\right)^2.
\]
It is at most $W^2/2$ exactly when $y=W/2$. This proves NP-hardness, and membership
of the threshold decision problem in NP is immediate by evaluating a proposed
partition. For this restricted positive-integer problem, a dynamic program indexed
by selected cardinality and cumulative weight decides whether $n$ items sum to
$W/2$ in time polynomial in the number of items and $W$. The restricted decision
problem is therefore weakly NP-complete. The reduction establishes NP-hardness of
the unrestricted minimization problem but leaves its classification as weakly or
strongly NP-hard open. \hfill$\square$

\paragraph{\Cref{thm:concentration,cor:fullcert} (one-sided concentration and certificate).}
Let $e_i$ denote row $i$ of $E_A$. Independence across rows gives
\[
\Fnorm{E_A B}^2=\sum_i e_i(BB^\top)e_i^\top.
\]
Write $e_i=D_i^{1/2}g_i$, where $D_i=\diag(v^A_{i,:})$ and $g_i$ is isotropic
sub-Gaussian. The $i$th summand is then a quadratic form with kernel
\[
D_i^{1/2}BB^\top D_i^{1/2}.
\]
Its nonzero spectrum equals that of $\Sigma_i^A=B^\top D_iB$.

Hanson--Wright~\cite{rudelson2013hanson} gives this quadratic form a sub-gamma
tail. Its variance proxy is of order $\Fnorm{\Sigma_i^A}^2$, and its scale is
of order $\norm{\Sigma_i^A}_{\mathrm{op}}$. Summing the independent rows adds the
variance proxies to $s_A^2$ and takes their largest scale, $\ell_A$. The corresponding
mean is $\sum_i\tr\Sigma_i^A=\mu_A$, yielding \cref{eq:concentration}.

Applying the same argument to the independent columns of $E_B$ gives the symmetric
bound with $(\mu_B,s_B,\ell_B)$. Set $t=t_\delta=\log(4/\delta)$. A union bound
bounds the propagated terms by $U_A(\delta)$ and $U_B(\delta)$ with probability at
least $1-\delta$.

For non-overloading subtractive dither,
$\abs{(E_A)_{ik}}\le d^A_{ik}$ and
$\abs{(E_B)_{kj}}\le d^B_{kj}$ deterministically. Hence
\[
\Fnorm{E_AE_B}
\le \Fnorm{E_A}\norm{E_B}_{\mathrm{op}}
\le \Fnorm{E_A}\Fnorm{E_B}
\le Q_AQ_B.
\]
On the simultaneous event, the triangle inequality applied to
$E_AB+AE_B+E_AE_B$ yields
\[
\Fnorm{\hat A\hat B-AB}
\le\sqrt{U_A(\delta)}+\sqrt{U_B(\delta)}+Q_AQ_B.
\]
Squaring proves \cref{eq:full-certificate}; the square explicitly retains the
cross terms between all three propagated components. \hfill$\square$

\subsection{Low-dimensional and tropical profiles}
\label{app:latent-profiles}

Rank-one profiles and arbitrary-profile $g$-center are two endpoints of the
partitioning problem. When profiles exhibit low-dimensional structure, the
latent dimension rather than the ambient dimension $K$ determines the
clustering guarantee.

\begin{proposition}[Latent-dimension spread control]
\label{prop:latent}
Let $\phi: Z \to \ell_\infty^K$ be $\kappa_\phi$-Lipschitz from a metric space
$(Z, d_Z)$, and suppose the regularized profiles satisfy
$\norm{x_i - \phi(z_i)}_\infty \le \varepsilon$ for latent points $z_i$. If a block
$I$ has latent diameter $r$, then
$\mathrm{diam}_\infty\{x_i : i\in I\} \le \kappa_\phi r + 2\varepsilon$, and
$\tilde\gamma_I \le e^{2(\kappa_\phi r+2\varepsilon)}$.
\end{proposition}

\begin{corollary}[Latent-space covering]
\label{cor:latent-cover}
Under the hypotheses of \Cref{prop:latent}, Gonzalez's traversal in $Z$ returns
a latent radius of at most $2r^\star_Z$, where $r^\star_Z$ is the optimal latent
$g$-center radius. If $(Z,d_Z)$ has doubling dimension $d$ and diameter $D$,
then for $\log\Gamma>4\varepsilon$,
$O((\kappa_\phi D/(\log\Gamma-4\varepsilon))^d)$ blocks suffice to reach spread
$\Gamma$, up to constant factors and independently of $K$.
\end{corollary}

\paragraph{Proof of \Cref{prop:latent,cor:latent-cover}.}
For $i,i'\in I$, $\norm{x_i - x_{i'}}_\infty \le \norm{\phi(z_i)-\phi(z_{i'})}_\infty
+ 2\varepsilon \le \kappa_\phi\,d_Z(z_i,z_{i'}) + 2\varepsilon$, so the block's
$\ell_\infty$-diameter is at most $\kappa_\phi r + 2\varepsilon$. Substituting into
\Cref{thm:cluster}'s $e^{2\,\mathrm{diam}}$ bound gives $\tilde\gamma_I \le
e^{2(\kappa_\phi r+2\varepsilon)}$. If $\log\Gamma>4\varepsilon$, cover $Z$ by balls of
radius $(\log\Gamma-4\varepsilon)/(4\kappa_\phi)$. Each resulting block has twice that
latent diameter, so its exponent is at most $\log\Gamma$. The standard
doubling-dimension cover uses
$O((\kappa_\phi D/(\log\Gamma-4\varepsilon))^d)$ such balls. Gonzalez traversal supplies the
stated constant-factor $g$-center radius. \hfill$\square$

\paragraph{Tropical specialization.}
Suppose the log-magnitude profiles admit a max-plus approximation with
$q_{\mathrm{tpl}}\ge1$ templates:
\[
x_i(k)=\max_{\ell\le q_{\mathrm{tpl}}}
       (u_{i\ell}+v_{\ell k})+e_{ik},
\qquad \norm{e_i}_\infty\le\varepsilon.
\]
For the associated coefficient map, fix $k$ and set
$a_\ell=u_\ell+v_{\ell k}$ and $b_\ell=u'_\ell+v_{\ell k}$. Then
$\abs{\max_\ell a_\ell-\max_\ell b_\ell}
\le\max_\ell\abs{a_\ell-b_\ell}
=\max_\ell\abs{u_\ell-u'_\ell}$, uniformly in $k$. Thus, the coefficient map is
$1$-Lipschitz from $(\R^{q_{\mathrm{tpl}}},\ell_\infty)$, and
\Cref{prop:latent} applies directly.
The case $q_{\mathrm{tpl}}=1$ recovers the one-dimensional rank-one geometry.
For $q_{\mathrm{tpl}}\ge2$, finding an optimal factorization is generally hard,
but an approximate factorization remains useful because its residual enters the
bound explicitly as $\varepsilon$. Because crossing templates can order rows
differently across coordinate ranges, whether an exact dynamic program can compute
the $g$-block optimum in this setting remains open.

\paragraph{\Cref{thm:clip} (clipped product-error identity).}
We decompose $\hat A\hat B - \tilde A\tilde B = (A'B' - \tilde A\tilde B) + E_A B' +
A' E_B + E_A E_B$. The first summand is deterministic and equals $C^A\tilde B +
\tilde A C^B + C^A C^B$ by expanding $A' = \tilde A + C^A$, $B' = \tilde B + C^B$.
Every cross term between this deterministic part and a one-sided error term carries
a single zero-mean factor. The cross term between the deterministic part and
$E_AE_B$ also vanishes because
$\E(E_AE_B)=0$ by mutual independence. The error--error cross terms vanish as in
\Cref{thm:master}. The remaining expectation is the product-error identity
evaluated at the clipped factors $(A',B')$, giving \cref{eq:clip}.

Each term $c_b\tau^2\sum_s w_s$ and $w_s(\abs{z_s}-\tau)_+^2$ has nonnegative
second derivative in $\tau$, so $M(\tau)$ is convex. Between
breakpoints,
\[
M'(\tau)=2c_b\tau\sum_s w_s
-2\sum_s w_s(\abs{z_s}-\tau)_+.
\]
If the total weight is positive and some weighted sample is nonzero, this derivative
is strictly increasing from a negative value at zero to a positive value beyond
the largest sample, so its root is unique. Finally,
$M(\tau^\star)>c_b(\tau^\star)^2\sum_sw_s$ whenever the optimum clips a weighted
sample, whereas max scaling has zero overload. Thus
$M(\tau_{\max})/M(\tau^\star)<(\tau_{\max}/\tau^\star)^2$ in that case, proving
that the squared-threshold ratio overestimates the clipping improvement whenever
the optimum clips a weighted sample. \hfill$\square$

\paragraph{\Cref{prop:foldsplit} (computable identity-fold criterion).}
Write $a_k=\norm{A_{I,k}}_2^2$ and $b_\ell=\norm{B_{\ell,J}}_2^2$. Expanding the
three products in \cref{eq:block-fold-full} gives terms of the forms
\begin{align*}
R_A^{\mathrm{fold}}(x)W_B^{\mathrm{fold}}(x)
&=\sum_{i,\ell}\max_k
  \left\{A_{ik}^2b_\ell e^{2x_k-2x_\ell}\right\},\\
R_B^{\mathrm{fold}}(x)W_A^{\mathrm{fold}}(x)
&=\sum_{j,k}\max_\ell
  \left\{B_{\ell j}^2a_k e^{2x_k-2x_\ell}\right\},\\
R_A^{\mathrm{fold}}(x)R_B^{\mathrm{fold}}(x)
&=\sum_{i,j}\max_{k,\ell}
  \left\{A_{ik}^2B_{\ell j}^2e^{2x_k-2x_\ell}\right\}.
\end{align*}
Zero-coefficient terms may be omitted. Each displayed maximum is log-convex
because its logarithm is a maximum of affine functions of $x$; it is therefore
convex. Their nonnegative weighted sum $F_{I,J}(x)$ is convex. The explicit
expansion establishes convexity. The function is invariant under
$x\mapsto x+\gamma\mathbf1$, so the
constraint $\mathbf1^\top x=0$ fixes only the scale-gauge invariance of the
log-fold parameterization and leaves the broader class of contraction-gauge
transformations unrestricted.

The directional derivative of each maximum is the maximum over its active set;
differentiating the four factors and applying the product rule gives
\cref{eq:fold-directional}. This derivative is convex and piecewise linear;
rewriting each negative minimum as a maximum yields an epigraph linear program.
Since $d=0$ is feasible, its optimum is at most zero. First-order
optimality for convex functions forces that optimum to zero exactly when the identity-fold
point is optimal; a negative optimum supplies a strict descent direction.
\hfill$\square$

\section{Additional Trained-Classifier Results}
\label{app:trained-digits-results}

\begin{figure}[t]
\centering
\includegraphics[width=\textwidth]{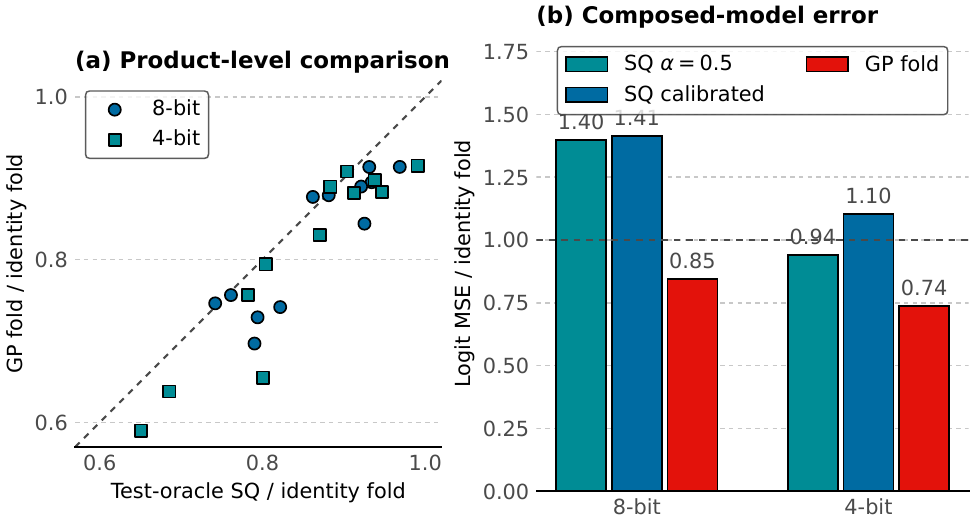}
\caption{\textbf{Product-level and composed deterministic-RTN error.}
\textup{(a)} Held-out GP-fold error versus the best test-selected point on the
eleven-value SmoothQuant-style $\alpha$ grid, each normalized by the identity fold. The GP
is lower on ten of twelve products at both precisions.
\textup{(b)} Relative logit MSE after quantizing all twelve products
simultaneously. At 8 bits, the ratios for $\alpha=0.5$, calibration-selected
$\alpha$, and the GP are $1.399$, $1.413$, and $0.846$; at 4 bits they are
$0.941$, $1.104$, and $0.736$. Bars have visible borders and display their
values above the fill. See \Cref{sec:exp-trained-digits}.}
\label{fig:trained-layer-composition}
\end{figure}

\Cref{fig:trained-layer-composition} separates the product-level comparison
from composed-model error for the experiment in
\Cref{sec:exp-trained-digits}. Each marker in panel (a) is one of the twelve
block-linear products; points below the diagonal favor the leading-objective GP
fold over the test-selected SmoothQuant-style grid. In panel (b), we apply all
twelve stored folds simultaneously while leaving the patch embedding and
classifier head in full precision. Because the products belong to one trained network, the
plot reports descriptive ratios for every product rather than independent-product
confidence intervals.

%% file: src/reproducibility.tex
\section{Reproducibility and Artifact Release}
\label{app:reproducibility}
\label{sec:reproducibility}

\paragraph{Repository and release status.}
The code, numerical outputs, and vector figures are versioned in the
\texttt{qmm-transformations} repository at
\url{https://github.com/piyush314/qmm-transformations}. This paper's source tree
includes the repository as a Git submodule pinned to artifact commit
\texttt{edd9ee8}.
The repository is private while ORNL
release clearance, the approved copyright notice, and the open-source license
are finalized; the same URL will host the public artifact.  Within the
repository, the checked-in results are divided into a controlled synthetic suite
and a trained-classifier suite using round-to-nearest (RTN). Each suite has its
own tested interpreter, top-level package pins, and checksum manifest.

\paragraph{One-command regeneration.}
The synthetic suite uses CPython 3.10.16 with the top-level pins in
\texttt{scripts/requirements-figures.txt}; the trained-classifier suite uses
CPython 3.13.11 with the top-level pins in
\texttt{experiments/trained\_digits/requirements.txt}.  After creating these
environments, running
\begin{center}
\texttt{scripts/reproduce\_all.sh --paper-root ../..}
\end{center}
from the code repository regenerates both suites, runs their focused tests,
and mirrors byte-identical PDFs into this paper tree.  This command fixes
\texttt{SOURCE\_DATE\_EPOCH=0} and uses repository-relative paths.

\paragraph{Synthetic suite.}
The synthetic runner executes 13 scripts, producing 13 plotted PDFs and 14
machine-readable JSON files; the additional JSON records the unplotted
hierarchy-depth calculation.  The synthetic manifest,
\texttt{results/manifest.json}, records every
script, seed, output path, statistical estimand, package pin, and SHA-256
checksum.  Repeated-trial means use two-sided $95\%$ Student-$t$ intervals;
paired comparisons use geometric-mean ratios and Student-$t$ intervals on
their log ratios.  Exact-identity checks and solver diagnostics remain in the
raw records as implementation tests.
In the released result schema, the machine-readable identifier
\texttt{no\_fold} denotes the identity-fold baseline used in the paper and
figures.

\paragraph{Trained-classifier suite.}
The trained-classifier suite evaluates 12 linear products, two bit widths, and
13 fold candidates, giving 312 held-out product measurements.  The evaluation loads the
tensor-only checkpoint with \texttt{weights\_only=True}; fixed split indices
identify disjoint training, calibration, and test sets.  The calibration stage
writes the fitted fold vectors once, and the composed-model evaluation reloads
that archive, so the reported product and network measurements use identical
folds.  The suite manifest,
\texttt{results/trained\_digits/manifest.json}, hashes the checkpoint, split,
numerical outputs, three PDFs, environment specifications, and experiment
source.  Because the 12 products belong to one trained network,
their summaries are descriptive geometric means, quartiles, ranges, rank
correlations, and selection regrets rather than independent-sample confidence
intervals.

\paragraph{Randomness and metrics.}
The synthetic scripts use local NumPy generators with seeds
$0,1,0,5,3,17,11,13,7,9,2,31,$ and $21$ in manifest order.  The trained-classifier
suite uses seed 1234 for training and 20260720 for calibration and evaluation.  Its
stored checkpoint and split make retraining optional for reproducing the
reported measurements.  Matrix-product panels report squared relative
Frobenius error,
\[
  \lVert \widehat C-C\rVert_{\mathrm F}^{2}/
  \lVert C\rVert_{\mathrm F}^{2},
\]
whereas the scalar clipping panel reports per-entry mean squared error.

%% file: src/related_chronology.tex
\clearpage
\section{Selected Chronology of Related Work}
\label{app:chronology}

\Cref{sec:related} organizes prior work by technical thread. The table below
adds a chronological view of representative milestones that directly motivate
our noise model, transform families, gauge terminology, quantizer choices, and
product-weighted objective. Year ranges are ordered by their earliest publication.

\begin{table}[!ht]
\centering
\fontsize{9}{10.5}\selectfont
\setlength{\tabcolsep}{3pt}
\setlength{\aboverulesep}{0.25ex}
\setlength{\belowrulesep}{0.35ex}
\renewcommand{\arraystretch}{1.02}
\caption{\textbf{Representative milestones for
reuse-aware contraction-gauge design.}}
\label{tab:chronology}
\rowcolors{2}{gray!12}{white}
\begin{tabular}{@{}
  >{\raggedright\arraybackslash}p{0.075\textwidth}
  >{\raggedright\arraybackslash}p{0.31\textwidth}
  >{\raggedright\arraybackslash}p{0.555\textwidth}
  @{}}
\toprule
Years & Representative work & Connection to this paper \\
\midrule
1948--82 &
\citeauthor{bennett1948spectra}; \citeauthor{max1960quantizing};
\citeauthor{lloyd1982least} &
\textsc{Scalar quantization.} Companding, point densities, and least-squares
codebooks; their distortion objectives apply to individual factors. \\

1961--77 &
\citeauthor{widrow1961statistical}; \citeauthor{schuchman1964dither};
\citeauthor{sripad1977quantization} &
\textsc{Noise model.} Lattice-frequency analysis and exact dither conditions;
we propagate entrywise variances through a two-factor product. \\

1963 &
\citeauthor{huang1963block} &
\textsc{Rate allocation.} Transform-coefficient allocation under a fixed rate;
we split a fixed bit-width sum between operands using product weights. \\

2006--17 &
\citeauthor{ailon2006fast}; \citeauthor{suresh2017distributed};
\citeauthor{alistarh2017qsgd} &
\textsc{Rotate--quantize.} Fast or structured rotations and stochastic
quantization guarantees; we apply these tools to two-factor products under
transform reuse. \\

2018--23 &
\citeauthor{evenbly2018gauge}; \citeauthor{tindall2023gauging} &
\textsc{Gauge terminology.} Inverse-pair freedom on contracted tensor bonds;
here the bond is the QMM contraction index and sharing determines copy count. \\

2019 &
\citeauthor{meller2019factorization}; \citeauthor{nagel2019equalization} &
\textsc{Scaling.} Function-preserving channel and range equalization; we solve
the bounded domain-shared range-law fold objective as a geometric program. \\

2019--24 &
\citeauthor{banner2019aciq}; \citeauthor{dong2019hawq};
\citeauthor{young2024cvxq} &
\textsc{Quantizer control.} Tensor clipping and network- or size-level bit
budgets; our rules operate on both operands of one product. \\

2021--22 &
\citeauthor{kovaleva2021bertbusters}; \citeauthor{dettmers2022llmint8} &
\textsc{Outliers.} Persistent transformer dimensions and their low-precision
failure motivate profile-aware scaling, grouping, and rotation. \\

2022 &
\citeauthor{kuzmin2022fp8}; \citeauthor{vargaftik2022eden} &
\textsc{Error and rotation.} Scalar-product cross terms and structured-rotation
guarantees; we use fixed matrices and entrywise variance fields. \\

2023--24 &
\citeauthor{xiao2023smoothquant}; \citeauthor{yuan2023rptq};
\citeauthor{lin2024awq} &
\textsc{Folds and grouping.} Diagonal transfer, range-aware clustering, and
data-driven scales; we add certified optimization of the bounded domain-shared
range-law fold objective. \\

2023 &
\citeauthor{frantar2023gptq} &
\textsc{Product weighting.} A one-sided product-reconstruction loss; our
product-error identity treats both quantized operands and their cross term. \\

2023--24 &
\citeauthor{chee2023quip}; \citeauthor{ashkboos2024quarot};
\citeauthor{tseng2024quipsharp} &
\textsc{Fixed rotations.} Incoherence processing and Hadamard or lattice
implementations; we derive block-local and hierarchical upper-bound criteria. \\

2024--25 &
\citeauthor{shao2024omniquant}; \citeauthor{ma2024affinequant};
\citeauthor{liu2025spinquant}; \citeauthor{hu2025ostquant};
\citeauthor{sun2025flatquant} &
\textsc{Learned transforms.} Diagonal, orthogonal, and structured affine
families; we analyze restricted families with explicit criteria. \\

2025--26 &
\citeauthor{savkin2025nestquant}; \citeauthor{kaplan2025highrate};
\citeauthor{ordentlich2024optimal}; \citeauthor{ang2026product} &
\textsc{Product QMM.} Matrix-product codebooks, high-rate rules, and
information-theoretic limits; we focus on per-instance transform and reuse
selection. \\

2026 &
\citeauthor{sanjeet2026blockrot}; \citeauthor{feng2026rht};
\citeauthor{federici2026cat} &
\textsc{Rotation analysis.} Finite-block and randomized-Hadamard guarantees
and concentration--alignment diagnostics complement our selection criteria. \\
\bottomrule
\end{tabular}
\end{table}

\FloatBarrier